\documentclass[journal,onecolumn]{IEEEtran}
\usepackage[utf8]{inputenc}
\usepackage{xcolor}
\usepackage{mathrsfs}
\usepackage{multirow}
\usepackage{multicol}
\usepackage{graphicx}
\usepackage{amsmath, amssymb, latexsym}
\usepackage{tikz}
\usetikzlibrary{decorations.pathreplacing}
\usetikzlibrary{fadings}
\usepackage[font=small,skip=0pt]{caption}
\usepackage{tikz}
\usepackage[edges]{forest}
\usepackage{caption}
\usepackage{longtable}
\usepackage{hyperref}
\usepackage{booktabs}
\usepackage{tabularray}
\newcommand{\ra}[1]{\renewcommand{\arraystretch}{#1}}
\usepackage[most]{tcolorbox}
\tikzset{parent/.style={align=center, text width=8cm,rounded corners=3pt},
    child/.style={align=center,text width=3cm,rounded corners=3pt}
    }
\usepackage{lscape}
\usepackage[normalem]{ulem}
\usetikzlibrary{trees}
\usepackage{array}
\newcolumntype{L}{>{\arraybackslash}m{3cm}}
\usepackage{amsmath}
\usepackage{amssymb}
\usepackage[top=2.5cm,left=2.5cm,right=2.5cm,bottom=2.5cm]{geometry}
\usepackage{pifont}
\title{A Survey on Training Challenges in Generative Adversarial Networks for Biomedical Image Analysis}

\begin{document}

\author{\IEEEauthorblockN{Muhammad Muneeb Saad\IEEEauthorrefmark{1},
Ruairi O'Reilly\IEEEauthorrefmark{2}, and Mubashir Husain Rehmani\IEEEauthorrefmark{3}} \\
\IEEEauthorblockA{Department of Computer Science,
Munster Technological University (MTU), Ireland\\
Email: \IEEEauthorrefmark{1}muhammad.saad@mycit.ie,
\IEEEauthorrefmark{2}ruairi.oreilly@mtu.ie,
\IEEEauthorrefmark{3}mubashir.rehmani@mtu.ie}}

\maketitle

\begin{abstract} 
In biomedical image analysis, the applicability of deep learning methods is directly impacted by the quantity of image data available. This is due to deep learning models requiring large image datasets to provide high-level performance. Generative Adversarial Networks (GANs) have been widely utilized to address data limitations through the generation of synthetic biomedical images. GANs consist of two models. The generator, a model that learns how to produce synthetic images based on the feedback it receives. The discriminator, a model that classifies an image as synthetic or real and provides feedback to the generator. Throughout the training process, a GAN can experience several technical challenges that impede the generation of suitable synthetic imagery. First, the mode collapse problem whereby the generator either produces an identical image or produces a uniform image from distinct input features. Second, the non-convergence problem whereby the gradient descent optimizer fails to reach a Nash equilibrium. Thirdly, the vanishing gradient problem whereby unstable training behavior occurs due to the discriminator achieving optimal classification performance resulting in no meaningful feedback being provided to the generator. These problems result in the production of synthetic imagery that is blurry, unrealistic, and less diverse. To date, there has been no survey article outlining the impact of these technical challenges in the context of the biomedical imagery domain. This work presents a review and taxonomy based on solutions to the training problems of GANs in the biomedical imaging domain. This survey highlights important challenges and outlines future research directions about the training of GANs in the domain of biomedical imagery. 
\end{abstract}

\providecommand{\keywords}
{
  \small	
  \textbf{\textit{Keywords---}}
}

\section{Introduction}
\label{section:Introduction}
Generative adversarial networks (GANs) refer to the class of generative models that generate synthetic data by learning through probability distributions of real data \cite{goodfellow2014generative}. GANs are designed with generator and discriminator models. The generator produces realistic-looking synthetic data while taking random vectors as inputs. The discriminator's task is to classify real data from generated (synthetic) data. GANs use an objective function as a joint loss function with minimax optimization. The generator aims to produce realistic data and misguides the discriminator to classify it as real. Contrarily, the discriminator aims to classify synthetic data as fake and real data as real. The discriminator backpropagates its gradient feedback to the generator. The generator updates its learning to generate realistic synthetic data based on the discriminator's gradient feedback. Ideally, the training of the GANs should be continued until it achieves the Nash equilibrium so that the actions of the generator and discriminator models do not affect each other's performance. At this stage, the generator becomes well-trained so that it uses random vectors to generate synthetic data that closely resemble the real data. 

In healthcare technology, GANs have been widely utilized for several tasks such as pattern analysis of biomedical imagery \cite{bhattacharya2020gan} \cite{qin2020gan} \cite{shi2020knowledge}, electronic health records \cite{lee2020generating}, as well as drug discovery \cite{zhao2020gansdta}. Recently, GANs have also been contributing in the context of Coronavirus disease (COVID-19), i.e., disease detection from chest radiography \cite{waheed2020covidgan}. In the domain of biomedical imagery, the availability of data is an obstacle to the application of deep learning. Deep learning models are composed of deep neural networks, that require large training datasets for better predictive analytics \cite{bhattacharya2020gan}. Thus, enhancing the size of biomedical datasets is a challenging problem. Another dilemma in the biomedical imaging domain is class-imbalanced datasets. It refers to the datasets with skewed classes when dealing with multiple disease classes. With class-imbalanced datasets, deep neural networks train better on the classes with a large number of images rather than the class with a limited number of images \cite{saini2020deep}. Data augmentation is one of the potential solutions to address the class imbalance, as well as data limitation problems \cite{qasim2020red}. 

The utility of GANs in biomedical image analysis has been extensively investigated to perform image recognition \cite{mao2020abnormality}, image synthesis \cite{zhou2020hi}, image reconstruction \cite{li2020high}, and image segmentation \cite{liu2019joint}. GANs have demonstrated a capacity to support deep learning models through the generation of synthetic images and thus enlarging the size of biomedical datasets \cite{tegang2020gaussian} \cite{han2020infinite} \cite{pollastri2020augmenting}. GANs suffer from training challenges such as mode collapse, non-convergence, and instability problems. With these limitations, GANs can generate unrealistic, blurry, and less diverse images. The mode collapse problem occurs when the generator produces similar output images while taking different input features. In the domain of biomedical imaging, the mode collapse problem of GANs has been addressed by using minibatch discrimination \cite{xue2019synthetic}, skip connections \cite{segato2020data}, VAEGAN \cite{kwon2019generation}, varying layers of generator and discriminator \cite{qin2020gan}, spectral normalization \cite{xu2020low}, perceptual image hashing \cite{Neffseg2017}, Gaussian mixture model as generator \cite{wu2018end}, discriminator with conditional information vector \cite{modanwal2021normalization}, self-attention mechanism \cite{saad2022self} \cite{abdelhalim2021data}, and adaptive input-image normalization \cite{saad2022addressing}. The non-convergence problem occurs due to the lack of GAN's ability to reach Nash equilibrium. This problem has been addressed by using modified training updates of generator and discriminator \cite{biswas2019synthetic}, Whale optimization algorithm \cite{goel2021automatic}, and two time-scale update rules \cite{abdelhalim2021data}. The instability problem of GANs occurs due to the vanishing gradient problem. The Wasserstein loss \cite{xue2019synthetic} \cite{segato2020data} \cite{kwon2019generation} \cite{deepak2020msg}, residual connections \cite{wei2020predicting}, multi-scale generator \cite{ciGAN2018}, and Relativistic hinge loss \cite{saad2022self} techniques are identified to address the instability problem in the biomedical imagery.

Several survey articles have identified technical solutions to address the problems of mode collapse, non-convergence, and instability \cite{wiatrak2019stabilizing} \cite{jabbar2020survey} \cite{sampath2021survey} \cite{saxena2020generative}. In the general imaging domain, few survey articles discuss each problem with solutions based on objective functions and modified architectures of GANs while missing the definition, identification, and quantification methodologies The quantification methods are discussed as evaluation metrics in two survey articles \cite{pan2019recent} \cite{gui2020review} while covering almost all aspects of each problem. The existing literature discussed these training challenges of GANs in general and did not cover the significant solutions to address these challenges in the domain of biomedical imaging. There are only four survey articles \cite{kazeminia2020gans} \cite{singh2021medical} \cite{you2022application} \cite{alamir2022role} that only cover these challenges with their definitions and identifications in the biomedical imaging domain. These survey articles outline application-based problems of GANs and have no information about quantification, and solutions to the training challenges of GANs in the biomedical imaging domain. In this survey article, we define each training problem of GANs with their definition, identification, quantification, and existing solutions. A detailed comparison of this work with the existing survey articles is indicated in Table \ref{tabsummary}.

\subsection{Contributions of this Paper}

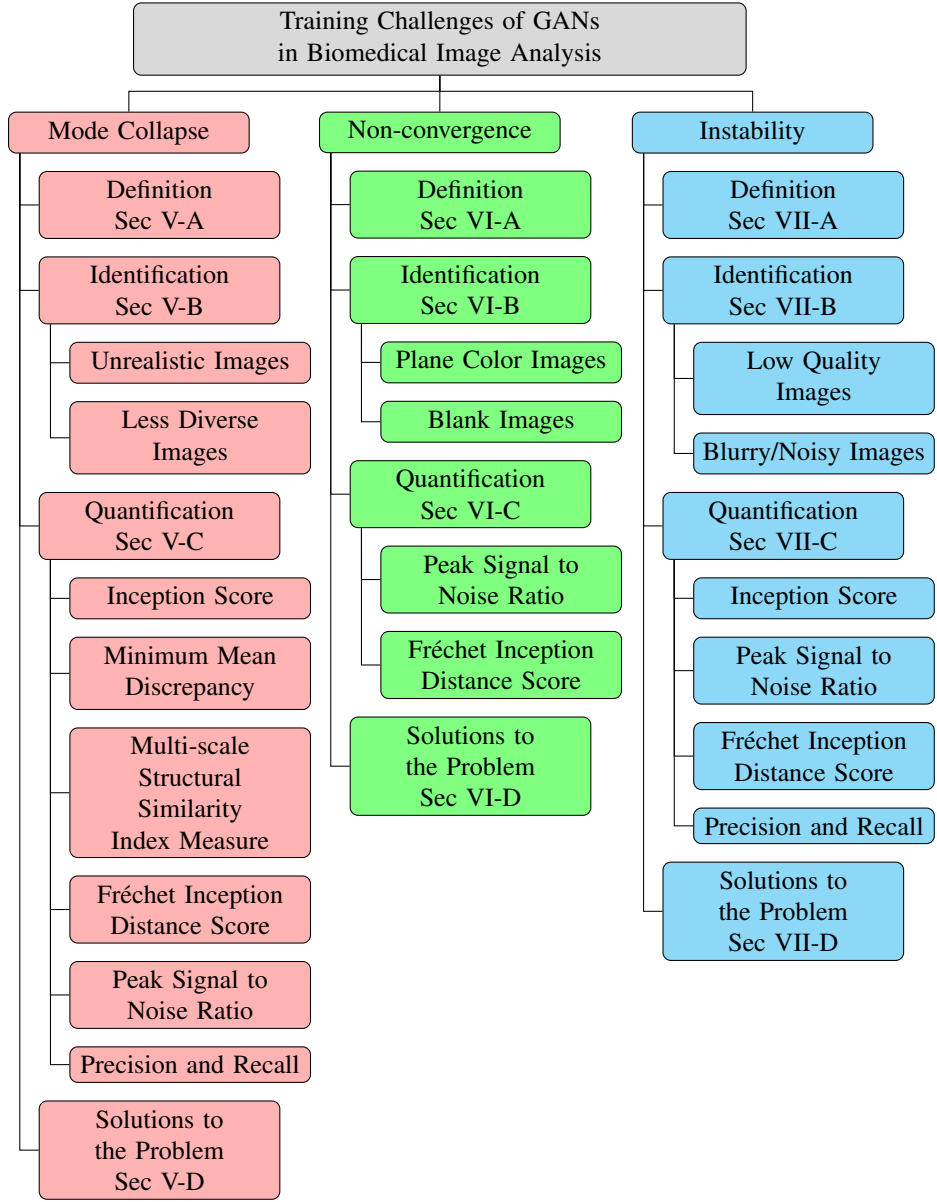
\begin{figure}
    \centering
\begin{center}
        \resizebox{0.75\textwidth}{!}{%
            \begin{forest}
                for tree={
                    draw,
                    rounded corners,
                    node options={align=center},
                    text width=2.7cm,
                    anchor=center,
                },
                where level=0{%
      }{%
        folder,
        grow'=0,
        if level=1{%
          before typesetting nodes={child anchor=north},
          edge path'={(!u.parent anchor) -- ++(0,-20pt) -| (.child anchor)},
        }{},
        }
                [ Training Challenges of GANs in Biomedical Image Analysis, fill=gray!30, parent
                [Mode Collapse, for tree={fill=red!30, child}
                [Definition \\ Sec \ref{section: mode collapse: Def}]
                [Identification \\ Sec \ref{section: mode collapse: Ident}
                [Unrealistic Images]
                [Less Diverse \\ Images]]
                [Quantification \\ Sec \ref{section: mode collapse: Quant}
                [Inception Score]
                [Minimum Mean Discrepancy]
                [Multi-scale Structural \\ Similarity Index Measure]
                [Fr\'echet Inception Distance Score]
                [Peak Signal to Noise Ratio]
                [Precision and Recall]
                ]
                [Solutions to the Problem \\ Sec \ref{section: mode collapse: Soltn}]]
                [Non-convergence, for tree={fill=green!50,child}, calign with current edge
                [Definition \\ Sec \ref{section: Non-convergence: Def}]
                [Identification \\ Sec \ref{section: Non-convergence: Ident}
                [Plane Color Images]
                [Blank Images]
                ]
                [Quantification \\ Sec \ref{section: Non-convergence: Quant}
                [Peak Signal to Noise Ratio]
                [Fr\'echet Inception Distance Score]
                ]
                [Solutions to the Problem \\ Sec \ref{section: Non-convergence: Soltn}]]
                [Instability, for tree={fill=cyan!40, child}
                [Definition \\ Sec \ref{section: Instability: Def}]
                [Identification \\ Sec \ref{section: Instability: Ident}
                [Low Quality \\ Images]
                [Blurry/Noisy Images]
                ]
                [Quantification \\ Sec \ref{section: Instability: Quant}
                [Inception Score]
                [Peak Signal to Noise Ratio]
                [Fr\'echet Inception Distance Score]
                [Precision and Recall]
                ]
                [Solutions to the Problem \\ Sec \ref{section: Instability: Soltn}]]]
            \end{forest}
        }
    \end{center}
      \caption{Taxonomy of training challenges in GANs for biomedical image analysis.}
    \label{fig: GANs training}
    \end{figure}

The main contributions of this survey article are listed as follows:

\begin{itemize}
    \item In this article, we discuss training challenges of GANs like mode collapse, non-convergence, and instability in detail.
    \item We classify each of these training challenges into four different categories i.e., Definition, Identification, Quantification, and available solutions as shown in Fig. \ref{fig: GANs training}.
    \item We also review the existing approaches in terms of different biomedical imaging modalities and classify them into applications-based taxonomies for each problem.
    \item This survey identifies research gaps and provides future research directions for GANs in the domain of biomedical imagery.
\end{itemize}

\begin{table}[htp!]
\centering
\caption{\textbf{An overview of existing survey articles discussing three training problems of GANs based on definition, identification, quantification, and solution from the technical literature in the general and biomedical imagery domain.}.
}

\begin{tabular}{p{2.3cm}p{3cm}p{0.5cm}p{0.2cm}p{0.4cm}p{0.4cm}p{0.4cm}p{0.2cm}p{0.4cm}p{0.4cm}p{0.4cm}p{0.2cm}p{0.4cm}p{0.4cm}p{0.4cm}}

\toprule

\textbf{Main Domain} & \textbf{References} & \textbf{Year} & \multicolumn{4}{c|}{\textbf{Mode Collapse}} & \multicolumn{4}{c|}{\textbf{Non-Convergence}} & \multicolumn{4}{c|}{\textbf{Instability}} \\
&  &  & Def: & Ident: & Quant: & Sol: & Def: & Ident: & Quant: & Sol: & Def: & Ident: & Quant: & Sol: \\

\midrule

\multirow{10}{*}{General Imagery} & Pan et al. \cite{pan2019recent} & 2019 & x & \checkmark & \checkmark & \checkmark & x & \checkmark & \checkmark & \checkmark & x & \checkmark & \checkmark & \checkmark \\
& Hong et al. \cite{hong2019generative} & 2019 & \checkmark & \checkmark & x & \checkmark & \checkmark & \checkmark & x & \checkmark & \checkmark & \checkmark & x & \checkmark \\
& Wiatrak et al. \cite{wiatrak2019stabilizing} & 2019 & \checkmark & \checkmark & x & \checkmark & \checkmark & \checkmark & x & \checkmark & \checkmark & \checkmark & x & \checkmark \\
& Lee et al. \cite{lee2020regularization} & 2020 & \checkmark & x & x & \checkmark & x & x & x & x & x & x & x & x \\
& Jabbar et al. \cite{jabbar2020survey} & 2021 & \checkmark & \checkmark & x & \checkmark & \checkmark & \checkmark & x & \checkmark & \checkmark & \checkmark & x & \checkmark \\
& Gui et al. \cite{gui2020review} & 2021 & \checkmark & \checkmark & \checkmark & \checkmark & \checkmark & \checkmark & \checkmark & \checkmark & \checkmark & \checkmark & \checkmark & \checkmark \\
& Shamsolmoali et al. \cite{shamsolmoali2021image} & 2021 & \checkmark & \checkmark & x & \checkmark & \checkmark & \checkmark & x & \checkmark & \checkmark & \checkmark & x & \checkmark \\
& Wang et al. \cite{wang2021generative} & 2021 & \checkmark & \checkmark & x & \checkmark & x & \checkmark & x & \checkmark & x & \checkmark & x & \checkmark \\
& Sampath et al. \cite{sampath2021survey} & 2021 & \checkmark & \checkmark & x & \checkmark & \checkmark & \checkmark & x & \checkmark & \checkmark & \checkmark & x & \checkmark \\
& Saxena et al. \cite{saxena2020generative} & 2021 & \checkmark & \checkmark & x & \checkmark & \checkmark & \checkmark & x & \checkmark & \checkmark & \checkmark & x & \checkmark \\
& Kossale et al. \cite{kossale2022mode} & 2022 & \checkmark & \checkmark & \checkmark & \checkmark & \checkmark & \checkmark & x & x & \checkmark & \checkmark & x & x \\
&&&&&&&& \\
\multirow{5}{*}{Biomedical Imagery} & Kazeminia et al. \cite{kazeminia2020gans} & 2020 & \checkmark & x & x & x & \checkmark & x & x & x & x & x & x & x \\
& Singh et al. \cite{singh2021medical} & 2021 & \checkmark & x & x & x & x & x & x & x & \checkmark & x & x & x \\
& Aram You et al. \cite{you2022application} & 2022 & x & \checkmark & x & x & x & x & x & x & x & x & x & x \\
& Al Amir et al. \cite{alamir2022role} & 2022 & \checkmark & x & \checkmark & x & \checkmark & x & \checkmark & x & \checkmark & x & \checkmark & x \\
& This Work & 2023 & \checkmark & \checkmark & \checkmark & \checkmark & \checkmark & \checkmark & \checkmark & \checkmark & \checkmark & \checkmark & \checkmark & \checkmark \\

\bottomrule
\multicolumn{15}{l}{Def: Definition; Ident: Identification; Quant: Quantification; Sol: Solution}
\end{tabular}
\label{tabsummary}
\end{table}
\subsection{Organization of the Paper}
The rest of the article is organized as follows; Section \ref{section:GANs} presents the detailed working of GANs including background, basic architecture, and popular variants. Section \ref{section:application_of_GANs} highlights the applications of GANs in biomedical imagery. Section \ref{eval_metrics} discusses the benchmark evaluation metrics used for quantifying the training challenges of GANs. Section \ref{section:modecollapse} discusses the mode collapse problem definition, identification, quantification, and existing solutions. Section \ref{section:Non-convergence} elaborates on the non-convergence problem in the training of GANs, its identification methods, and how to quantify the problem and its possible existing solutions. Section \ref{section:Instabilityproblem} explains the instability problem in the training of GANs while providing a literature review of existing identification and quantification methods, and possible solutions in biomedical imagery. Section \ref{section:comparisongans for covid} provides a comparative analysis of existing GANs architectures for the biomedical imaging domain. Section \ref{section:challengesanddirections} discusses the important challenges and future research directions. Finally, Section \ref{section:conclusion} concludes the paper.

\section{Generative Adversarial Networks}
\label{section:GANs}
GANs are advanced machine learning models that are introduced to generate synthetic images by learning the probability distributions of real images. GANs work as learning agents that try to produce realistic images using probability distributions. To gain an understanding of GANs; the architecture, training, objective function, and GANs variants are elaborated as follows:

\subsection{Architecture of GANs}
\begin{figure}[htp]
    \centering
    \includegraphics[width=1\textwidth]{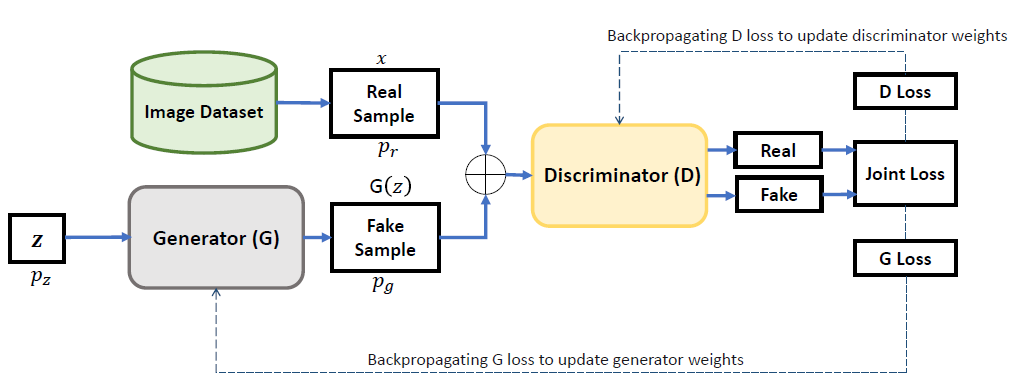}
    \caption{Architecture of Vanilla GANs. The generator $G$ and the discriminator $D$ are trained in an adversarial manner so that $G$ can generate plausible fake samples while $D$ can classify them from real samples. $G$ uses random vector input $z$ for generating fake samples. $G$ loss is described as $log (1-D(G(z)))$ while $D$ loss is $log (D(x))$. The figure is redesigned from \cite{goodfellow2014generative}.}
    \label{Fig.1}
\end{figure}

GANs are composed of two models; the generator and the discriminator. The generator's primary task is to create synthetic data that resembles the real data distribution such as images, sounds, or texts \cite{wolterink2021generative}. For image data, it takes random vector $z$ with probability distribution $p_{z}$ (usually drawn from a normal distribution) as input and generates synthetic image samples $G(z)$ with probability distribution $p_{g}$. The generator is designed with a series of learnable layers, typically consisting of fully connected (dense) or transposed convolutional (deconvolutional) layers. These layers help the generator to upsample the random noise vector $z$ and generate synthetic images in the desired format. The discriminator consists of learnable layers, such as fully connected or convolutional layers for downsampling the images. The discriminator distinguishes the synthetic image samples from real samples. It aims to output high values (close to 1) for real image data and low values (close to 0) for synthetic image data. Goodfellow et al. \cite{goodfellow2014generative} proposed the idea of vanilla GANs (baseline GAN) as shown in Fig. \ref{Fig.1}. The vanilla GAN's generator and discriminator models are composed of fully connected layers using multi-layer perceptron (MLP) neural networks.

\subsection{Training of GANs}
In GANs, adversarial training is the fundamental training technique that involves training two neural networks, the generator, and the discriminator, in a competitive manner, where they learn from each other through an adversarial process. A GAN's training initializes with random weights of the generator and the discriminator. The generator takes the random noise vector $z$ as input and produces synthetic images. The synthetically generated images are fed into the discriminator, along with real images from the actual dataset. The discriminator's task is to distinguish between real and synthetic images and assigns probabilities to each image sample being real or fake. The generator aims to generate images that are realistic enough to misguide the discriminator into classifying it as real. It tries to minimize the discriminator's ability to differentiate between real and synthetic image samples. The discriminator tries to correctly classify real images as real (assigning high probabilities) and synthetic images as fake (assigning low probabilities) \cite{wiatrak2019stabilizing} \cite{jabbar2020survey} \cite{goodfellow2016nips}.

The training process of GANs continues iteratively, with the generator and discriminator playing a minimax game against each other. The generator aims to generate images that look increasingly realistic, while the discriminator strives to become better at distinguishing real from synthetic images. The training converges to Nash equilibrium when the generator generates images that are indistinguishable from real images, and the discriminator can no longer differentiate between the real and synthetic images. The key idea behind adversarial training in GANs is that the generator gets better at producing realistic images by trying to outsmart the discriminator, and the discriminator becomes more adept at distinguishing real from fake images by learning from the generator's synthetic image samples. This competition and feedback loop between the generator and discriminator lead to the emergence of a well-trained GAN capable of generating high-quality synthetic images \cite{salimans2018improving} \cite{saxena2020generative} \cite{wang2021generative}.

\subsection{Objective Function of GANs}

The objective function of a GAN is defined by the distance between the probability distribution of the generated samples ($p_{g}$) and the probability distribution of real samples ($p_{r}$). The binary cross-entropy loss is used to evaluate the objective function. The binary cross-entropy $V(D, G)$ is a joint loss function of the discriminator and the generator. It minimizes the Jensen-Shannon divergence (JSD) between the distribution of generated data as well as real data distribution. The JSD is defined as Eq. \eqref{eq:JSD} \cite{goodfellow2014generative}.

\begin{equation}
\begin{aligned}
\operatorname{JSD}\left(\mathbb{P}_{r} \| \mathbb{P}_{g}\right)=\frac{1}{2} \mathrm{KL}\left(\mathbb{P}_{r} \| \mathbb{P}_{A}\right)+\frac{1}{2} \mathrm{KL}\left(\mathbb{P}_{g} \| \mathbb{P}_{A}\right)\label{eq:JSD}
\end{aligned}
\end{equation}

In Eq. \eqref{eq:JSD}, KL is defined as the Kullback-Leibler divergence, $\mathbb{P}_{r}$ and $\mathbb{P}_{g}$ represent the real and generated data distributions. $\mathbb{P}_{A}$ denotes the average distribution between real and generated distributions. The objective function becomes minimax $V(D, G)$ of $G$ and $D$ as presented in Eq. \eqref{eq:1} reproduced from \cite{gui2020review}. 

\begin{equation}
\begin{aligned}
\min _{G} & \max _{D} V(D, G)=E_{x \sim p_{\text {r}}}[\log D(x)] &+E_{z \sim p_{z}}[\log (1-D(G(z)))]\label{eq:1}
\end{aligned}
\end{equation}

In Eq. \eqref{eq:1}, minimax is considered as a game in the context of GANs. Generally, the minimax is an optimization problem that aims to optimize the objective function using the given constraints of $G$ loss and $D$ loss. The use of the gradient descent method for an optimization of the objective function is discouraged as it may converge the function to a saddle point. At the saddle point, the objective function gives a minimal value for one model's weight parameters while the maximal value for the other model's weight parameters. Hence, the objective function is optimized using the minimax game to find a Nash equilibrium. 

\subsection{Variant of GANs}
In this section, we discuss the three most practiced variants of GANs that are proposed with some advancement in architecture and loss functions to the vanilla GAN to address the underlying training challenges.

\subsubsection{Deep Convolutional GAN (DCGAN)}
One of the popular variants of GANs is deep convolutional GAN (DCGAN) \cite{radford2015unsupervised}. The DCGAN adopted convolutional neural networks instead of fully connected networks as in vanilla GAN for the generator and the discriminator. Besides, batch normalization is used in most of the layers. The ADAM optimizer \cite{kingma2014adam} is adopted instead of SGD. DCGAN provides a meaningful solution in terms of a stable architecture as compared to a vanilla GAN. However, DCGAN lack in generating diverse, realistic, and free of artifacts images which are fundamental challenges that need more advanced solutions.

\subsubsection{Conditional GAN (CGAN)}
\begin{figure}[htp]
    \centering
    \includegraphics[width=1\textwidth]{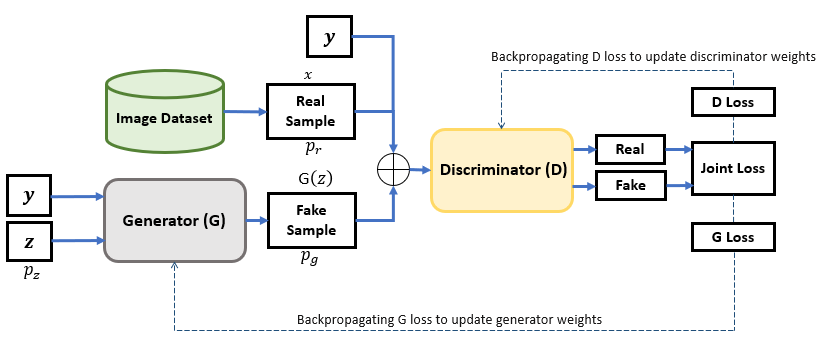}
    \caption{Architecture of CGANs. The generator $G$ and the discriminator $D$ are trained in an adversarial manner so that $G$ can generate plausible fake samples while $D$ can classify them from real samples. $y$ is a class label or any additional information conditioned with input samples for $G$ and $D$. $G$ loss is described as $log (1-D(G(z)))$ while $D$ loss is $log (D(x))$. The figure is redesigned from \cite{mirza2014conditional}.}
    \label{Fig.2}
\end{figure}

In vanilla GAN, the generator produces synthetic images only based on latent input $z$ which is considered to be limited information for high-performance image synthesis. Authors \cite{mirza2014conditional} proposed an idea of conditional GAN that utilizes additional information $y$ together with the random vector input $z$ as well as input to the discriminator. The $y$ can be a class label or any other conditional information that acts as an additional information feed to the generator as well as the discriminator. The CGAN architecture is presented in Fig. \ref{Fig.2}. The modified objective function is shown in Eq. \eqref{eq:2} that is reproduced from \cite{gui2020review}. The idea of CGAN has proven to be advantageous in terms of image synthesis as it can generate realistic and diverse images. CGAN shows a more stable training behavior as compared to vanilla GAN and DCGAN.

\begin{equation}
\begin{aligned}
\min _{G} & \max _{D} V(D, G)=E_{x \sim p_{\text {r}}}[\log D(x|y)] &+E_{z \sim p_{z}}[\log (1-D(G(z|y)))]\label{eq:2}
\end{aligned}
\end{equation}

\subsubsection{Wasserstein GAN (WGAN)}

To address the instability problem in vanilla GANs caused by the use of Jensen-Shannon divergence, authors in \cite{arjovsky2017wasserstein} proposed the idea of measuring the distance between two data distributions instead of minimizing the divergence. So, an Earth-mover (EM) or Wasserstein-1 distance is introduced in the Wasserstein-GAN (WGAN). The Wasserstein-1 distance is described as a metric instead of cross-entropy to measure the loss for optimizing the objective function. The objective function of the WGAN is shown in Eq. \eqref{eq: WGAN} that is reproduced from \cite{wang2021generative}.

\begin{equation}
W\left(p_{r}, p_{g}\right)=\inf _{\gamma \in \prod\left(p_{r}, p_{g}\right)} \mathbb{E}_{(\mathbf{x}, \mathbf{y}) \sim \gamma}\|\mathbf{x}-\mathbf{y}\|\label{eq: WGAN}
\end{equation}

In Eq. \eqref{eq: WGAN}, $\Pi\left(p_{r}, p_{g}\right)$ denotes all the joint distributions and $\gamma(\mathbf{x}, \mathbf{y})$ based on the marginals of $p_{r}$ and $p_{g}$. During the training of GAN, when there is no overlap between $p_{r}$ and $p_{g}$, the Jensen-Shannon divergence returns no values. However, the EM distance can reflect the distance measured continuously. Thus, WGAN can propagate meaningful gradient feedback to train the generator and avoid vanishing gradient problems. The main contribution of the WGAN is the use of a discriminator as a regressor instead of a binary classifier.

\subsubsection{StyleGAN}

\begin{figure}[htp!]
    \centering
    \includegraphics[width=1\textwidth]{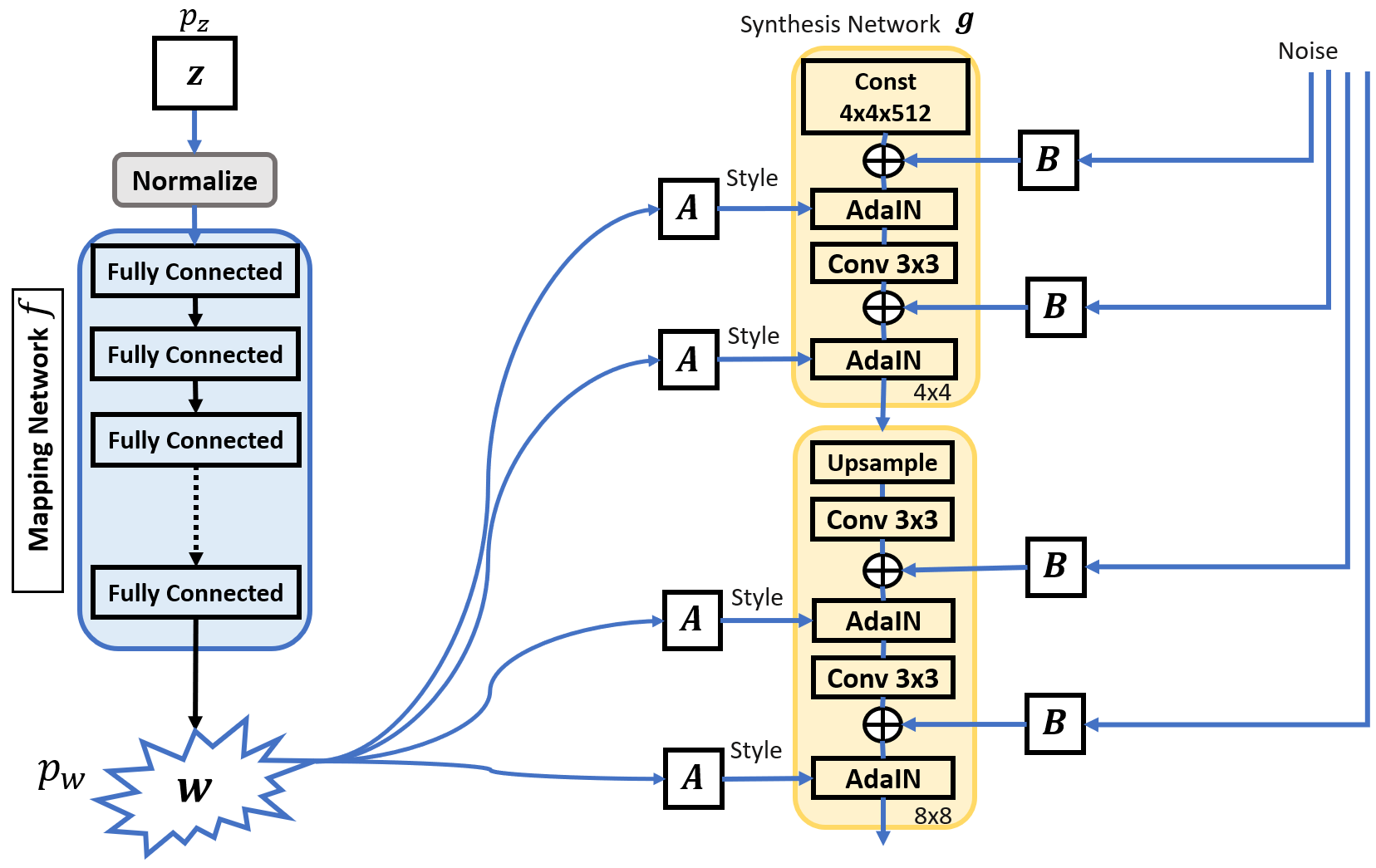}
    \caption{Generator architecture of StyleGAN. This figure is redesigned from \cite{karras2019style}.}
    \label{Fig:stylegan_arch}
\end{figure}

StyleGAN is a state-of-the-art GAN variant that was proposed with several key features to generate diversified and high-quality synthetic images \cite{karras2019style}. The architecture of StyleGAN is designed with a style generator, adaptive instance normalization (AdaIN), and a progressive growing training technique as depicted in Fig. \ref{Fig:stylegan_arch}. Unlike traditional GANs where the generator directly maps noise to images, StyleGAN separates the learned "style" (high-level features) from the learned "structure" (low-level features) of the image using a mapping network $f$. This separation allows for more control over the generation process and results in more realistic and appealing images. AdaIN is used to combine the learned style and structure information in StyleGAN. It aligns the statistics (mean and variance) of the intermediate feature maps to match the desired style \cite{wang2021generative}. The progressive growing training of StyleGAN starts with a low resolution and gradually increases the resolution of generated images during training. This approach helps stabilize the training process and allows the generator to focus on generating coarse details first before adding finer details, resulting in more coherent and realistic images.

StyleGAN is known for its ability to generate diverse and unique images from the same latent code. By controlling the style and structure separately, it allows for the manipulation of individual aspects of the generated image, such as changing the pose, color, and facial expressions while keeping the underlying structure consistent \cite{saxena2020generative}.

\subsubsection{CycleGAN}

\begin{figure}[htp!]
    \centering
    \includegraphics[width=1\textwidth]{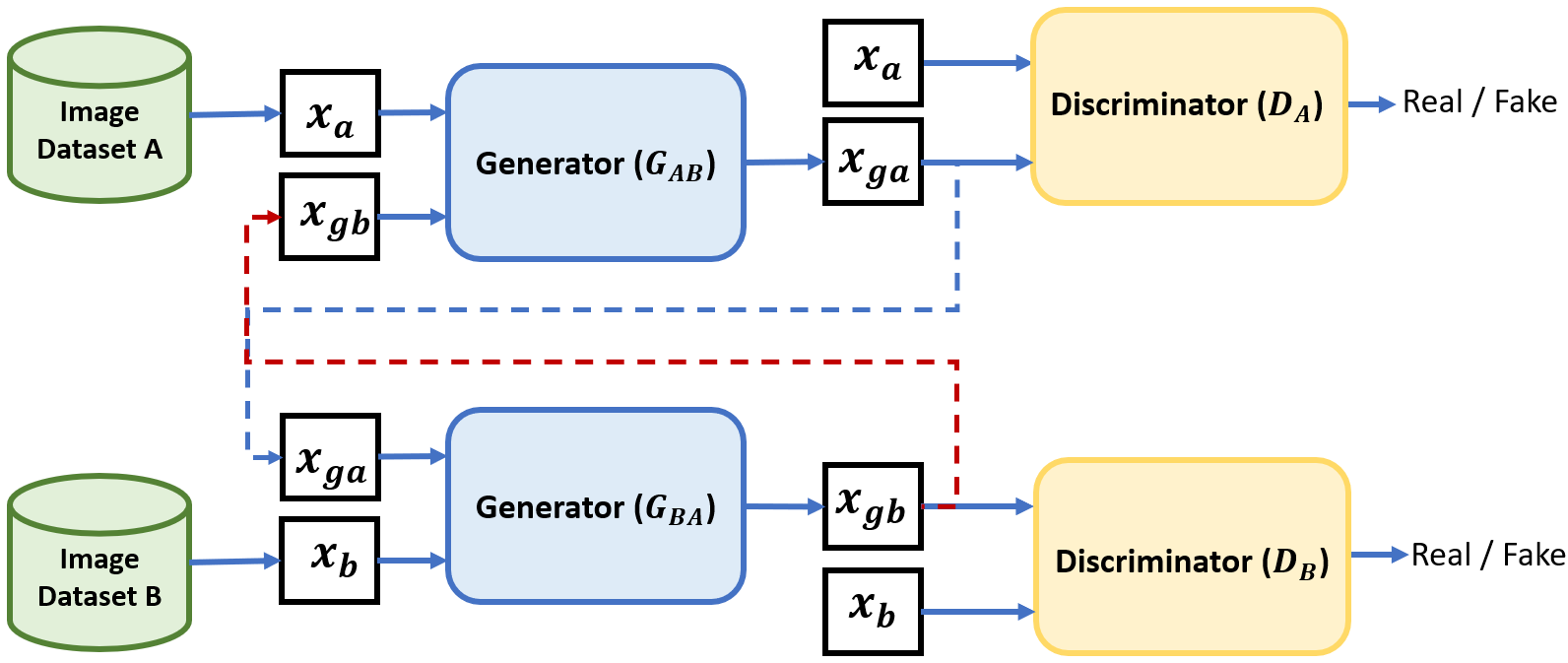}
    \caption{Architecture of CycleGAN. The generators $G_{AB}$ and $G_{BA}$ are trained in an adversarial manner by taking real samples from one domain as input and generating plausible fake image samples for another domain as output. $x_{a}$ and $x_{b}$ are two different unaligned image domains. The discriminators $D_{A}$ and $D_{B}$ distinguish the generated fake samples from real samples and provide feedback to the generators to update their learning accordingly. This figure is redesigned from \cite{singh2021medical}.}
    \label{Fig:cyclegan_arch}
\end{figure}

The variants of GANs, such as vanilla GAN, DCGAN, CGAN, and WGAN, are limited to the generation of a single image domain using latent input $z$. However, architectures of these GANs variants were designed to synthesize training images to similar domains and synthetic images have the same mapping as real training images.

The idea of generating images of different mappings and different modalities as compared to the real training images is known as image-to-image translation \cite{zhu2017unpaired}. For this purpose, CycleGAN architecture is proposed. The CycleGAN learns a mapping using the generators $G$: A $\xrightarrow{}$ B such as image distributions of A from $G(A)$ must be indistinguishable from the image distributions of B using an adversarial loss \cite{singh2021medical}. To this end, two generators and two discriminators with a cycle consistency loss are proposed in CycleGAN architecture as depicted in Fig. \ref{Fig:cyclegan_arch}. In CycleGAN, the generator $G_{AB}$ and discriminator $D_{B}$ work for a single pair using an adversarial loss $L_{\mathrm{GAN}}\left(G_{A B}, D_B\right)$ as defined in Eq. \ref{loss:firstpair}. However, the adversarial loss for reverse mapping pair $G_{BA}$ and $D_{A}$ is denoted as $L_{\mathrm{GAN}}\left(G_{B A}, D_A\right)$. So, a cycle-consistency loss is proposed to minimize the reconstruction error from image translation of one domain to another domain. The cycle-consistency loss is defined in Eq. \ref{loss:cycle-consistency}. The final loss of the CycleGAN is formulated as Eq. \ref{loss:finalCyclegan}.
\begin{equation}
L_{\mathrm{GAN}}\left(G_{A B}, D_B\right)=E_{b \sim P_B(b)}\left[\log D_B(b)\right]+E_{a \sim P_A(a)}\left[1-\log \left(D_B\left(G_{A B}(a)\right)\right)\right]\label{loss:firstpair}
\end{equation}
\begin{equation}
L_{\mathrm{cyc}}\left(G_{A B}, G_{B A}\right)=E_{a \sim P_A(a)}\left[a-G_{B A}\left(G_{A B}(a)\right) \|_1\right]+E_{b \sim P_B(b)}\left[b-G_{A B}\left(G_{B A}(b)\right) \|_1\right]\label{loss:cycle-consistency}
\end{equation}
\begin{equation}
L\left(G_{A B}, G_{B A}, D_A, D_B\right)=L_{\mathrm{GAN}}\left(G_{A B}, D_B\right)+L_{\mathrm{GAN}}\left(G_{B A}, D_A\right)+L_{\mathrm{cyc}}\left(G_{A B}, G_{B A}\right)\label{loss:finalCyclegan}
\end{equation}
\begin{equation}
G_{A B}^*, G_{B A}^*=\arg \min _{G_{A B}, G_{B A}} \max _{D_A, D_B} L\left(G_{A B}, G_{B A}, D_A, D_B\right)\label{loss:cycelganobjective funct}
\end{equation}
The CycleGAN is trained using the objective function defined in Eq. \ref{loss:cycelganobjective funct}. The equations \ref{loss:firstpair}, \ref{loss:cycle-consistency}, \ref{loss:finalCyclegan}, and \ref{loss:cycelganobjective funct} are reported from \cite{singh2021medical}.

\subsubsection{DiscoGAN}
\begin{figure}[htp!]
    \centering
    \includegraphics[width=1\textwidth]{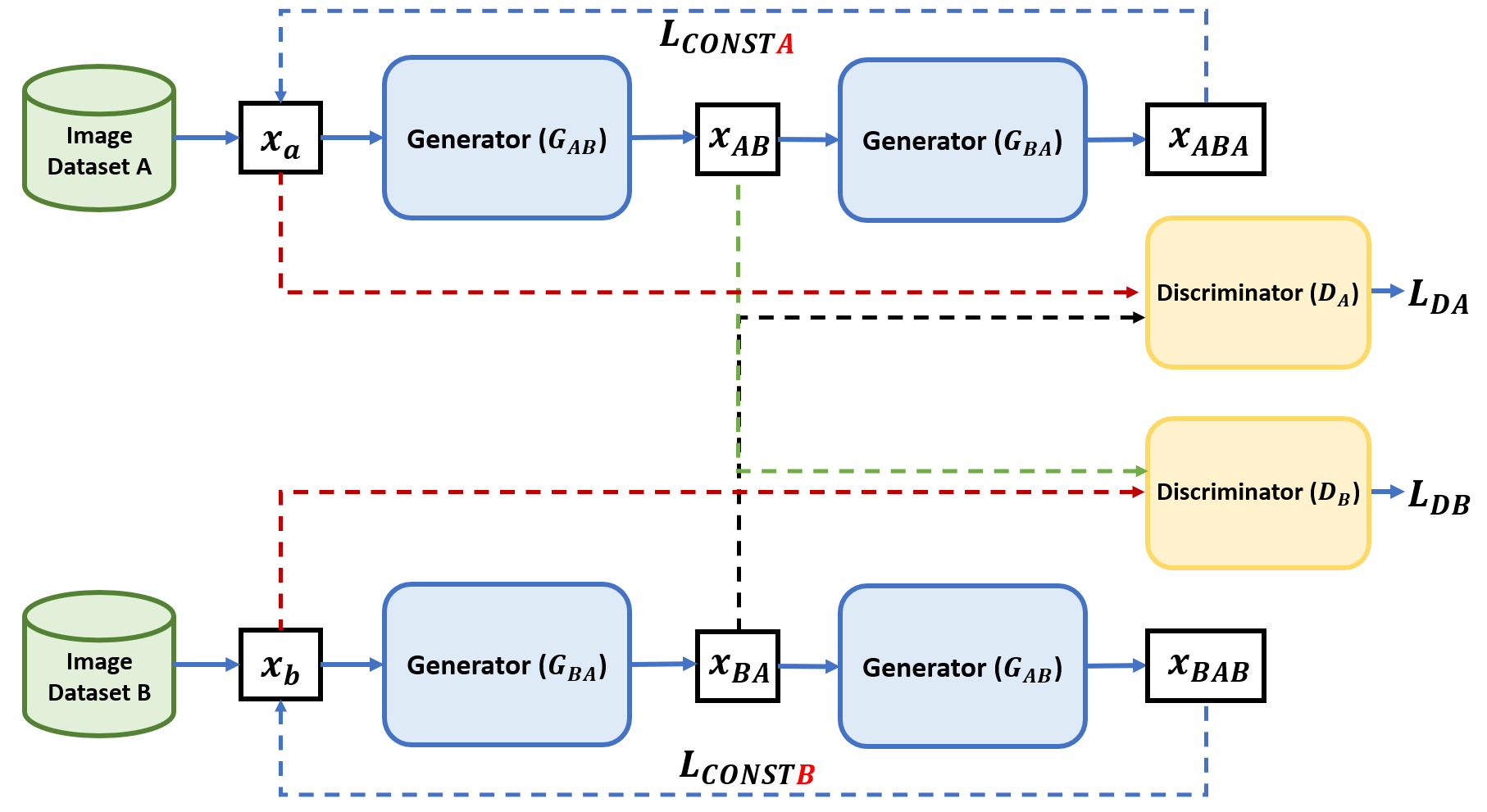}
    \caption{Architecture of DiscoGAN. The generator and discriminator models are designed with reconstruction losses ($L_{const}$) to discover the cross-domain relationship between two unpaired, unlabeled datasets. \cite{kim2017learning}.}
    \label{Fig:discogan_arch}
\end{figure}

DiscoGAN is another unsupervised GAN variant used for image-to-image translation tasks, but it focuses on discovering cross-domain relations between two distinct domains \cite{kim2017learning}. The main goal of DiscoGAN is to learn the cross-domain relationships between two unpaired datasets, without using any paired data during the training process. DiscoGAN uses reconstruction losses to discover relations among different domains as depicted in Fig. \ref{Fig:discogan_arch}. It aims to learn the shared structure between two domains, allowing for translation between the two domains in both directions.

\subsubsection{U-Net}
\begin{figure}[htp!]
    \centering
    \includegraphics[width=1\textwidth]{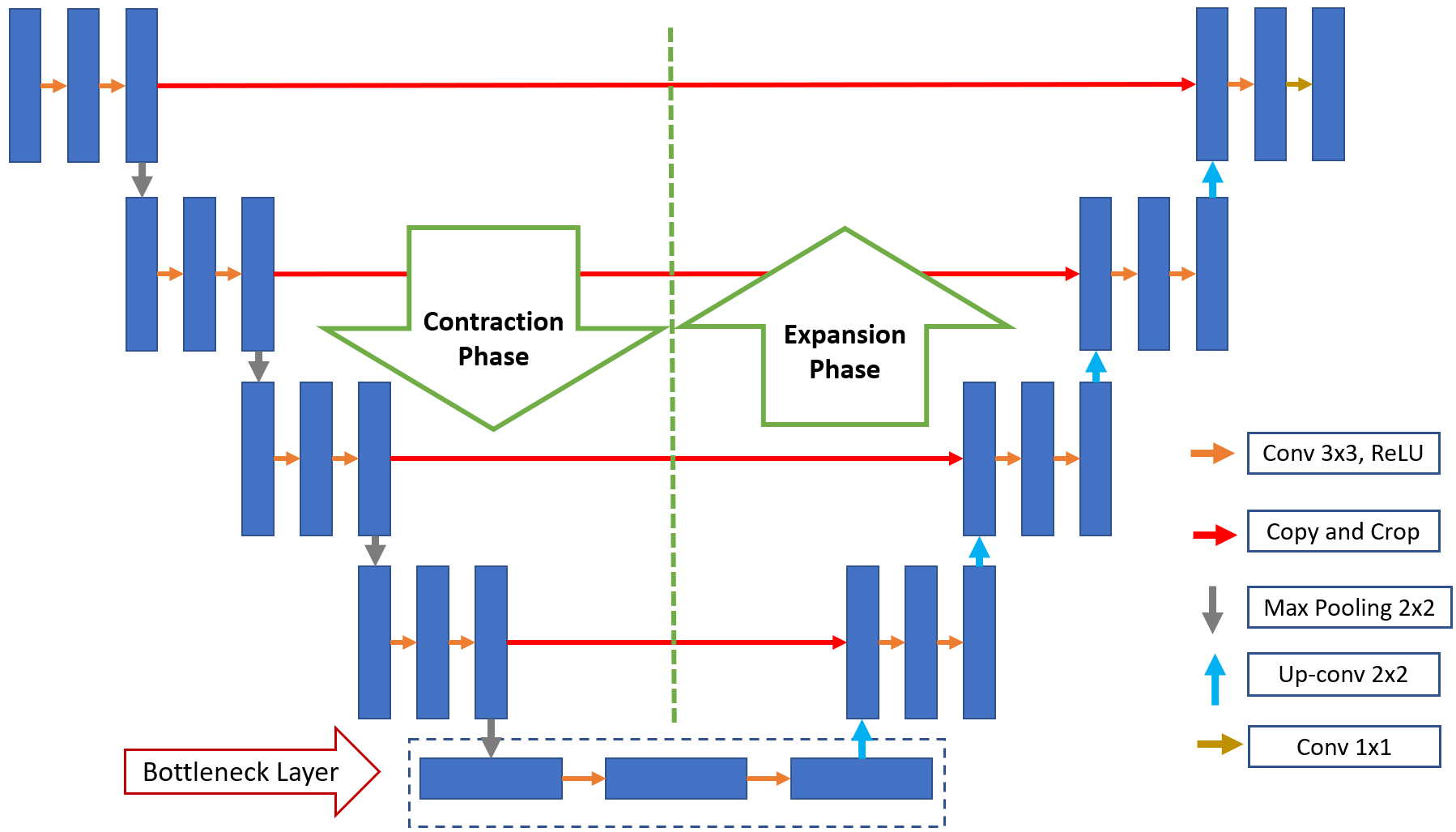}
    \caption{Architecture of U-Net.  
    \cite{ronneberger2015u}.}
    \label{Fig:unet_arch}
\end{figure}

The U-Net is a popular model that is widely used for image segmentation tasks in the domain of biomedical image analysis \cite{punn2022modality}. In GANs, U-Net is integrated into GAN architectures to perform segmentation tasks efficiently for biomedical images \cite{mubashar2022r2u++}.

U-Net is a U-shaped network that combines low-level and high-level information to extract the complex features of segmented regions. The U-Net is proposed by Ronneberger et al. \cite{ronneberger2015u}. The architecture of U-Net is depicted in Fig. \ref{Fig:unet_arch}. U-Net is designed with a symmetrical ordering of encoder-decoder blocks to distinguish every pixel by extracting multi-scale feature maps using encoding the input and decoding it to output using the same resolution \cite{punn2022modality}. The U-Net is operated to segregate the overlapping regions using background pixels with an individual loss of each pixel. This process is defined through an energy function $E$ as represented in Eq. \ref{energyfunction}. 
\begin{equation}
E=\sum_{x \in \Omega}\left(w_c(x)+w_0 \cdot \exp \left(-\frac{\left(d_1(x)+d_2(x)\right)^2}{2 \sigma^2}\right)\right) \log \left(p_{\ell(x)}(x)\right)\label{energyfunction}
\end{equation}
In Eq. \ref{energyfunction} reported from \cite{punn2022modality}, softmax is defined as Eq. \ref{unetsoftmax}.
\begin{equation}
p_k(x)=\exp \left(a_k(x)\right) /\left(\sum_{k^{\prime}=1}^K \exp \left(a_{k^{\prime}}(x)\right)\right)\label{unetsoftmax}
\end{equation}
In \ref{energyfunction} and \ref{unetsoftmax}, The $w_c$ indicates a weight map while $d_1$ and $d_2$ denote the distances to the boundary pixels at the first and second nearest positions respectively. $w_0$ and $\Omega$ are the constants. The $a_k(x)$ denotes an activation for channel k with pixel $x \in \Omega$ and $\Omega \in \mathbb{Z}^2$.

\section{Applications of GANs in Biomedical Image Analysis}
\label{section:application_of_GANs}
\begin{figure}
    \centering
\begin{center}
        \resizebox{1\textwidth}{!}{%
            \begin{forest}
                for tree={
                    forked edges,
                    draw,
                    rounded corners,
                    node options={align=center},
                    text width=2.7cm,
                    anchor=center,
                 },
                where level=0{%
      }{%
        folder,
        grow'=0,
        if level=1{%
          before typesetting nodes={child anchor=north},
          edge path'={(!u.parent anchor) -- ++(0,7pt) -| (.child anchor)},
        }{},
        }
                [Applications of GANs in Biomedical Image Analysis, rotate = 0, fill=gray!25, parent
                [Image \\ Reconstruction \\ \cite{yedder2020deep} (2020), rotate = 0, for tree={fill=green!25, child}]
                [Image \\ Segmentation \\ \cite{yi2019generative} (2019) \\ \cite{nalepa2019data} (2019) \\ \cite{roman2020medical} (2020), rotate = 0, for tree={fill=teal!25,child}]
                [Image \\ Denoising \\ \cite{kazeminia2020gans} (2020) \\ \cite{denoising2020} (2020), rotate = 0, for tree={fill=brown!25, child}]
                [Image Synthesis, rotate = 0, for tree={fill=lime!25, child}, calign with current edge
                [Conditional \\ Image \\ Synthesis \\ \cite{singh2021medical} (2021), rotate = 0]
                [Unconditional \\ Image \\ Synthesis \\ \cite{yi2019generative} (2019) \\ \cite{kazeminia2020gans} (2020), rotate = 0]]
                [Image \\ Detection \\ \cite{yi2019generative} (2019) \\ \cite{kazeminia2020gans} (2020), rotate = 0, for tree={fill=yellow!25, child}]
                [Image \\ Registration \cite{haskins2020deep} (2020), rotate = 0, for tree={fill=purple!25, child}]
                [Image Super \\ Resolution \cite{li2020review} (2020), rotate = 0, for tree={fill=blue!25, child}]]
            \end{forest}
        }
    \end{center}
    \caption{Applications of GANs in biomedical image analysis.}
    \label{fig: appplications GANs training}
    \end{figure}
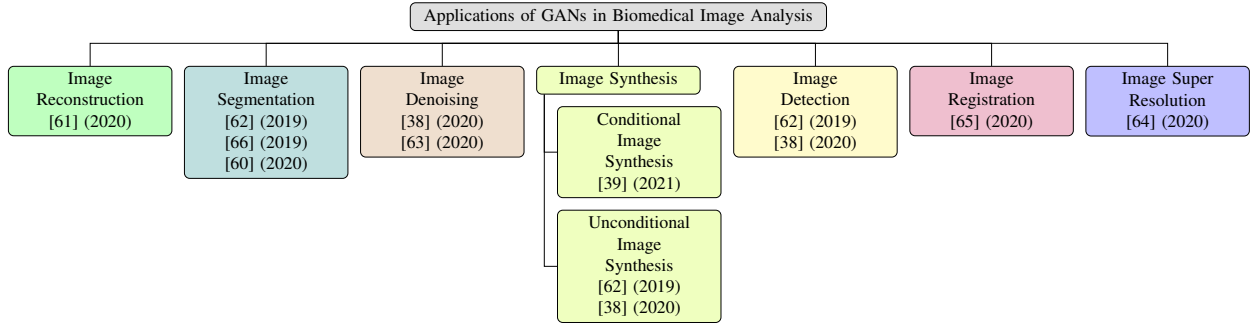

In the domain of biomedical imaging, GANs have been utilized in several applications such as image synthesis \cite{kazeminia2020gans}, image segmentation \cite{roman2020medical}, image reconstruction \cite{yedder2020deep}, image detection \cite{yi2019generative}, image denoising \cite{denoising2020}, image super-resolution \cite{li2020review}, and image registration \cite{haskins2020deep}. The performance of these applications is affected by the training challenges of GANs. This section presents a high-level discussion on the impact of training challenges of GANs for the applications such as image synthesis, image segmentation, image reconstruction, image detection, image denoising, image super-resolution, and image registration in biomedical image analysis. How these training challenges affect applications is also discussed. A few state-of-the-art survey papers are identified to get insights into these applications for readers that are shown in Fig. \ref{fig: appplications GANs training}.

\subsection{Image synthesis}
GANs are used to generate synthetic images of training images. Conventionally, GANs are introduced as unsupervised models and can be leveraged with unannotated image datasets. Therefore, synthesizing training images using GANs is known as image synthesis. Training challenges of GANs can affect the synthetic images during the image synthesis process. For example, the generation of similar synthetic images for distinct input images, blurry images, and low-quality images indicates the training challenges of GANs. GANs have been used for two types of image synthesis; unconditional image synthesis and conditional image synthesis \cite{singh2021medical} \cite{kazeminia2020gans} \cite{yi2019generative}. Each type of image synthesis is discussed as follows;

\subsubsection{Unconditional Image Synthesis}
In unconditional image synthesis, GANs rely only on random noisy inputs in the latent space without any prior conditions to generate new synthetic image samples. The unconditional image synthesis of biomedical images is affected most by the training challenges of GANs such as mode collapse and training instability. For example, direct generation of magnetic resonance images, computed tomography images, cell images, and dermoscopic images encounter these training challenges. Being an unsupervised framework, this approach has been widely utilized for biomedical image analysis to address data limitation and class imbalance issues. A detailed discussion and technical papers can be found in \cite{kazeminia2020gans} \cite{yi2019generative}.

\subsubsection{Conditional Image Synthesis}
In conditional image synthesis, GANs consider some prior conditional information together with $z$ to generate new synthetic images. This type of image synthesis face training challenges of GANs during the image-to-image translation tasks. When a GAN generates a biomedical image from the same modality input or cross-modality input images, it can miss salient features of input images during the training to translate into new images. Due to instability problems, the quality of synthetic images can be affected during the generation of biomedical images. There are two types of applications in conditional image synthesis. Generation of new images from real images with some prior conditions in the same modalities such as CT to CT, MRI to MRI, and PET to PET. Generation of new images from different modalities like MRI to CT, MRI to PET, etc. The survey paper \cite{singh2021medical} discussed these applications in detail and can be studied. 

\subsection{Image Segmentation}
GANs provide a significant contribution to the domain of biomedical imagery for image segmentation tasks. It has been utilized for the segmentation of tumors, pathology, and lesions from different body parts like the brain or liver, etc. GANs use segmented masks with input images to generate synthetic images with the segmentation of the target masks. Sometimes, during the training of GANs, the segmented masks are difficult to learn and GANs generate poorly segmented synthetic images or low-quality images. The literature \cite{yi2019generative} \cite{nalepa2019data} \cite{roman2020medical} \cite{iqbal2022generative} can be explored for more discussion on biomedical image segmentation.

\subsection{Image Reconstruction}
GANs have been utilized to improve the quality of reconstructed images like estimating full-dose CT images from low-dose CT images with reduced aliasing artifacts. Usually, GANs do not reduce these aliasing artifacts effectively due to the training instability problem. GANs face difficulty to generate plausible images reconstructed from training images due to poor image quality. The mode collapse occurs during the training of GANs while learning the distribution of low-quality images. The reader can be referred to the survey paper \cite{yedder2020deep} for a detailed insight on biomedical image reconstruction using GANs.

\subsection{Image Detection}
GANs have been used for unsupervised anomaly detection in biomedical imagery. The discriminator model can be used to detect anomalies like lesions or tumors. This contribution helps to work with unannotated data and address the problem of anomaly detection. The survey papers \cite{kazeminia2020gans} \cite{yi2019generative} are identified for more detail on the underlying GANs application.

\subsection{Image Denoising}
Image denoising techniques are required to remove the noise and recover the original latent information from the noisy images. GANs can be used as an excellent tool to produce sharp, plausible, and noise-free images. A powerful GAN model is required to denoise biomedical images because it is usually incorporated with the training challenges of GANs. These challenges can affect the denoising of biomedical images as GANs are unable to learn low-quality or noisy images effectively and can reflect a poor generation of output images. A more detailed overview of biomedical image denoising techniques with the utility of GANs can be studied in the literature \cite{kazeminia2020gans} \cite{denoising2020}.

\subsection{Image Super Resolution}
GANs can be utilized to produce super-resolution images from low-resolution images. The training instability problem should be addressed completely to achieve better high-resolution biomedical images as the optimality of the GANs is difficult to achieve. The mode collapse and non-convergence problems can also degrade the quality of synthetic images. GANs have performed various super-resolution tasks in biomedical image analysis and the reader can find a detailed review of those tasks in the review paper \cite{li2020review}.

\subsection{Image Registration}
Conventional registration techniques suffer from parameter dependency problems and high optimization loads. GANs have good capabilities of image transformations that can serve as excellent candidates for the extraction of a more optimal registration mapping. GANs have limitations of training challenges as they can miss the location of an object or feature in the biomedical image during the image registration process. Usually, 3D volumes of biomedical images face these challenges as the generator can not learn 3D volumes effectively to generate diverse, un-blurred, and high-quality synthetic images. More details can be found in the survey paper of \cite{haskins2020deep}.

\section{Evaluation Metrics}
\label{eval_metrics}
Several evaluation metrics have been proposed to assess the technical training challenges of GANs, such as mode collapse, non-convergence, and unstable training. These metrics include Inception Score (IS), Maximum Mean Discrepancy (MMD), Multi-scale Structural Similarity Index Measure (MS-SSIM), Fr\'echet Inception Distance (FID), Peak signal-to-noise ratio (PSNR), Dice Score (DS), and classification performance metrics (Precision and Recall). Each metric is discussed in detail as follows:
\subsection{Inception Score (IS)}
\label{IS_eval}
Inception Score is a metric used for the evaluation of GANs \cite{salimans2016improved}. It provides an assessment of generated images for high-quality and diverse characteristics. IS utilizes a pre-trained Inception-Net \cite{szegedy2016rethinking} and measures the KL divergence between class conditional probability distribution $p(y \mid \mathbf{x})$ of generated sample and the marginal probability distribution $p(y)$ obtained from a set of generated images.

\begin{equation}
\exp \left(\mathbb{E}_{\mathbf{x}}[\mathbb{K} \mathbf{L}(p(\mathbf{y} \mid \mathbf{x}) \| p(\mathrm{y}))]\right)=\exp \left(H(y)-\mathbb{E}_{\mathbf{x}}[H(y \mid \mathbf{x})]\right)\label{Eq. IS}
\end{equation}

In Eq. \eqref{Eq. IS} that is reproduced from \cite{borji2019pros}, $p(y \mid \mathbf{x})$ shows the class conditional probability distribution with image x, $p(y)$ is a marginal probability distribution, and $H(x)$ denotes the entropy of variable x. \cite{borji2019pros}. IS measures the lowest score as 1 while the highest score depends on the number of classes of the dataset. The higher IS score shows that the model can generate high-quality as well as diverse images.

\subsection{Maximum Mean Discrepancy (MMD)}
\label{MMD_eval}
The maximum mean discrepancy is used to measure the dissimilarity between real image distribution $p_{r}$ and generated image distribution $p_{g}$ \cite{mmd}. The higher value of MMD indicates that the generator is collapsing and doesn't generate realistic and diverse images.

\begin{equation}
\operatorname{MMD}(Pr, Pg)=\left\|\mu_{R}-\mu_{G}\right\|_{\mathcal{H}}^{2}\label{Eq. mmd}
\end{equation}

Mathematically, it uses Hilbert's space of functions. In Hilbert space functions, two functions are supposed to be point-wise closed if they are closed in the norm \cite{segato2020data}. So, MMD can be calculated by measuring the squared distance between the embeddings of $p_{r}$ and $p_{g}$ as shown in Eq. \eqref{Eq. mmd} that is reproduced from \cite{borji2019pros}.

\subsection{Multi-scale Structural Similarity Index Measure (MS-SSIM)}
\label{MSSSIM_eval}
MS-SSIM is a metric that is used to assess the diversity of synthetic images in GANs. MS-SSIM is introduced to measure the similarity score using human perception similarity analysis. It computes the similarity between two images with the help of pixels and structures \cite{odena2017conditional}. MS-SSIM considers luminance (realizing the brightness of a color) and contrast estimations for a metric score. Luminance ($l$), contrast ($c$), and structure ($s$) can be computed using Eq. \eqref{Eq. cont} as reproduced from \cite{borji2019pros}.

\begin{equation}
I(x, y)=\frac{2 \mu_{x} \mu_{y}+C_{1}}{\mu_{x}^{2}+\mu_{y}^{2}+C_{1}} \quad C(x, y)=\frac{2 \sigma_{x} \sigma_{y}+C_{2}}{\sigma_{x}^{2}+\sigma_{y}^{2}+C_{2}} \quad S(x, y)=\frac{\sigma_{x y}+C_{3}}{\sigma_{x} \sigma_{y}+C_{3}}\label{Eq. cont}
\end{equation}

In Eq. \eqref{Eq. cont}, $x$ and $y$ are two images. $\mu_{x}$ and $\mu_{y}$ represent the mean, whereas $\sigma_{x}$ and $\sigma_{y}$ denote the variance (standard deviation) of pixel intensities. The correlation between corresponding pixels is represented by $\sigma_{xy}$. For the numerical stability of the fractions, constant C is added in all three quantities. The single-scale similarity index is then computed by Eq. \eqref{Eq.ssim} (reproduced from \cite{borji2019pros}) by considering the fixed distance perspective, as well as sampling density of images \cite{wang2004image}.

\begin{equation}
\operatorname{SSIM}(x, y)=I(x, y)^{\alpha} C(x, y)^{\beta} S(x, y)^{\gamma}\label{Eq.ssim}
\end{equation}

The multi-scale SSIM is a variant of the single-scale SSIM metric. It considers all scales of iteratively downsampled images for computing contrast and structural scores. The luminance quantity is measured at the last iteration known as the coarsest scale (M). Conversely, it gives weightage to the contrast and structure at each scale. The MS-SSIM is computed by Eq. \eqref{Eq. ms-ssim} as reproduced from \cite{borji2019pros}.

\begin{equation}
\operatorname{MS}-\operatorname{SSIM}(x, y)=I_{M}(x, y)^{\alpha_{M}} \prod_{j=1}^{M} C_{j}(x, y)^{\beta_{j}} S_{j}(x, y)^{\gamma_{j}}\label{Eq. ms-ssim}
\end{equation}

The range of MS-SSIM scores lies between 0.0 and 1.0. An important point to note is that a higher MS-SSIM score shows lower diversity between images of the same class. This metric is useful for evaluating GANs to compute the diversity between generated images of a single class.

\subsection{Fr\'echet Inception Distance (FID)}
\label{FID_eval}
FID is an evaluation metric used to assess the quality of synthetic images. It is proposed by Heusel et al \cite{heusel2017gans}. FID computes the mean and covariance of synthetic and real images as shown in Eq.\eqref{FID} that is reproduced from \cite{borji2019pros}. It visualizes an embedded layer that contains a set of synthetic images in the Inception-Net and uses it as the continuous multivariate Gaussian.

\begin{equation}
F I D(r, s)=\left\|\mu_{r}-\mu_{s}\right\|_{2}^{2}+\operatorname{Tr}\left(\Sigma_{r}+\Sigma_{s}-2\left(\Sigma_{r} \Sigma_{s}\right)^{\frac{1}{2}}\right)\label{FID}
\end{equation}

In Eq.\eqref{FID}, r and s shows real and synthetic images while $\left(\mu_{r}, \Sigma_{r}\right)$ and $\left(\mu_{s}, \Sigma_{s}\right)$ denote mean and covariances of real and synthetic images. FID score measures the distance between real and synthetic images in GANs. A higher FID score shows a larger distance between synthetic and real data distributions \cite{borji2019pros}.

\subsection{Peak signal-to-noise ratio (PSNR)}
\label{PSNR_eval}
In GANs, PSNR is used to check the quality of synthetic images to the corresponding real images. PSNR is applied to monochrome images. It is measured in decibels (dB). The higher value of PSNR represents a better quality of synthetic images. PSNR is computed as shown in Eq. \eqref{psnr} reproduced from \cite{borji2019pros}.   

\begin{equation}
\operatorname{PSNR}(I, K)=10 \log _{10}\left(\frac{M A X_{I}^{2}}{M S E}\right)\label{psnr}
\end{equation}

By simplifying,

\begin{equation}
\operatorname{PSNR}(I, K)=20 \log _{10}\left(M A X_{I}\right)-20 \log _{10}\left(M S E_{I, K}\right)\label{max}
\end{equation}

Whereas

\begin{equation}
M S E_{I, K}=\frac{1}{m n} \sum_{i=0}^{m-1} \sum_{i=0}^{n-1}(I(m, n)-K(m, n))^{2}\label{mse}
\end{equation}

The Eq.\eqref{psnr}, \eqref{max}, and \eqref{mse} are reported in \cite{borji2019pros}. I and K represent two monochrome images. In Eq. \eqref{max}, MAXI denotes the highest possible pixel value of an image such as 255 in the case of 8-bit representation.

\subsection{Dice Score (DS)}
Dice score is a popular metric that is used to evaluate the targeted segmented images as compared to their real ground truth images \cite{bertels2019optimizing}. In GANs, DS is also utilized to assess the quality of synthetic segmented images. DS compares the area of segmented regions of the generated synthetic images and real ground truth images with the total area of both regions \cite{ghaffari2019automated}. The formula for DS is calculated using Eq. \ref{dice_score}:  
\begin{equation} DS = 2 \times \frac{{{Y_{true}} \times {Y_{pred}}}}{{{Y_{true}} + {Y_{pred}} + \varepsilon }}\label{dice_score} 
\end{equation}
In Eq. \ref{dice_score} reported in \cite{ghaffari2019automated}, $Y_{pred}$ indicates the ground truth, $Y_{pred}$ indicates the predicting label and $\varepsilon$ is a small number used for avoiding division by zero. Perfect segmentation is indicated by the DS of 1.0.

\subsection{Classification Performance Metrics (Precision and Recall)}

In GANs, classification metrics such as recall and precision are also used to evaluate the quality and diversity of synthetic images \cite{borji2019pros}. In literature, studies have been proposed to measure the recall and precision to quantify the mode collapse and instability problem \cite{lucic2018gans} \cite{sajjadi2018assessing}. In \cite{lucic2018gans}, authors argued that these classification metrics can evaluate the quality and diversity of synthetic images. Sajjadi et al., \cite{sajjadi2018assessing} argued that high precision and low recall scores indicate low quality and diversity of synthetic images while higher quality and diversity of synthetic images are indicated by low precision and high recall scores.

\begin{figure}[ht]
    \centering
    \includegraphics[width=1\textwidth]{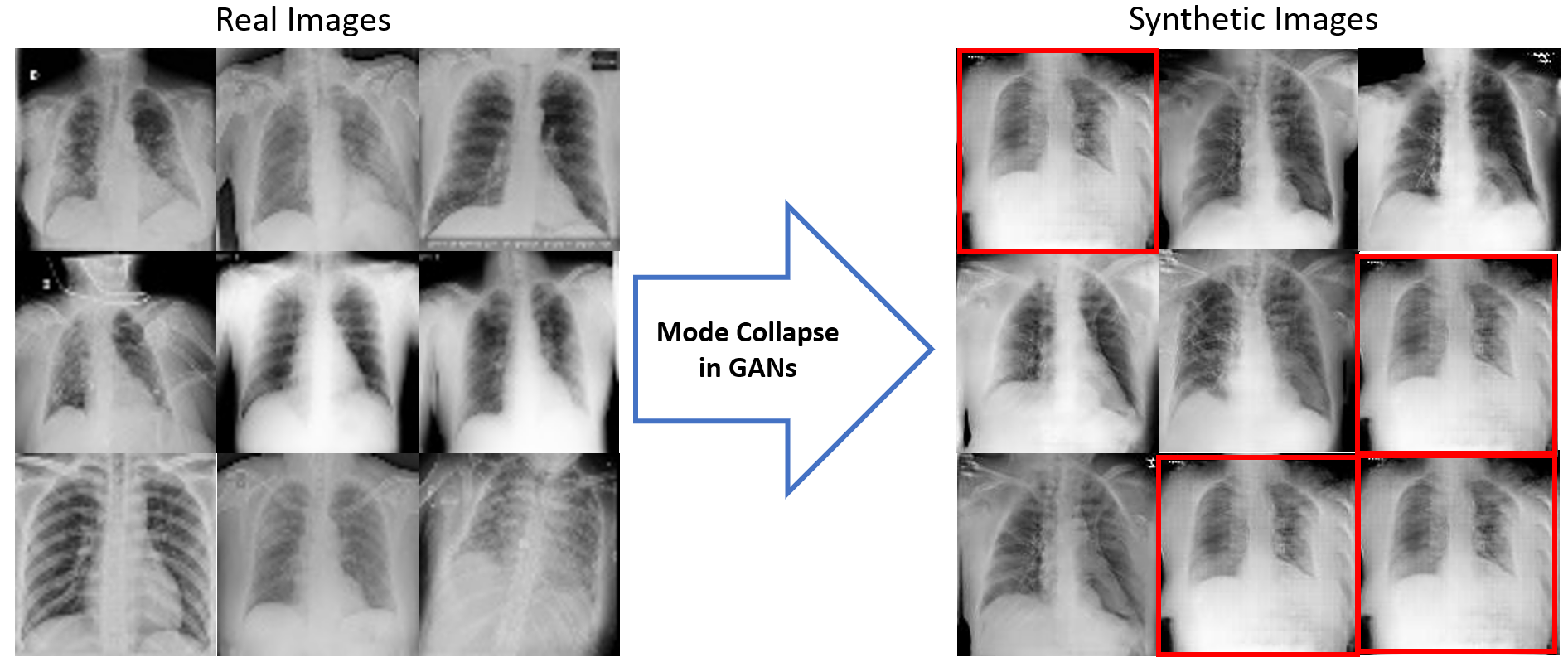}
    \caption{Identification of mode collapse in GANs for X-ray image synthesis. The red areas highlighted illustrate the repetition of synthetic X-ray images with a similar distribution of features such as lungs. The chest bones are also suppressed indicating the occurrence of a mode collapse problem in GANs.}
    \label{Fig.mode_collapse}
\end{figure}

\begin{figure}[ht]
    \centering
    \includegraphics[width=1\textwidth]{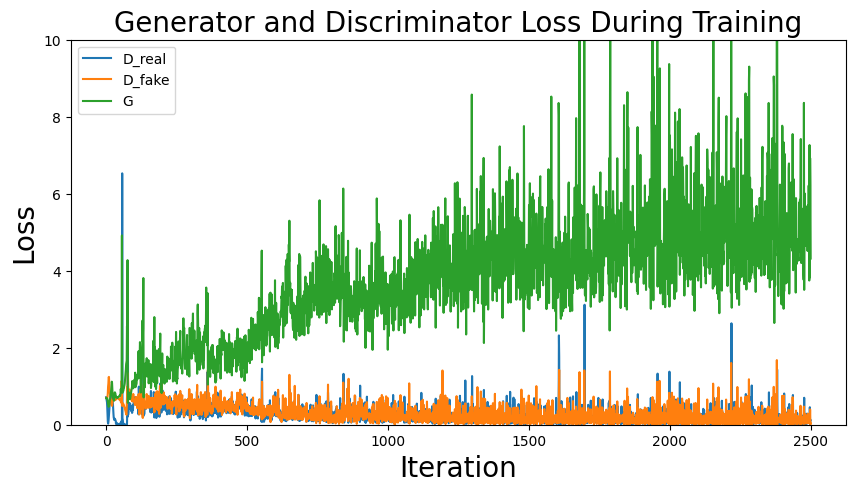}
    \caption{Identification of the mode collapse problem using the non-converging generator loss of GANs for X-ray image synthesis. The generator loss depicted by label ($G$) illustrates the non-converging behavior as compared to the discriminator losses ($D\_real$ and $D\_fake$).}
    \label{Fig.mode_collapsed_generator}
\end{figure}

\begin{figure}[ht]
    \centering
    \includegraphics[width=1\textwidth]{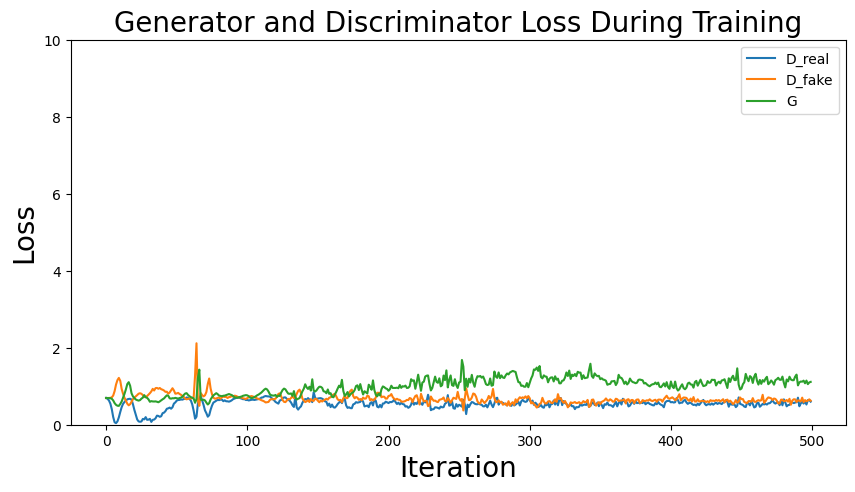}
    \caption{Identification of the no-mode collapse in GANs for X-ray image synthesis. The generator loss depicted by label ($G$) illustrates the converging and balanced behavior as compared to the discriminator losses ($D\_real$ and $D\_fake$).}
    \label{Fig.converging_generator}
\end{figure}

\section{The Mode Collapse Problem}
\label{section:modecollapse}

\subsection{Definition}
\label{section: mode collapse: Def}

The basic purpose of the GANs is to produce realistic and a variety of synthetic output images. The synthetic images should be of different styles (modes of distribution) for each random input. In practice, the generator learns to produce synthetic images just to misguide the discriminator for being classified as real. Once the generator finds the best way to fool the discriminator by producing particular plausible images, it focuses on the generation of similar images repetitively. The discriminator gets fooled each time and classifies the synthetic images as real. Eventually, the discriminator gets stuck in this trap and is unable to get out of this trap. Consequently, the generator starts producing a similar style of images. The underlying problem is known as mode collapse \cite{goodfellow2016nips}. 

\subsection{Identification}
\label{section: mode collapse: Ident}
The mode collapse problem is identified during the training of GANs by looking at the nature of generated images. The mode collapse refers to the generation of less diversified synthetic images where salient features of input (real) images are overlooked by the generator during the training of GANs \cite{saad2022addressing}. Therefore, GANs with mode collapse generate synthetic images with similar distribution modes repetitively rather than having input images with diverse distribution modes as indicated in Fig. \ref{Fig.mode_collapse}. The mode collapse problem can be divided into two categories based on the number of classes within the datasets \cite{alot2020}. Firstly, when the generator produces a similar style of output images for multi-class input images then it will affect the inter-class diversity, and the problem is known as inter-class mode collapse. Secondly, when the generator produces a similar style of output images for single-class input images then the problem is termed as intra-class mode collapse and affects the intra-class diversity. The mode collapse problem can also be identified using the loss curves of the generator during the training of GANs. Figure \ref{Fig.mode_collapsed_generator} illustrates the mode collapse during the training of GANs using a non-converging generator loss (G) for X-ray image synthesis. Consequently, a converging generator loss (G) in Fig. \ref{Fig.converging_generator} shows the balanced training of GANs indicating no mode collapse for X-ray image synthesis.

\subsection{Quantification}
\label{section: mode collapse: Quant}
The diversity and similarity of generated synthetic images can be computed by several evaluation metrics. The occurrence of mode collapse and diversity of synthetic images is quantified by MS-SSIM \cite{wang2003multiscale} \cite{odena2017conditional} using image similarity features while IS \cite{salimans_training}, MMD \cite{mmd}, and FID \cite{heusel2017gans} using distance measures as discussed in Section \ref{eval_metrics}. However, PSNR, SSIM, and classification metrics such as recall and precision are also used to quantify the diversity of synthetic images.
\begin{figure}[htp!]
    \centering
    \includegraphics[width=1\textwidth]{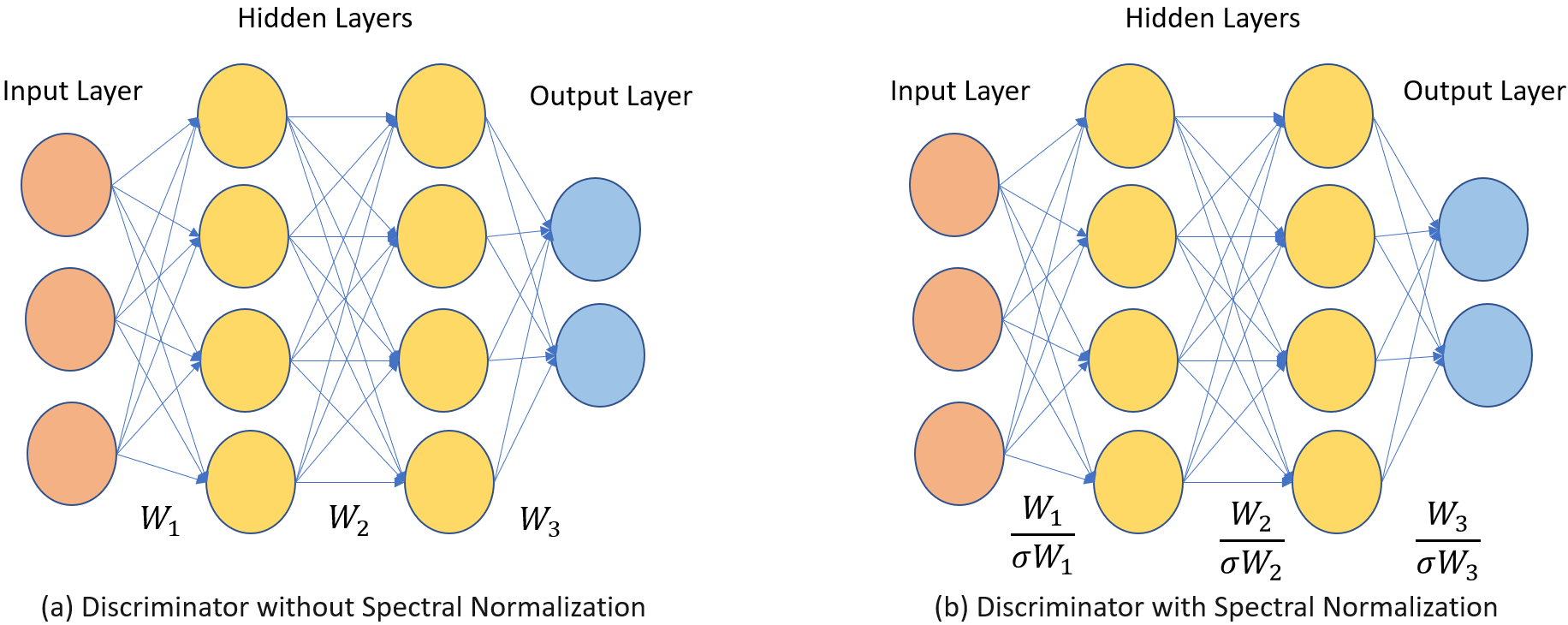}
    \caption{Spectral normalization utilizes the largest singular values of $W_i$ as their spectral norms ($\sigma$$W_i$) to divide the actual gradient weights of the discriminator. The figure is redesigned from \cite{miyato2018spectraliclr}.}
    \label{Fig.spectralnorm}
\end{figure}

\begin{figure}[htp!]
    \centering
    \includegraphics[width=1\textwidth]{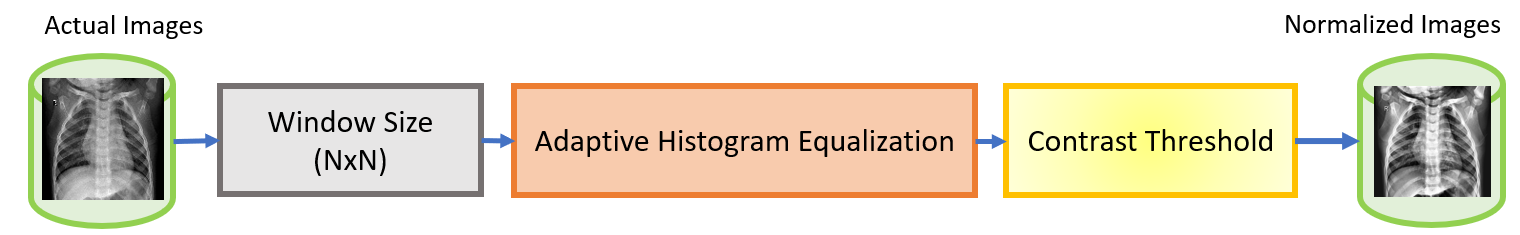}
    \caption{The block diagram of an adaptive input-image normalization. The figure is redesigned from \cite{saad2022addressing}.}
    \label{Fig.AIINDCGAN}
\end{figure}
\subsection{Solutions to the Problem}
\label{section: mode collapse: Soltn}
\subsubsection{Regularization}
In deep learning models, we aim to find minimum loss that is difficult to achieve when using large weight sizes. This will lead the model to overfit the data and provide poor prediction results. To alleviate this problem, a regularization term is used to reduce the weight size of the network or limit the model capacity \cite{goodfellow2016deep}. In GANs, neural networks are used in the generator as well as in the discriminator. So, when the discriminator produces ambiguous gradients as feedback to the generator continuously, the generator learns to generate similar images again and again to fool the discriminator which leads to the mode collapse problem. Here, regularization is used as weight normalization. 

\paragraph{Weight Normalization (WN):}
In GANs, weight normalization (WN) uses specialized training algorithms to update the weight matrices regularly while training the GANs. WN does not use additional loss. It backpropagates the gradients by computing them according to the normalized weights during the training of GANs \cite{lee2020regularization}. Several normalization techniques such as spectral normalization \cite{miyato2018spectraliclr}, batch normalization \cite{radford2015unsupervised}, and self-normalization \cite{klambauer2017self} have been proposed to use as a weight normalization in GANs.

Xu et al. \cite{xu2020low} alleviated the mode collapse problem in a GAN using spectral normalization for a super-resolution of low-dose X-ray images. Spectral Normalization is a type of weight normalization that employs the spectral norm of weight matrices as shown in Fig. \ref{Fig.spectralnorm} while training GANs. The spectral norm is equivalent to the L2 norm and corresponds to the largest singular vector. The largest singular vector can be approached to the Lipschitz constant. The spectral normalization is used to normalize the weight matrices in the discriminator of the proposed Spectral Normalization Super Resolution GAN (SNSRGAN) which controls the Lipschitz constant to 1. The authors utilized IS and MS-SSIM scores to evaluate the diversity of super-resolution synthetic images generated by the SNSRGAN. Results demonstrate that SNSRGAN achieved improved scores of IS with 6.56 and MS-SSIM with 0.986 as compared to the baseline SRGAN \cite{ledig2017}.

\paragraph{Input Normalization (IN)}
Input normalization refers to the normalization of input image features so that a GAN can better train on those normalized images and alleviate the mode collapse problem for biomedical image synthesis.

A similar idea of input image normalization is proposed by Saad et al. \cite{saad2022addressing} to the DCGAN for generating diversified chest X-ray images. The authors alleviated the mode collapse in the DCGAN using a preprocessing technique namely an adaptive input-image normalization (AIIN). The AIIN normalizes the input X-ray images using a contrast-based histogram equalization to highlight the diverse features of X-ray images as depicted in Fig. \ref{Fig.AIINDCGAN}. A DCGAN learns X-ray image features more accurately with these normalized images having highlighted features and can generate improved diversified X-ray images. Several experiments with varying batch sizes, window sizes, and contrast thresholds have been conducted. They used MS-SSIM and FID evaluation metrics to evaluate the mode collapse problem in DCGAN and the diversity of synthetic images.

The authors demonstrated improved results of AIIN-DCGAN over DCGAN with high diversity scores using the MS-SSIM and FID evaluation metrics. Moreover, synthetic images with the best MS-SSIM and FID scores are used to augment the imbalanced dataset. A baseline CNN classifier is trained on the standard and augmented datasets to compare the classification score including accuracy, recall, specificity, etc. The improved accuracy of 91.50 \% and specificity of 0.79 are achieved with the augmented dataset having AIIN-DCGAN synthetic images as compared to the alternate datasets.

\subsubsection{Modified Architecture}
In GANs, if a new architecture is defined with an alternative generator or discriminator or both as compared to the vanilla GAN then we describe it as modified architecture.

\paragraph{Generator}
An alternative generator introduced in the proposed architecture of GAN is described as the modified generator. To avoid the mode collapse problem, a widely adopted approach is to use multiple generators instead of a single as in vanilla GAN which has proved effective to alleviate the problem \cite{hoang2018mgan}. However, optimizing multiple generators is complicated and costs extensively large computations. 

\begin{figure}[htp!]
    \centering
    \includegraphics[width=0.7\textwidth]{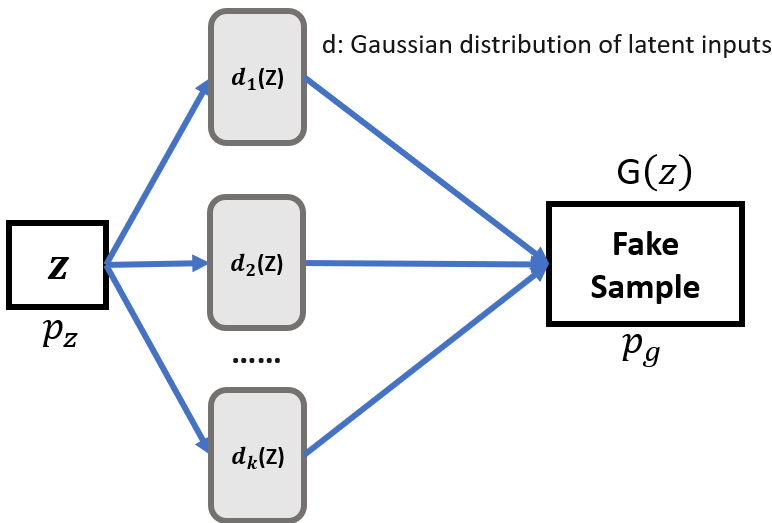}
    \caption{The Generator architecture of MDGAN. The figure is redesigned from \cite{wu2018end}.}
    \label{Fig.GenMDGAN}
\end{figure}
To address this limitation, Wu et al. \cite{wu2018end} proposed the idea to use multiple distributions instead of using multiple generators to synthesize human cell images. A Gaussian Mixture Model (GMM) based generator is used to cover each data distribution in the latent space as indicated in Fig. \ref{Fig.GenMDGAN}. It helps the proposed MDGAN to generate diverse image samples using a mixture of data distributions. Moreover, the authors argued that more distributions can aid in generating more diverse synthetic image samples but can lead to huge computational costs. The generated human cell images are then used to augment the dataset for classification tasks. To evaluate generated images, no quantitative analysis is reported in the paper. While authors discussed that the generated synthetic images aid in data augmentation and improve the classification performance of CNN by 4.6 \% precision value.

The hierarchy of layers of the generator and discriminator models. To interpret this idea, Qin et al. \cite{qin2020gan} proposed an extension to the StyleGAN as skin-lesion StyleGAN (SL-StyleGAN) for synthesizing skin lesion images. In \cite{qin2020gan}, the authors discussed that changing the number of fully-connected layers in a mapping network of the generator can control the generation of different modes of images. In baseline Style-GAN \cite{karras2019style}, a generator consists of a non-linear mapping network that maps latent input $z$ to an intermediate latent space $W$ using MLP network and then passes the $W$ information to the original generator model. Furthermore, the authors attempted 2, 4, and 6 fully-connected layers and evaluated the generated images with a recall score. They investigated that the generator with 2 fully-connected layers can generate relatively more diverse images than 6 but results in scattered defects like artifacts, etc. The generator model with 4 fully-connected layers can generate relatively good diverse images with no artifacts. The final SL-StyleGAN architecture with a generator of 4 fully-connected layers achieved a 0.263 recall score which is higher than alternate fully-connected layer combinations. The authors concluded that the final synthetic images are not fully diverse as indicated by the lower recall score which needs more work in the future to address this problem.

\begin{figure}[htp!]
    \centering
    \includegraphics[width=1\textwidth]{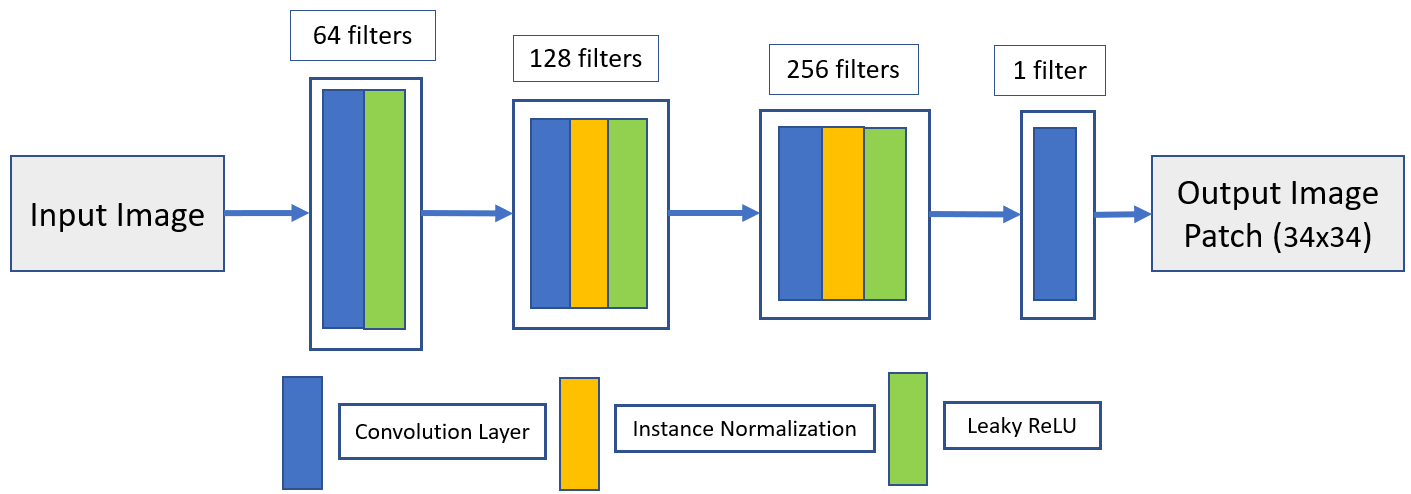}
    \caption{Patch Discriminator of CycleGAN. The figure is redesigned from \cite{modanwal2021normalization}.}
    \label{Fig.DisCycleGAN}
\end{figure}
\paragraph{Discriminator}
An alternative discriminator introduced in the proposed architecture of GAN is known as the modified discriminator. In GANs, when the generator collapses to a single mode and produces identical image samples then the discriminator backpropagates identical gradients for several generator updates. There is no coordination between the discriminator and its gradients because it deals with each training sample independently. So, no mechanism guides the generator to produce diverse image samples. To address this problem in MR to MR image translation of breast slices, Modanwal et al. \cite{modanwal2021normalization} use a small field of view 34x34 instead of 70x70 in standard Patch discriminator as depicted in Fig. \ref{Fig.DisCycleGAN} in the CycleGAN. The small field of view encourages the transformation learned by the generator to maintain the sharp and high-frequency details. This modification of the CycleGAN preserves the structural information of breast and dense tissues during the training of GAN to perform image translation tasks.

The generated images are evaluated by dice coefficient and compared with the standard CycleGAN. The standard CycleGAN has a mean value of 0.8913 and a standard deviation of 0.0941 for GE to SE translation while the mean value of 0.9089 and a standard deviation of 0.0471 for SE to GE translation. GE Healthcare and Siemens are the two source scanners for image acquisition. Authors have achieved an improved mean value of 0.9801 and a standard deviation of 0.0061 for GE to SE translation while a mean value of 0.9813 and a standard deviation of 0.0049 for SE to GE translation on the test data.

\begin{figure}[htp!]
    \centering
    \includegraphics[width=1\textwidth]{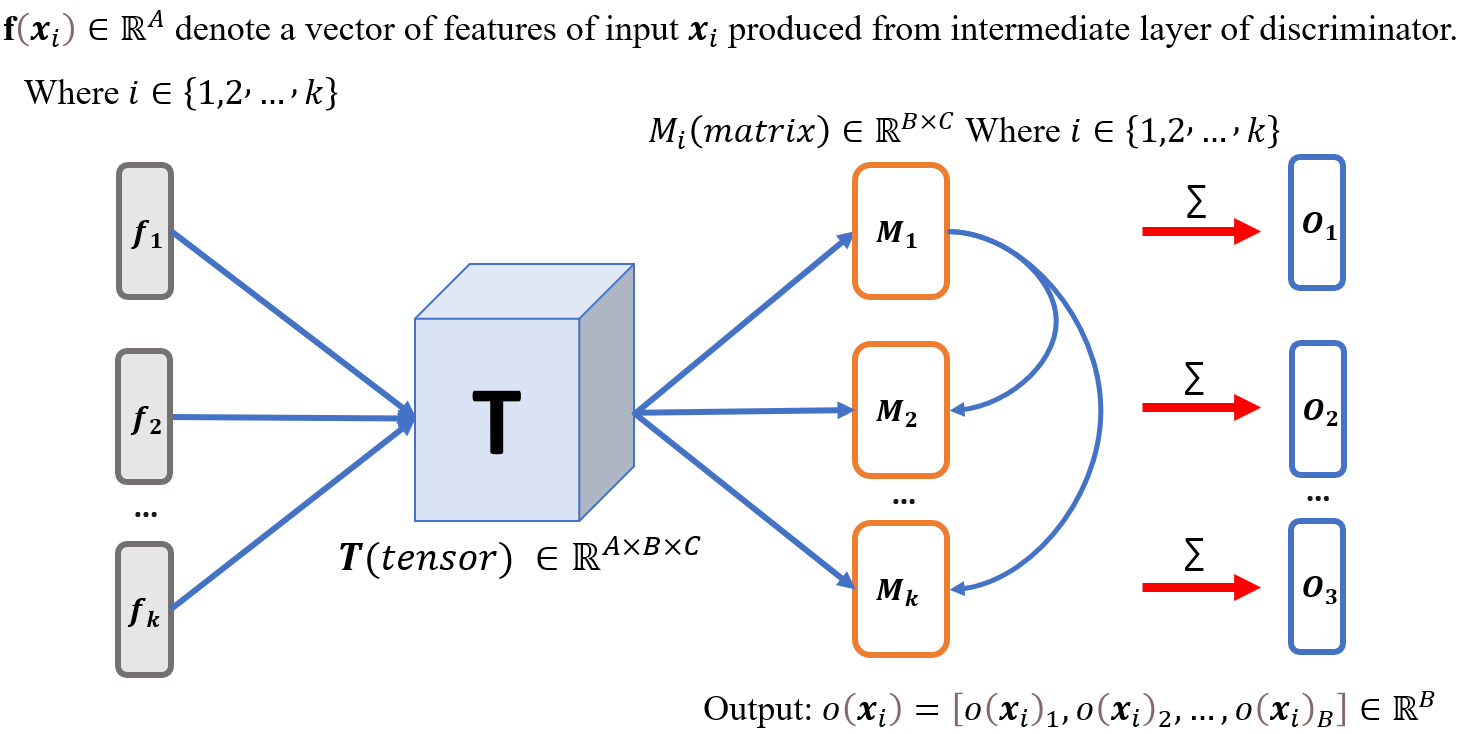}
    \caption{The workflow of Minibatch Discrimination. The figure is redesigned from \cite{salimans2016improved}.}
    \label{Fig.minibatch}
\end{figure}
Cervical histopathology images contain fine-grained information that is difficult to learn by GANs and can cause the mode collapse problem. To address the mode collapse in synthesizing cervical histopathology images, authors in \cite{xue2019synthetic} utilize mini-batch discrimination in the discriminator of CGAN to generate realistic diverse samples. The Minibatch discrimination enables the coordination between gradients of discriminator and training samples using mini-batches for training image samples as depicted in Fig. \ref{Fig.minibatch}. In this way, the generator is penalized if it collapses to a single mode and is regulated to produce diverse images \cite{salimans2016improved}. The synthetic images are not evaluated by any metric to check the diversity or similarity measures with real images. The generated synthetic images are then used to augment the dataset for classification tasks.

\begin{figure}[htp!]
    \centering
    \includegraphics[width=1\textwidth]{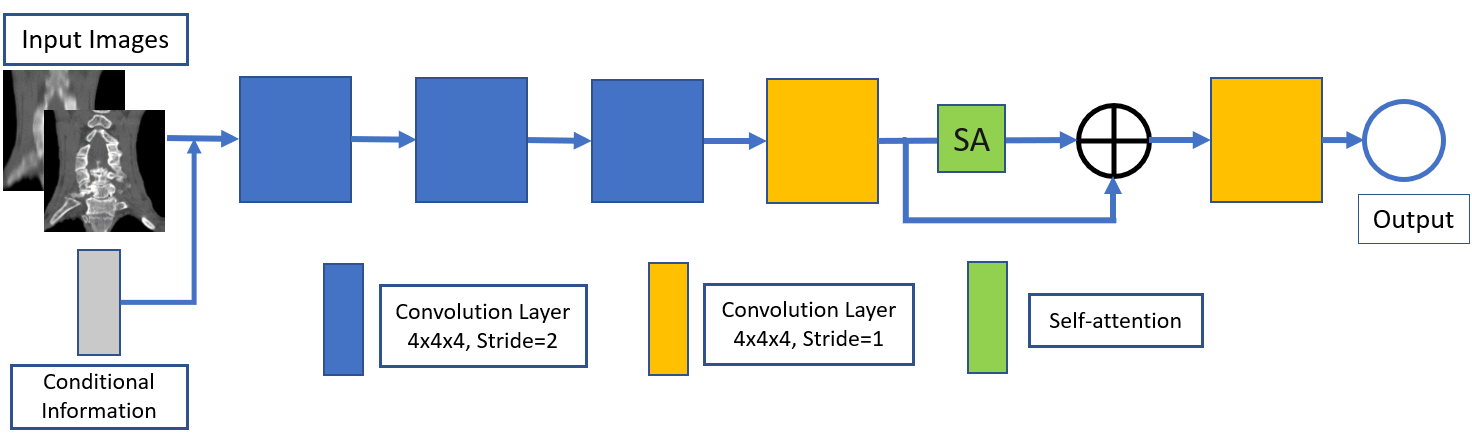}
    \caption{Conditional Informational based three dimensional CNN Discriminator of CGAN \cite{kudo2019virtual}. The figure is redesigned from \cite{kudo2019virtual}.}
    \label{Fig.DisCGAN}
\end{figure}
A similar problem of generating diverse synthetic image samples occurs in CGAN when dealing with distinct CT scans of different body parts for a super-resolution task. To address this problem, a conditional information vector $w$ based modified discriminator is proposed in \cite{kudo2019virtual}. The discriminator is composed of a 3-dimensional fully convolutional neural network as shown in Fig. \ref{Fig.DisCGAN}. The conditional vector $w$ contains information about input image data such as leg, head, abdomen, or chest. This information is used by the discriminator to evaluate the generated slices of CT data and encourages the generator to produce diverse image samples. The generated super-resolution images are evaluated through SSIM and PSNR scores. The highest score of SSIM (0.933) and PSNR (35.73) are achieved respectively as compared to the CGAN without conditional vector $w$. The SSIM score shows a similarity measure and realistic nature of generated images towards ground truth images.

\begin{table}[htp!]
\centering
\caption{\textbf{A comparative analysis of contributing papers highlighting training problems of GANs based on GAN variant, proposed solution, image modality, and evaluation metric.}}
\begin{tabular}{lp{2cm}p{2cm}p{2cm}p{3cm}p{3cm}p{0.2cm}}

\toprule

\textbf{Training Problem} & \textbf{References} & \textbf{GAN Variant} & \textbf{Image Modality} & \textbf{Proposed Solution} & \textbf{Evaluation Metric} \\

\midrule

\multirow{13}{*}{Mode Collapse} & Qin et al. \cite{qin2020gan} & SL-StyleGAN & Dermoscopic Images & Varying number of fully-connected layers & Recall \\
& Lau et al. \cite{lau2018scargan} & ScarGAN & MR Images & Experience replay buffer & - \\
& Wu et al. \cite{wu2018end} & MDGAN & Chromosome Cell Images & Gaussian Mixture Model as generator & - \\
& Modanwal et al. \cite{modanwal2021normalization} & CycleGAN & MR Images & Patch discriminator (34x34) & dice coefficient \\
& Xue et al. \cite{xue2019synthetic} & Modified CGAN & Histopathology Images & Minibatch discrimination & - \\
& Kudo et al. \cite{kudo2019virtual} & CGAN & CT Images & Discriminator based on 3D CNN with conditional information vector & - \\
& Segato et al. \cite{segato2020data} & DCR Auto-Encoding Alpha GAN & MR Images & Skip connections & MMD and MS-SSIM \\
& Kwon et al. \cite{kwon2019generation} & Auto-Encoding GAN & MR Images & VAEGAN & MMD and MS-SSIM \\
& Abdelhalim et al. \cite{abdelhalim2021data} & SPGGAN & Dermoscopic Images & Self attention mechanism & Feature level maps \\
& Neff et al. \cite{Neffseg2017} & DCGAN & X-ray Images & Perceptual image hashing & - \\
& Saad et al. \cite{saad2022addressing} & AIIN-DCGAN & X-ray Images & Adaptive Input-image Normalization & MS-SSIM and FID \\
& Saad et al. \cite{saad2022self} & MSG-SAGAN & X-ray Images & Self-attention and multi-scale gradients & MS-SSIM and FID \\
& Xu et al. \cite{xu2020low} & SNSRGAN & X-ray Images & Spectral Normalization & - \\
&&&&&& \\
\multirow{3}{*}{Non-convergence} & Abdelhalim et al. \cite{abdelhalim2021data} & SPGGAN-TTUR & Dermoscopic Images & Two Time-scale Update Rule (TTUR) & Paired t-test \\
& Goel et al. \cite{goel2021automatic} & Optimized GAN & CT Images & Whale optimization algorithm & - \\
& Biswas et al. \cite{biswas2019synthetic} & uGAN & Retinal Images & Modified training updates of generator and discriminator & SSIM \\
&&&&&& \\
\multirow{8}{*}{Instability} & Xue et al. \cite{xue2019synthetic} & Modified CGAN & Histopathology Images & WGAN-GP loss & - \\ 
& Segato et al. \cite{segato2020data} & DCR Auto-Encoding Alpha GAN & MR Images & WGAN-GP loss & - \\
& Kwon et al. \cite{kwon2019generation} & Auto-Encoding GAN & MR Images & WGAN-GP loss & - \\ 
& Wei et al. \cite{wei2020predicting} & CF-SAGAN & MR Images & Residual connections & PSNR \\ 
& Wu et al. \cite{ciGAN2018} & ciGAN & Mammography Images & Multi-scale generator & - \\ 
& Zhao et al. \cite{zhao2020study} & S-CycleGAN & PET Images & WGAN loss & learned perceptual image patch similarity (LPIPS) score \\
& Deepak et al. \cite{deepak2020msg} & MSG-GAN & MR Images & WGAN-GP loss & - \\ 
& Saad et al. \cite{saad2022self} & MSG-SAGAN & X-ray Images & Relativistic Hinge loss & MS-SSIM and FID \\

\bottomrule
\end{tabular}
\label{tabchalleng}
\end{table}

\paragraph{Generator-Discriminator Combined}

\begin{figure}[htp!]
    \centering
    \includegraphics[width=1\textwidth]{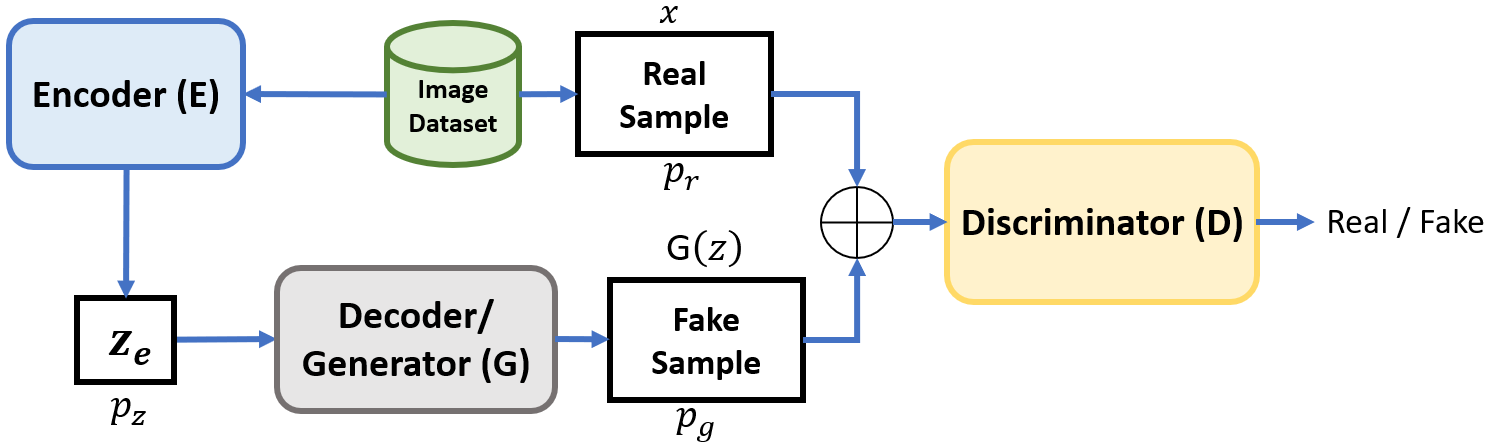}
    \caption{The architecture of VAEGAN. The figure is redesigned from \cite{kwon2019generation}}
    \label{Fig.VAEGAN}
\end{figure}
In this section, we describe the architecture of GANs where the generator and the discriminator are updated or modified. The generation of diversified synthetic 3-dimensional (3D) Magnetic Resonance images is a challenging task. This is due to the complexity of the structure of 3D image data. To address this limitation, authors in \cite{kwon2019generation} adopted an $\alpha$-GAN with few modifications in the activation functions, batch normalization, and loss function. The $\alpha$-GAN is composed of a Variational Auto-encoder (VAE) and a code discriminator network. The VAE is a generative model that explicitly learns the likelihood distributions of training data rather than the other model's feedback as in GANs to generate synthetic image samples \cite{kingma2014stochastic}. A GAN combined with VAE can learn the likelihood distributions of images which results in the generation of diversified synthetic images as shown in Fig. \ref{Fig.VAEGAN}. In contrast, VAE generates blurry images. $\alpha$-GAN utilizes the advantage of VAE in alleviating the mode collapse problem in 3D MR image generation. The authors of \cite{kwon2019generation} proposed an Auto-encoding GAN and generate 3D MR images with different latent input $z$ sizes like 100, 1000, and 2048. With a latent vector input of 1000, the proposed Auto-encoding GAN can generate diverse image samples while it fails to escape mode collapse with too small (100) or too large (2048) latent vector input sizes. 

To evaluate the diversity of synthetic images, authors \cite{kwon2019generation} calculated average MMD x $10^{-4}$ and MS-SSIM scores. The results show that the proposed GAN can perform better with a latent input value of 1000 with an average MMD x $10^{-4}$ score of 0.072 and MS-SSIM of 0.829. The MS-SSIM of real data is 0.846. MS-SSIM score of synthetic 3D MR images shows a good similarity measure with the real data and can be a good candidate for generating diverse images. However, there is a gap in generating more robust and diverse images with smooth and artifact-free images. 

\begin{figure}[htp!]
    \centering
    \includegraphics[width=1\textwidth]{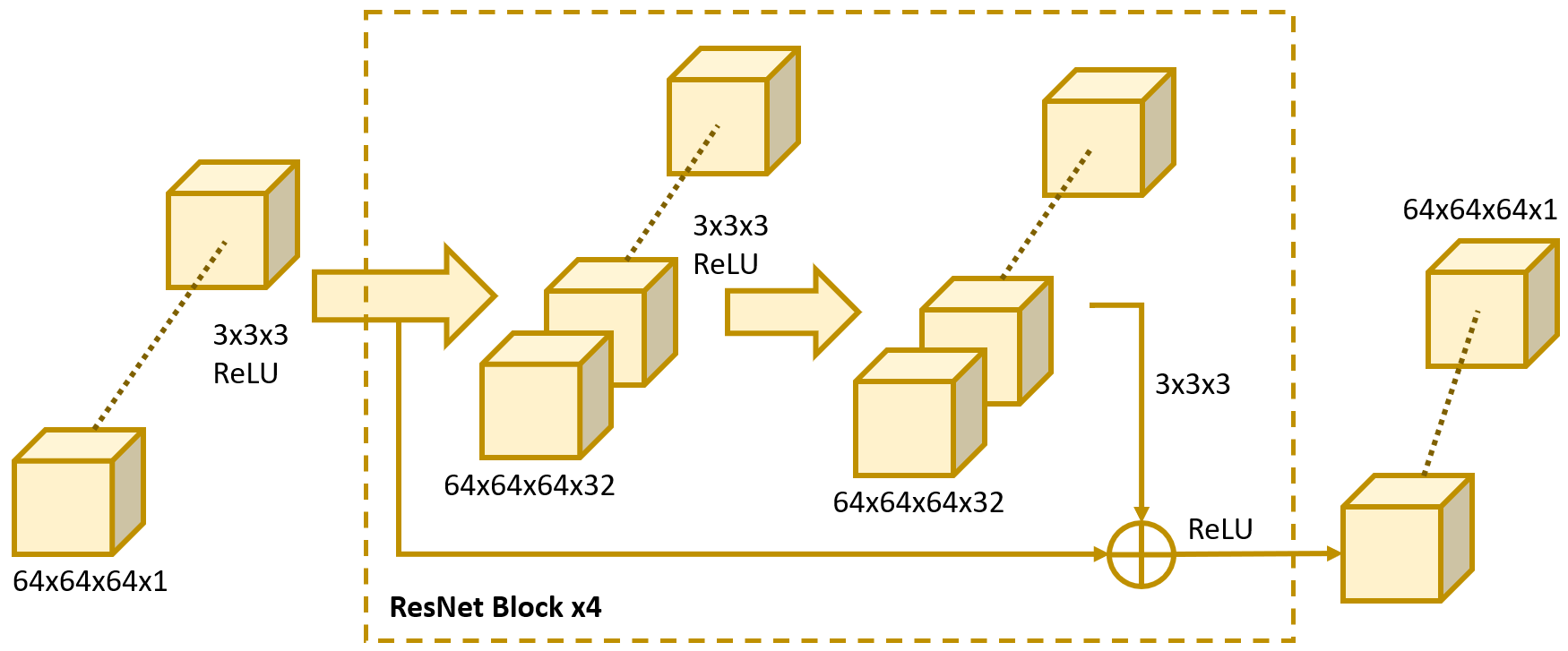}
    \caption{The deep convolutional refiner architecture of DCR-AEGAN. The figure is redesigned from \cite{segato2020data}.}
    \label{Fig.RefineralphaGAN}
\end{figure}
To bridge this gap, authors in \cite{segato2020data} extend this work \cite{kwon2019generation} by applying a refiner network based on ResNet blocks \cite{targ2016resnet} to generate realistic 3D MR images. The ResNet uses skip connections with deep convolutions as shown in Fig. \ref{Fig.RefineralphaGAN} which controls the skipping of some training layers to smooth the shapes of generated images and make them more realistic. However, this work delivers a low diversity score evident from the MS-SSIM score of 0.9991 between generated images which indicates the lowest diversity of synthetic images as compared to the real images. The proposed deep convolutional refiner GAN \cite{segato2020data} achieved a good score of MMD as (0.2240 $\pm$ 0.0008) x $10^{4}$ as compared to the previous score of MMD as (0.5932 $\pm$ 0.0004) x $10^{4}$ which shows the realistic nature of generated images.

\begin{figure}[htp!]
    \centering
    \includegraphics[width=1\textwidth]{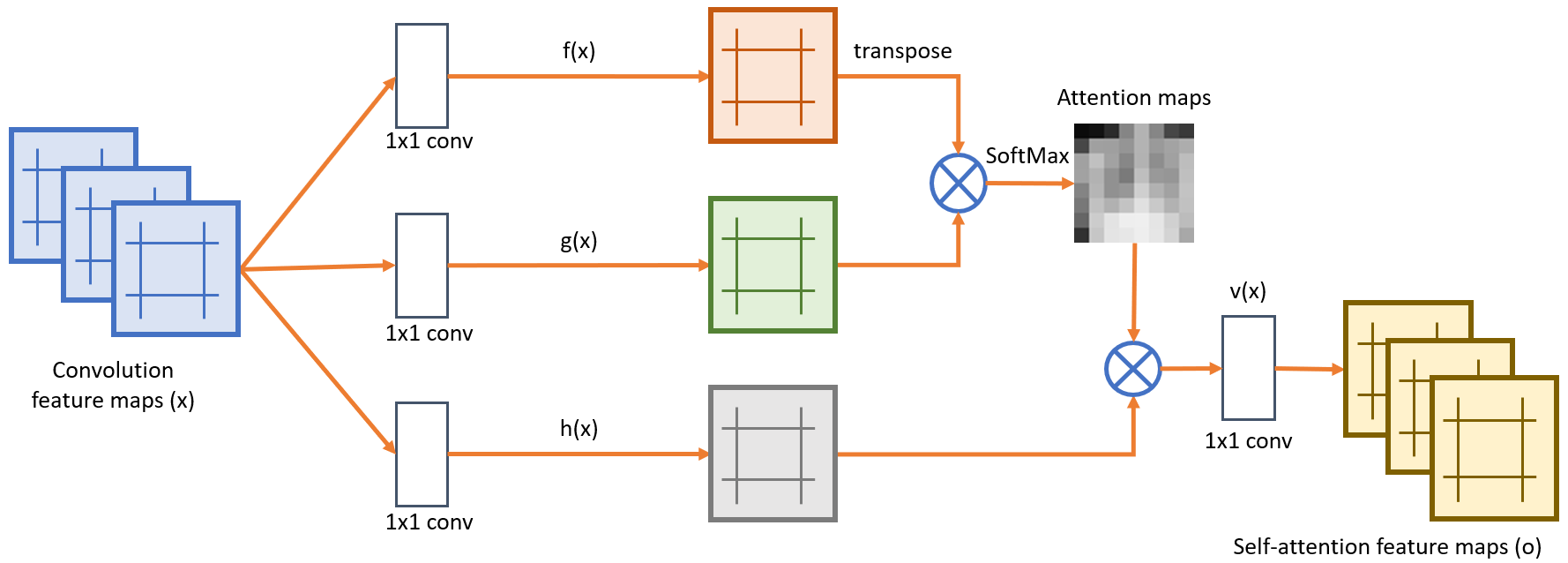}
    \caption{Self-attention mechanism. The figure is redesigned from \cite{zhang2019self}.}
    \label{Fig.selfattention}
\end{figure}
The mode collapse can occur in a GAN when biomedical images contain complex information of salient features which are difficult to learn and model a relationship between them. A similar type of limitation is addressed for Dermoscopic skin lesion images in a progressive growing GAN (PGGAN) using a self-attention mechanism by authors in \cite{abdelhalim2021data}. They discussed that most image synthesis tasks in biomedical imagery utilize PGGAN built with convolutional layers. While in convolutional layers, the convolutional filters are dependent on local neighborhood information to process the convolution operations. It is computationally inefficient for convolutional filters to capture the long-range dependencies in images by relying only on convolutional layers. So, a self-attention mechanism is adapted that enables the discriminator to preserve image features with relevant activations to a particular task. It utilizes feature attention maps that help the generator to produce synthetic images in which coordination should be observed between fine details at every location and fine details in distant portions of the images as shown in Fig. \ref{Fig.selfattention}. Besides, the discriminator can judge the consistency of highly detailed features in distant portions of the image. In this way, the generator becomes capable of generating diverse image samples using a self-attention mechanism in PGGAN (SPGGAN).

Different feature level maps are used for evaluating the performance of the self-attention mechanism in image synthesis of resolution 128 x 128 pixels. The (N-1)-to-(N) stage in SPGGAN and PGGAN is monitored which represents the $2^{N-1}$-to-$2^{N}$ level feature maps where $N = 7$. As a result, SPGGAN performs better with $70.1$\% as compared to PGGAN with $67.7$\% for the training set at $N = 6$. Similarly, SPGGAN performs better with $62.2$\% as compared to PGGAN with $60.8$\% for the test set at $N = 6$. However, the real dataset has feature maps of 78.2\%. It shows that the proposed SPGGAN attains better diversity and realistic image synthesis performance than PGGAN yet is distant from real images.

Saad et al. \cite{saad2022self} also utilized a self-attention mechanism in the multi-scale gradient GAN (MSG-GAN) to generate diversified X-ray images. They integrated a self-attention layer into each layer of the generator and discriminator models. The self-attention utilizes attention features maps to help the MSG-GAN to learn and focus on the diverse features of X-ray images as shown in Fig. \ref{Fig.selfattention}. The authors demonstrated an improvement in the diversity of generated synthetic images using an improved FID score of 139.6.

\subsubsection{Adversarial Training}
This section discusses the alterations made during the training of GANs such as making buffer storage \cite{lau2018scargan} or using perceptual image hash \cite{Neffseg2017} to identify and address the mode collapse problem.

\paragraph{Buffer Storage Scheme}
Generation or simulation of diverse scar tissues in the myocardium of the left ventricle from a segmented healthy Late-gadolinium enhancement (LGE) imaging scan using GANs is always a challenging task. Scar tissue is a fibrosis tissue that appears when healthy tissue gets destroyed by some disease. \cite{lau2018scargan} proposed a variant of GAN namely ScarGAN that is composed of a convolutional U-Net-based architecture \cite{ronneberger2015u} both in the generator as well as in the discriminator. In ScarGAN, an experience replay buffer scheme \cite{shrivastava2017learning} is used to prevent the generator from producing similar shapes of scar tissue. In this scheme, half of the generated masks are stored in a buffer for an experience replay. From this buffer, the discriminator uses half of the training batches randomly to check previously generated scar tissue samples and prevent the generator from producing similar shapes of scar tissue.

The generated images from ScarGAN \cite{lau2018scargan} are evaluated by experienced physicians. These physicians are provided with 15 generated and 15 real images in a mixed dataset. They classify them with an accuracy of 53 \% which reflects a good score for the realism of generated images. However, the authors concluded that ScarGAN still generates less diverse shapes of scar tissues i.e. similar shapes that require to be researched in the future.   

\paragraph{Perceptual Image Hashing}

\begin{figure}[htp!]
    \centering
    \includegraphics[width=1\textwidth]{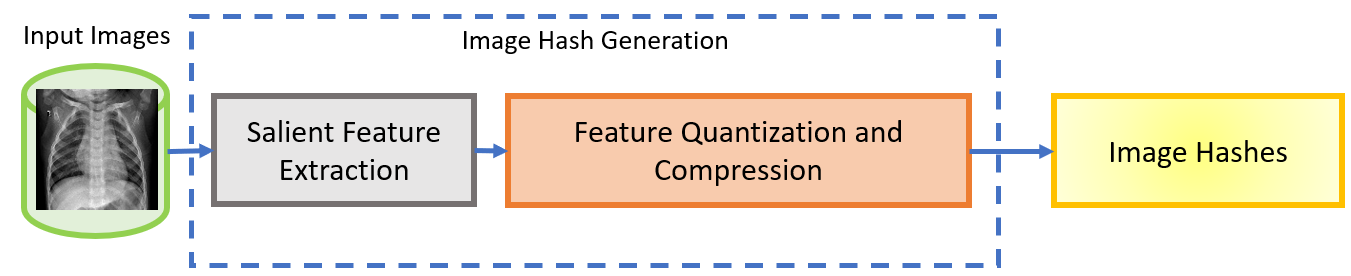}
    \caption{A flow methodology of image hash generation. The figure is redesigned from \cite{du2020perceptual}.}
    \label{Fig.image hash}
\end{figure}
Generating new segmentation masks and ground-truth images separately from GANs is a time taking task. To generate new chest X-ray images and segmentation masks, \cite{Neffseg2017} proposed a variant of DCGAN that forces the generator to produce a segmentation mask together with ground truth images. During the adversarial training, the generator starts producing identical image-segmentation pairs with few artifacts that lead to a mode collapse problem. To address this problem, the authors use the perceptual image hash function to remove the identical generated image-segmentation pair. Perceptual image hash functions calculate hash values of real and generated images based on specific image features as shown in Fig. \ref{Fig.image hash}. These hash values are compared further to evaluate the difference between generated and real images.

The generated image-segmentation pair is evaluated in data augmentation for the segmentation task. The U-Net is trained on 30 real and 120 generated images. The lowest Hausdorff distance of 7.2885 has been observed as compared to the results when U-Net trained on only real images or only generated images. However, the authors concluded that a mild form of mode collapse occurred which results in less diverse images.

\subsubsection{Summary}

\begin{figure}[htp!]
    \centering
\begin{center}
        \resizebox{0.6\textwidth}{!}{%
            \begin{forest}
                for tree={
                    draw,
                    rounded corners,
                    node options={align=center},
                    text width=2.7cm,
                    anchor=center,
                           },
                where level=0{%
      }{%
        folder,
        grow'=0,
        if level=1{%
          before typesetting nodes={child anchor=north},
          edge path'={(!u.parent anchor) -- ++(0,-20pt) -| (.child anchor)},
        }{},
        }
                [Proposed Solutions for Mode Collapse Problem \\ in GANs, rotate = 0, fill=gray!30, parent
                [Regularization, rotate = 0, for tree={fill=red!30, child}
                [Weight \\ Normalization: \\ SNSRGAN \cite{xu2020low} (2020), rotate = 0, for tree={fill=red!30, child}]
                [Input \\ Normalization: \\ AIIN-DCGAN \cite{saad2022addressing} (2022), rotate = 0, for tree={fill=red!30, child}]
                ]
                [Modified \\ Architecture, rotate = 0, for tree={fill=green!50, child}, calign with current edge
                [Generator: \\ MD-GAN \cite{wu2018end} (2018) \\ SL-StyleGAN \cite{qin2020gan} (2020), rotate = 0]
                [Discriminator: \\ CycleGAN \cite{modanwal2021normalization} (2021) \\
                Modified CGAN \cite{xue2019synthetic} (2019)
                \\ CGAN \cite{kudo2019virtual} (2019), rotate = 0]
                [Generator-Discriminator Combined: \\ Auto-Encoding GAN \cite{kwon2019generation} (2019)
                \\ DCR Auto-Encoding Alpha GAN \cite{segato2020data} (2020)
                \\ SPGGAN \cite{abdelhalim2021data} (2021)
                \\ MSG-SAGAN \cite{saad2022self} (2023), rotate = 0]]
                [Adversarial \\ Training, rotate = 0, for tree={fill=cyan!40, child}
                [Buffer Strategy: \\ ScarGAN \cite{lau2018scargan} (2018), rotate = 0]
                [Image Hashing: \\ DCGAN \cite{Neffseg2017} (2017), rotate = 0]]]
            \end{forest}
        }
    \end{center}
    \caption{Taxonomy of different proposed solutions for addressing the mode collapse problem of GANs in biomedical imagery analysis.}
    \label{fig: mode collapse}
    \end{figure}
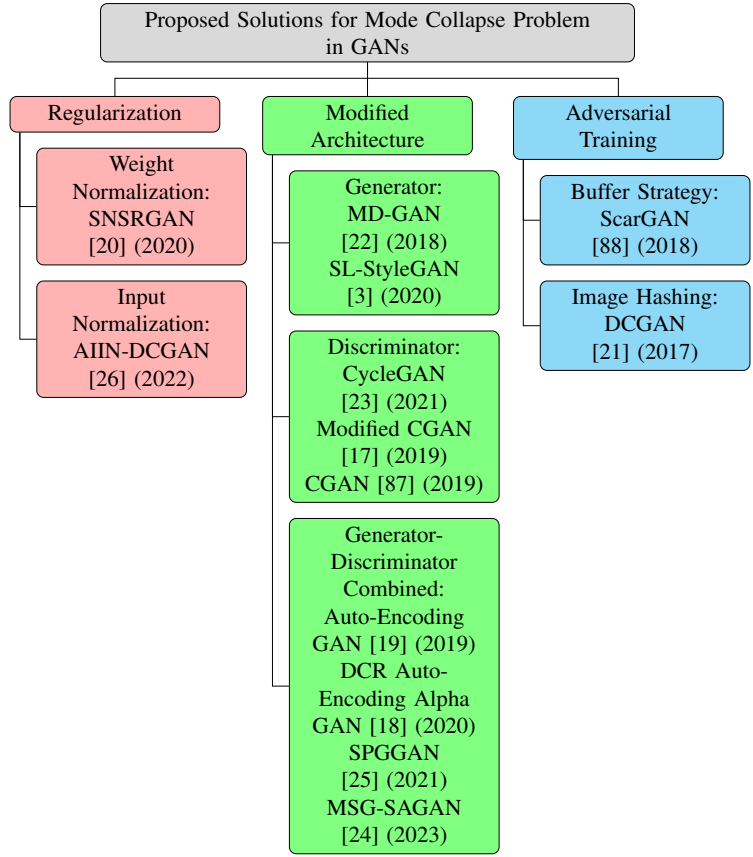

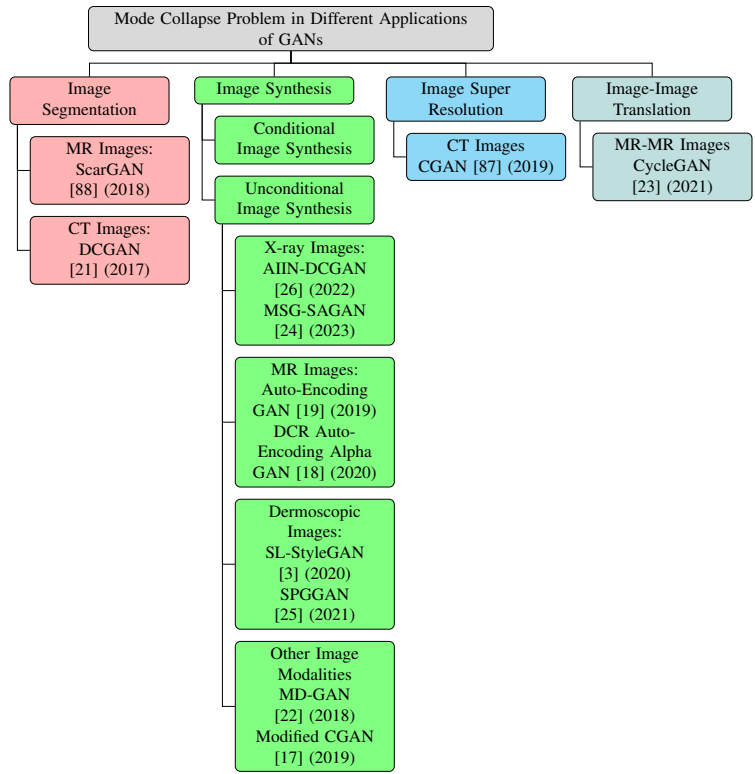
\begin{figure}[htp!]
    \centering
\begin{center}
        \resizebox{0.6\textwidth}{!}{%
            \begin{forest}
                for tree={
                    draw,
                    rounded corners,
                    node options={align=center},
                    text width=2.7cm,
                    anchor=center,
                                    },
                where level=0{%
      }{%
        folder,
        grow'=0,
        if level=1{%
          before typesetting nodes={child anchor=north},
          edge path'={(!u.parent anchor) -- ++(0,-20pt) -| (.child anchor)},
        }{},
        }
                [Mode Collapse Problem in Different Applications \\ of GANs, rotate = 0, fill=gray!30, parent
                [Image \\ Segmentation, rotate = 0, for tree={fill=red!30, child}
                [MR Images: \\ ScarGAN \cite{lau2018scargan} (2018), rotate = 0]
                [CT Images: \\ DCGAN \cite{Neffseg2017} (2017), rotate = 0]
                ]
                [Image Synthesis, rotate = 0, for tree={fill=green!50, child}
                [Conditional \\ Image Synthesis, rotate = 0]
                [Unconditional \\ Image Synthesis, rotate = 0,
                [X-ray Images: \\ AIIN-DCGAN \cite{saad2022addressing} (2022) \\ MSG-SAGAN \cite{saad2022self} (2023), rotate = 0]
                [MR Images: \\ Auto-Encoding GAN \cite{kwon2019generation} (2019) \\ DCR Auto-Encoding Alpha GAN \cite{segato2020data} (2020), rotate = 0]
                [Dermoscopic Images: \\ SL-StyleGAN \cite{qin2020gan} (2020) \\ SPGGAN \cite{abdelhalim2021data} (2021), rotate = 0]
                [Other Image Modalities \\ MD-GAN \cite{wu2018end} (2018) \\ Modified CGAN \cite{xue2019synthetic} (2019), rotate = 0]]]
                [Image Super \\ Resolution, rotate = 0, for tree={fill=cyan!40, child}
                [CT Images \\ CGAN \cite{kudo2019virtual} (2019), rotate = 0]]
                [Image-Image \\ Translation, rotate = 0, for tree={fill=teal!25,child}
                [MR-MR Images \\ CycleGAN \cite{modanwal2021normalization} (2021), rotate = 0]]
                ]
            \end{forest}
        }
    \end{center}
    \caption{An application-based taxonomy of different approaches for addressing the mode collapse problem of GANs in biomedical imagery analysis.}
    \label{fig: mode collapse appl}
    \end{figure}

\begin{longtable}[htp!]{lp{1.3cm}p{1.7cm}p{0.5cm}p{0.5cm}p{0.35cm}p{0.5cm}p{5.8cm}}

\caption{\textbf{An overview of existing solutions via preprocessing, modified architectures, adversarial training, and loss functions to address the mode collapse problem in GANs for biomedical image analysis.}}
\label{ganhowoverviewmodecollapse} \\

\toprule
\textbf{References} & \textbf{GANs Variant} & \textbf{Image Modality} & \textbf{Pre-proc:} & \textbf{Mod. Arch.} & \textbf{Adv. Tr.} & \textbf{Ls. Funs.} & \textbf{Summary of Solutions} \\

\midrule
\endfirsthead

\multicolumn{8}{l}{\footnotesize\itshape\tablename~\thetable:
continued from the previous page} \\
\toprule
\textbf{References} & \textbf{GANs Variant} & \textbf{Image Modality} & \textbf{Pre-proc:} & \textbf{Mod. Arch.} & \textbf{Adv. Tr.} & \textbf{Ls. Funs.} & \textbf{Summary of Solutions} \\
\midrule
\endhead

\midrule
\multicolumn{8}{l}{Pre-proc: Pre-processing; Mod. Arch: Modified Architectures; Adv. Tr: Adversarial Training; Ls. Func: Loss Function} \\
\multicolumn{8}{r}{\footnotesize\itshape\tablename~\thetable:
Continued on next page} \\
\endfoot

\bottomrule
\multicolumn{8}{l}{Pre-proc: Pre-processing; Mod. Arch: Modified Architectures; Adv. Tr: Adversarial Training; Ls. Func: Loss Function} \\
\multicolumn{8}{r}{\footnotesize\itshape\tablename~\thetable:
It ends from the previous page.} \\
\endlastfoot

Qin et al. \cite{qin2020gan} & SL-StyleGAN & Dermoscopic Images & x & \checkmark & x & x & The generator of StyleGAN is fine-tuned with the number of fully-connected layers to find the best co-ordination of skin lesion features during the training of GAN. The generator with 4 fully-connected layers is found optimal to generate the improved diversified skin lesion images as compared to the alternative combinations of layers.\\
& \\
Lau et al. \cite{lau2018scargan} & ScarGAN & MR Images & x & x & \checkmark & x & An experience replay buffer scheme is used during the training of a GAN to address the mode collapse problem. In this scheme, half of the generated masks are stored in a buffer for an experience replay. From this buffer, the discriminator uses half of the training batches randomly to check previously generated scar tissue samples and prevent the generator from producing similar shapes of scar tissue samples.\\
& \\
Wu et al. \cite{wu2018end} & MDGAN & Cell Images & x & \checkmark & x & x & A Gaussian Mixture Model-based generator is used to address the mode collapse in MDGAN for cell image synthesis. A Gaussian mixture of data distributions helps a GAN to cover each data distribution in the latent space for generating diversified synthetic image samples. \\
& \\
Modanwal et al. \cite{modanwal2021normalization} & CycleGAN & MR Images & x & \checkmark & x & x & A 34x34 patch discriminator is used to address the mode collapse in CycleGAN for MR-to-MR image translation. A small field of view 34x34 in the discriminator encourages the MR image transformation learned by the generator to maintain the sharp and high-frequency details while preserving the structural information of breast and dense tissues during the training of GAN for the image translation task. \\
& \\
Xue et al. \cite{xue2019synthetic} & Modified CGAN & Histopathology Images & x & \checkmark & x & x & A minibatch discrimination is used in the discriminator of CGAN to address the mode collapse problem and the generation of diversified histopathology images. The discriminator uses mini-batches to train the image samples as it creates coordination between the gradients of the discriminator and training samples. Minibatch discrimination is used to penalize the generator if it collapses to a single mode and regulates it to produce diverse histopathology images. \\
& \\ 
Kudo et al. \cite{kudo2019virtual} & CGAN & CT Images & x & \checkmark & x & x & A conditional information-based discriminator is proposed to address the mode collapse in CGAN for the generation of super-resolution CT images. The proposed discriminator uses conditional information on body parts such as the leg, head, abdomen, or chest from CT images to guide the generator in producing diversified super-resolution images. \\
& \\
Segato et al. \cite{segato2020data} & DCR AEGAN & MR Images & x & \checkmark & x & x & A deep convolutional refiner is used to generate the diversified MR images in AEGAN. A ResNet-based convolutional refiner uses skip connections with deep convolutions to control the skipping of some training layers for smoothing the shapes of generated images which results in the diversified generation of images. \\
& \\
Kwon et al. \cite{kwon2019generation} & AEGAN & MR Images & x & \checkmark & x & x & A Variational Auto-encoder (VAE) is used to address the mode collapse in AEGAN for diversified 3D MR image synthesis. VAE in the GAN learns the likelihood distributions of training images which helps to improve the generator's learning of generating diversified synthetic MR images. \\
& \\
Abdelhalim et al. \cite{abdelhalim2021data} & SPGGAN & Dermoscopic Images & x & \checkmark & x & x & A self-attention mechanism is proposed to address the mode collapse in the PGGAN for Dermoscopic image synthesis. Dermoscopic images contain fine-grained information about skin lesions which is important to learn during the generation of synthetic images. A self-attention mechanism utilizes feature attention maps to focus on skin lesion features that help the generator produce diversified synthetic images. \\
& \\
Neff et al. \cite{Neffseg2017} & DCGAN & X-ray Images & x & x & \checkmark & x & A perceptual image hash function is used to address the mode collapse in DCGAN for segmentation of X-ray images. The perceptual image hash function removes the identical generated image-segmentation pair during the training of GAN. It calculates hash values of real and generated images based on specific image features and then compared them to evaluate the difference between generated and real images. \\
& \\
Xu et al. \cite{xu2020low} & SNSRGAN & X-ray Images & x & \checkmark & x & x & Spectral Normalization is used to address the mode collapse in SNSRGAN for super-resolution of X-ray images. Spectral normalization is a regularization technique that employs the spectral norm of weight matrices corresponding to the largest singular vector in the discriminator which controls the Lipschitz constant to 1. It guides a generator to produce diversified synthetic X-ray images. \\
& \\
Saad et al. \cite{saad2022addressing} & AIIN-DCGAN & X-ray Images & \checkmark & x & x & x & AIIN is used to address the mode collapse in DCGAN. It normalizes the contrast of images while highlighting the salient features of interest in chest X-ray images. It helps a discriminator to learn real images more accurately and guides a generator to produce diversified synthetic X-ray images. \\
& \\
Saad et al. \cite{saad2022self} & MSG-SAGAN & X-ray Images & x & \checkmark & x & x & A self-attention mechanism is proposed in the layers of generator and discriminator of MSG-GAN to generate diversified synthetic X-ray images. Self-attention uses attention feature maps to help a GAN for modeling the long-range relationship between significant features of chest X-ray images. It also helps the MSG-GAN to emphasize selective features such as the spatial, shape, and structure of X-ray images during the training of GAN for the generation of diversified images. \\
\end{longtable}
    
In this section, technical papers are reviewed to address the mode collapse problem in the biomedical imagery domain. The mode collapse problem can be alleviated by using different methods such as regularization, modified architectures, and adversarial training. These methods are reviewed as solutions to the underlying problem in the domain of biomedical imagery. A taxonomy is created based on these solutions as shown in Fig. \ref{fig: mode collapse}. In Fig. \ref{fig: mode collapse}, each sub-category is further divided into different methods like regularization has weight normalization, modified architectures are divided into the generator, discriminator, and generator-discriminator combined. Similarly, adversarial training is further divided into possible solutions like buffer schemes and perceptual image hash. The application-based taxonomy is also created as shown in Fig. \ref{fig: mode collapse appl}. This taxonomy \ref{fig: mode collapse appl} helps to analyze the effect of mode collapse for the specific type of biomedical images.

From the technical literature, it is reviewed that all of the papers utilized approaches that partially alleviate the problem of mode collapse in biomedical imagery. The Auto-encoding GAN \cite{kwon2019generation} provides relatively more diverse synthetic images while addressing the problem in biomedical imagery. Table \ref{tabchalleng} provides a comparative analysis of contributing papers to address the underlying training challenges in GANs for biomedical imagery.

Moreover, a detailed overview of each solution is also listed in Table \ref{ganhowoverviewmodecollapse} where each solution based on three categories such as preprocessing, modified GAN architectures, and loss functions is outlined. Table \ref{ganhowoverviewmodecollapse} summarises how each solution addressed the mode collapse problem in GANs for the biomedical imagery domain.

\section{The Non-convergence Problem}
\label{section:Non-convergence}
\subsection{Definition}
\label{section: Non-convergence: Def}
In GANs, it is important that the training of the generator and the discriminator should converge at a global point (Nash equilibrium). The training of GANs is performed as a minimax game to reach this Nash equilibrium. The discriminator and the generator should be trained with the best training strategies to achieve better training. As the generator's performance improves, it gets difficult for the discriminator to distinguish synthetic images from real images. When the generator is producing the best plausible (realistic-looking) images, the discriminator will have a classification accuracy of 50\%. Consequently, the discriminator has no meaningful feedback to update the weights of the generator. This will affect the synthetic images produced by the generator. As a result, the training of GANs leads to a non-convergence problem \cite{arjovsky2017towards}.

\subsection{Identification}
\label{section: Non-convergence: Ident}
\begin{figure}[htp!]
    \centering
    \includegraphics[width=1\textwidth]{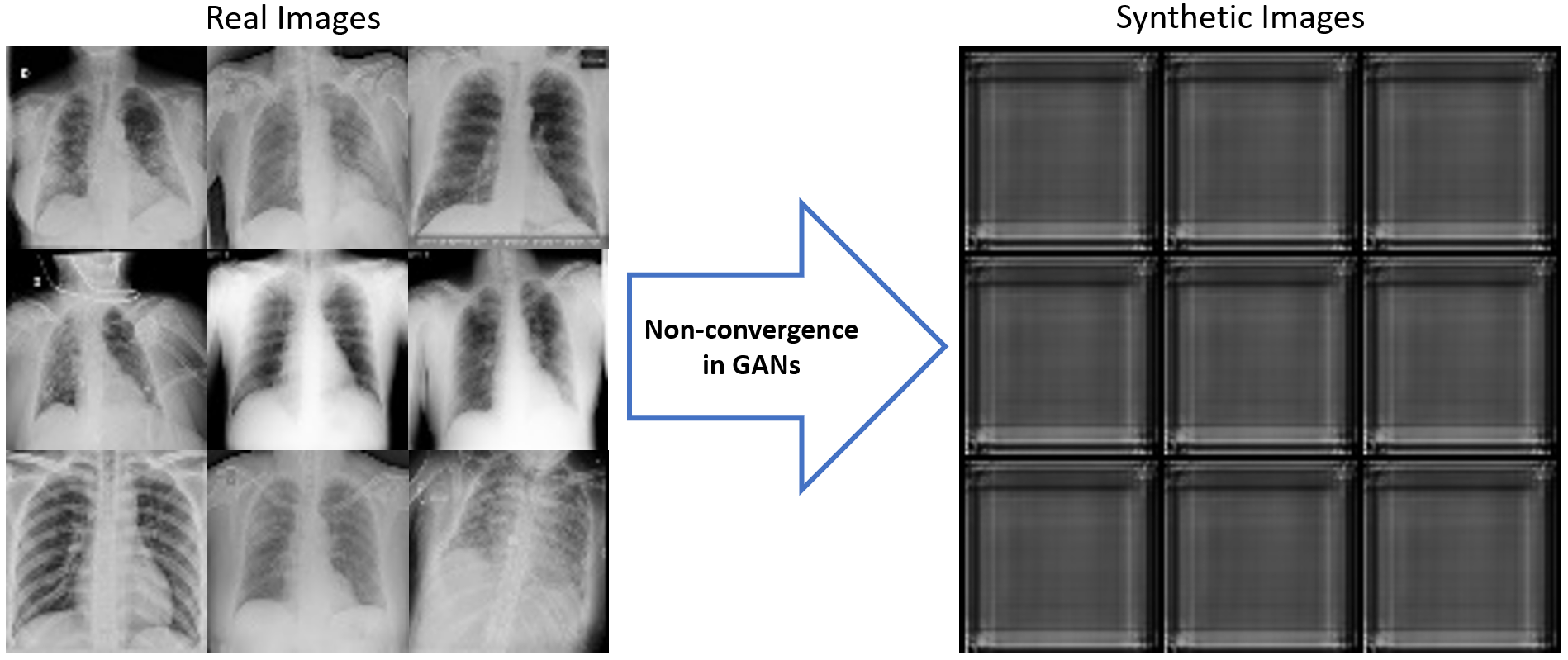}
    \caption{Identification of the non-convergence problem in GANs for X-ray image synthesis. The plane black color synthetic images with no information show the imbalanced training of the GANs for X-ray image synthesis. It shows a failure of the generator and discriminator models during the training of GANs for generating X-ray images.}
    \label{Fig.non-converge}
\end{figure}

The non-convergence problem has a direct effect on the generation of synthetic images. The underlying problem is identified by analyzing the nature of synthetic images. The non-convergence problem leads the generator to produce plane color images such as black or white in the case of gray-scale images as indicated in Fig. \ref{Fig.non-converge}.

\subsection{Quantification}
\label{section: Non-convergence: Quant}
To evaluate the problem of non-convergence in GANs, evaluation metrics are proposed to judge the quality of generated images. So, several evaluation metrics are proposed such as peak signal-to-noise ratio (PSNR) \cite{borji2019pros} and FID \cite{heusel2017gans} to quantify the quality of generated images as discussed in Section \ref{eval_metrics}.

\subsection{Solutions to the Problem}
\label{section: Non-convergence: Soltn}
\subsubsection{Nash Equilibrium}
This section discusses the possible solutions in terms of using optimization algorithms and controlling the training iteration ($k$) to find a Nash equilibrium.

In vanilla GAN \cite{goodfellow2014generative}, Goodfellow demonstrated that an equilibrium can be achieved with an optimal discriminator during the training of GAN. However, this is an ideal case, and in practice, GAN does not meet the condition. So, the author \cite{goodfellow2014generative} proposed an algorithm to update the discriminator multiple times $(k)$ per generator's training update to get the discriminator close to an ideal. In vanilla GAN, the discriminator is updated only once $(k = 1)$ per generator's training update which was suitable for that specific experiment. Similarly, WGAN \cite{arjovsky2017wasserstein} uses $(k = 5)$ for discriminator updates per generator's training update for attaining an equilibrium state.

\paragraph{Updating Algorithm}
It is a very critical and sensitive approach to control the training updates of the generator and discriminator models to reach a balanced state of training. Biswas et al. \cite{biswas2019synthetic} proposed a uGAN with separate parameters ($k$) for the discriminator and ($r$) for the generator to control the updates of the training iteration of both of these models. The authors investigated that the similar number of updates for both models yields balanced training and the generation of high-quality retinal synthetic images. It is also analyzed that $k$ with large values can generate high-quality realistic images by keeping $r = 1$. In contrast, noisy images are generated using larger values of $r$ with $k = 1$.

The synthetic images are evaluated with an SSIM metric. The mean, maximum, and mean-maximum values of SSIM are measured between synthetic and real images to check the quality and similarity between images. A higher score of SSIM shows higher similarity and high-quality measures. The mean SSIM score of 0.61, maximum SSIM score of 0.73, and mean-maximum SSIM score of 0.81 are achieved.

\paragraph{Learning Rate}
The idea of using learning rates to stabilize and balance the training of GANs is proposed by Heusel et al. \cite{heusel2017gans}. The authors introduced a novel algorithm namely the Two Time-scale Update Rule (TTUR) to achieve a local Nash equilibrium using distinct learning rates of the discriminator and the generator instead of using multiple update algorithms. TTUR uses stochastic gradient learning $\boldsymbol{g}(\boldsymbol{\theta}, \boldsymbol{w})$ of the discriminator's loss and $\boldsymbol{h}(\boldsymbol{\theta}, \boldsymbol{w})$ of the generator's loss. It defines the true gradients of $\boldsymbol{g}(\boldsymbol{\theta}, \boldsymbol{w})=\nabla_w \mathcal{L}_D$ and $\boldsymbol{h}(\boldsymbol{\theta}, \boldsymbol{w})=\nabla_\theta \mathcal{L}_G$ with random variables $\boldsymbol{M}^{(w)}$ and $\boldsymbol{M}^{(\theta)}$ as shown in Eq. \ref{randomvariable} reported from \cite{heusel2017gans}. So, it uses stochastic learning $b(n)$ and $a(n)$ for updating the discriminator and generator steps respectively as defined in Eq. \ref{ttur} reported from \cite{heusel2017gans}. However, the choice of appropriate learning rates depends on the GAN architecture, type of experiments, and nature of the datasets.

\begin{equation}
\tilde{\boldsymbol{g}}(\boldsymbol{\theta}, \boldsymbol{w})=\boldsymbol{g}(\boldsymbol{\theta}, \boldsymbol{w})+\boldsymbol{M}^{(w)} \text { and } \tilde{\boldsymbol{h}}(\boldsymbol{\theta}, \boldsymbol{w})=\boldsymbol{h}(\boldsymbol{\theta}, \boldsymbol{w})+\boldsymbol{M}^{(\theta)}\label{randomvariable}
\end{equation}

\begin{equation}
\boldsymbol{w}_{n+1}=\boldsymbol{w}_n+b(n)\left(\boldsymbol{g}\left(\boldsymbol{\theta}_n, \boldsymbol{w}_n\right)+\boldsymbol{M}_n^{(w)}\right), \boldsymbol{\theta}_{n+1}=\boldsymbol{\theta}_n+a(n)\left(\boldsymbol{h}\left(\boldsymbol{\theta}_n, \boldsymbol{w}_n\right)+\boldsymbol{M}_n^{(\theta)}\right)\label{ttur}
\end{equation}

Abdelhalim et al. \cite{abdelhalim2021data} investigated the use of both TTUR \cite{heusel2017gans} and discriminator updates in SPGGAN for skin lesion image synthesis. The authors updated the discriminator 5 times for every single update of the generator's training. The update algorithm slows down the training process while TTUR tries to balance it to generate noise-free images.

SPGGAN-TTUR \cite{abdelhalim2021data} shows visually appealing results of generated images as compared to SPGGAN. The results are evaluated through a paired t-test with $95\%$ confidence $(p-value < 0.05)$. Paired t-test gives the mean difference between two sample observations. The P-value of the t-test (PVT) is calculated to check the performance of SPGGAN-TTUR for generating synthetic train and test sets images. The PVT of $68.1 \pm 0.8\%$ for the training set while $60.8 \pm 1.5$ for test sets are achieved which outperformed the SPGGAN. However, SPGGAN-TTUR \cite{abdelhalim2021data} suffers from artifacts in the generated image that need to be researched.

\paragraph{Hyperparameter Optimization}
In GANs, the choice of appropriate hyperparameters to control the discriminator and the generator is a challenging task. To address this problem, optimization techniques can be used to obtain adaptive losses for updating the weights of the generator.

\begin{figure}[htp!]
    \centering
    \includegraphics[width=1\textwidth]{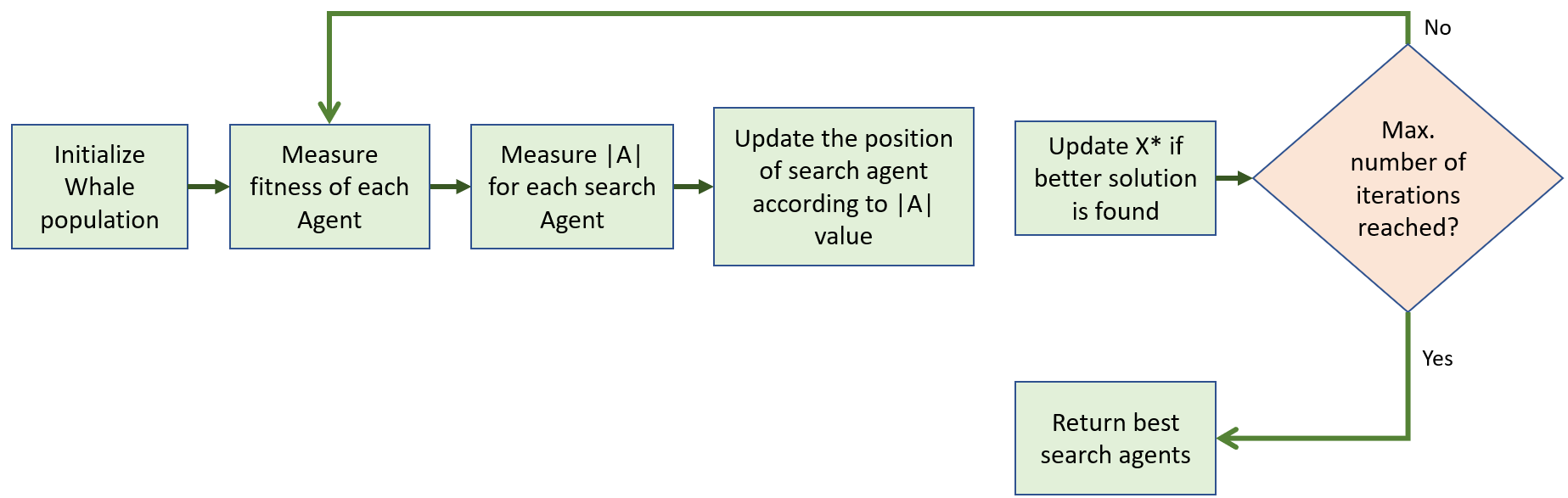}
    \caption{The flow diagram of the Whale optimization algorithm. The figure is redesigned from \cite{goel2021automatic}.}
    \label{Fig.whalealgo}
\end{figure}

Goel et al. \cite{goel2021automatic} proposed an optimized GAN to generate synthetic chest CT images of COVID-19 disease. The optimized GAN utilizes a CGAN with Whale Optimization Algorithm (WOA) \cite{mirjalili2016whale} to optimize its hyperparameters. A flow of the Whale optimization algorithm is shown in Fig. \ref{Fig.whalealgo}. In this algorithm, the hunting trick of humpback whales is adapted to optimize the prey's location. This hunting trick determines the generator's best search agents with the given discriminator. To update the position of search agents, the optimization of hyperparameters follows three rules; first, the leader whale finds the prey's position and encircles it. Similarly, the generator's search agents calculate the fitness function at each iteration to achieve the best position and then update their positions. Second, the distance between the prey and the location of the generator's search agents is measured and then the generator's search agents update their position based on these measures. Third, it is the same as the first rule but it updates the position of search agents based on the random search instead of the best search as in the first rule. The Optimized GAN \cite{goel2021automatic} improves the performance of the discriminator and can generate adaptive losses to update weights of the generator to produce good quality diverse images.

The performance of Optimized GAN \cite{goel2021automatic} is compared with the baseline CGAN. The generated images are used with training images for classification tasks. So, the F1-score and accuracy of $98.79\%$ and 98.78\% respectively are achieved with Optimized GAN while  91.60\% accuracy and 90.99\% F1-score are achieved with the baseline CGAN. It shows that Optimized GAN can perform better with accuracy and F1-score measures, as well as in optimizing hyperparameters for a balanced GAN.

\subsubsection{Summary}

\begin{table}[htp!]
\centering
\caption{\textbf{An overview of existing solutions via preprocessing, modified architectures, adversarial training, and loss functions to address the non-convergence problem in GANs for biomedical image analysis.}}

\begin{tabular}{lp{1cm}p{1.4cm}p{0.3cm}p{0.3cm}p{0.3cm}p{0.3cm}p{7cm}}

\toprule

\textbf{References} & \textbf{GANs Variant} & \textbf{Image Modality} & \textbf{Pre-proc:} & \textbf{Mod. Arch.} & \textbf{Adv. Tr.} & \textbf{Ls. Funs.} & \textbf{Summary of Solutions} \\

\midrule

 Abdelhalim et al. \cite{abdelhalim2021data} & SPGGAN-TTUR & Dermoscopic Images & x & x & \checkmark & x & The Two Time-scale Update Rule is used to achieve a local Nash equilibrium. The TTUR uses distinct learning rates of the discriminator and the generator instead of multiple update algorithms to balance the training and enable the generation of high-quality synthetic skin lesion images. \\
& \\
 Goel et al. \cite{goel2021automatic} & Optimized GAN & CT Images & x & x & \checkmark & x & A Whale Optimization Algorithm (WOA) is used to balance the training of CGAN for CT image synthesis. The WOA optimizes the hyperparameters of CGAN using the hunting trick of humpback whales. It uses a strategy of finding an optimal value of the prey's location. This hunting trick is used to determine the generator's best search agents with the given discriminator which yields balanced training of GAN. \\
& \\
 Biswas et al. \cite{biswas2019synthetic} & uGAN & Retinal Images & x & x & \checkmark & x & Similar number of training steps are proposed to update the generator and discriminator models to reach the balanced state during the training of uGAN. High-quality synthetic Retinal images are generated with a balanced training of uGAN. \\

\bottomrule
\multicolumn{8}{l}{Pre-proc: Pre-processing; Mod. Arch: Modified Architectures; Adv. Tr: Adversarial Training; Ls. Func: Loss Function}
\end{tabular}
\label{ganoverviewhownonconverge}
\end{table}

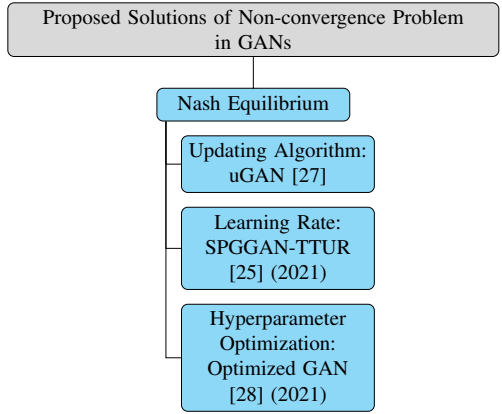
\begin{figure}
    \centering
\begin{center}
        \resizebox{0.4\textwidth}{!}{%
            \begin{forest}
                for tree={
                    draw,
                    rounded corners,
                    node options={align=center},
                    text width=2.7cm,
                    anchor=center,
                                 },
                where level=0{%
      }{%
        folder,
        grow'=0,
        if level=1{%
          before typesetting nodes={child anchor=north},
          edge path'={(!u.parent anchor) -- ++(0,-28pt) -| (.child anchor)},
        }{},
        }
                [Proposed Solutions of Non-convergence Problem \\ in GANs, rotate = 0, fill=gray!30, parent
                [Nash Equilibrium, rotate = 0, for tree={fill=cyan!40, child}, calign with current edge
                [Updating Algorithm: \\ uGAN \cite{biswas2019synthetic}, rotate = 0]
                [Learning Rate: \\ SPGGAN-TTUR \cite{abdelhalim2021data} (2021), rotate = 0]
                [Hyperparameter \\ Optimization: \\ Optimized GAN \cite{goel2021automatic} (2021), rotate = 0]]]
            \end{forest}
        }
    \end{center}
    \caption{Taxonomy of different proposed solutions for addressing the non-convergence problem of GANs in biomedical imagery analysis.}
    \label{fig: non-con}
    \end{figure}

\begin{figure}
    \centering
\begin{center}
        \resizebox{0.4\textwidth}{!}{%
            \begin{forest}
                for tree={
                    draw,
                    rounded corners,
                    node options={align=center},
                    text width=2.7cm,
                    anchor=center,
                          },
                where level=0{%
      }{%
        folder,
        grow'=0,
        if level=1{%
          before typesetting nodes={child anchor=north},
          edge path'={(!u.parent anchor) -- ++(0,-28pt) -| (.child anchor)},
        }{},
        }
                [Non-convergence Problem in Different Applications \\ of GANs, rotate = 0, fill=gray!30, parent
                [Image Synthesis, rotate = 0, for tree={fill=cyan!40, child}, calign with current edge
                [Unconditional \\ Image Synthesis, rotate = 0,
                [CT Images: \\ Optimized GAN \cite{goel2021automatic}, rotate = 0]
                [Dermoscopic Images \\ SPGGAN-TTUR \cite{abdelhalim2021data} (2021), rotate = 0]
                [Retinal Images \\ uGAN \cite{biswas2019synthetic}, rotate = 0]]]]
            \end{forest}
        }
    \end{center}
    \caption{An application-based taxonomy of different approaches for addressing the non-convergence problem of GANs in biomedical imagery analysis.}
    \label{fig: non-conv appl}
    \end{figure}
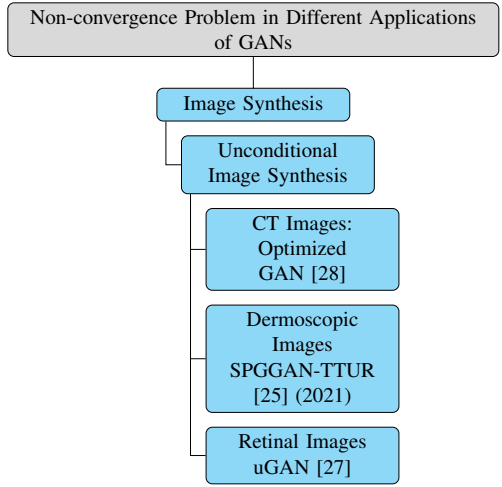
    
In this section, technical papers of GANs are reviewed to address the non-convergence problem in the domain of biomedical imagery. Achieving a Nash equilibrium during the training of GANs is a remedy to this non-convergence problem \cite{goodfellow2016nips}. Training GANs at an equilibrium state is not an easy task. By keeping this concept in mind, the reviewed papers are classified into three different categories as shown in Fig. \ref{fig: non-con}. First, updating algorithms \cite{biswas2019synthetic}, second, learning rate \cite{abdelhalim2021data}, and third is hyperparameter optimization \cite{goel2021automatic}. Another taxonomy is also proposed on application-based biomedical imagery as shown in Fig. \ref{fig: non-conv appl}. This is further classified into image modality types such as dermoscopic \cite{abdelhalim2021data}, CT \cite{goel2021automatic}, and retinal images \cite{biswas2019synthetic}.

The updating algorithm is reviewed for vanilla GAN \cite{goodfellow2014generative}, WGAN \cite{arjovsky2017wasserstein}, and then state-of-the-art uGAN \cite{biswas2019synthetic}. The updating algorithms in vanilla GAN \cite{goodfellow2014generative} and WGAN \cite{arjovsky2017wasserstein} are proposed for the general imagery domain while updating algorithm in uGAN \cite{biswas2019synthetic} is proposed for the biomedical imagery domain. All of these propose strategies to update discriminator time-steps per generator time-steps during the training of GANs. They show that their proposed solutions work better in attaining an equilibrium state while training the GANs.

Another idea of achieving equilibrium in training the GANs is proposed by \cite{heusel2017gans}. It also helps to achieve an equilibrium using adaptive learning rates for the discriminator and the generator. This technique is used by Abdelhalim et.al \cite{abdelhalim2021data} to address the non-convergence problem in the biomedical domain. The Hyperparameter optimization approach is also helpful in reaching the Nash equilibrium. For this, Goel et. al \cite{goel2021automatic} investigated the use of optimization algorithms such as the Whale optimization algorithm (WOA) \cite{mirjalili2016whale} for biomedical imagery.

To summarize this section, Table \ref{tabchalleng} shows a comparison of proposed techniques adapted by the contributing papers based on the underlying problem. It is observed that all of the technical papers belong to the image synthesis of CT, dermoscopic, and retinal image modalities. Among all of the contributed solutions, the TTUR \cite{heusel2017gans} scheme provides relatively good performance to address the non-convergence problem in the biomedical imaging domain. High-quality realistic images can be achieved using this approach in biomedical imagery.

Moreover, a detailed overview of existing solutions to address the non-convergence problem in GANs is also reported in Table \ref{ganoverviewhownonconverge} where the methodology of each solution to the non-convergence problem in GANs is summarized for the domain of biomedical imagery.

\section{The Instability Problem}
\label{section:Instabilityproblem}

\subsection{Definition}
\label{section: Instability: Def}
The training of the GANs can get unstable due to the vanishing gradient problem. The vanishing gradient problem occurs when the discriminator becomes an optimal classifier and produces smaller values of gradients (approaching zero) for back-propagation. These gradients are unable to update the weights of the generator due to which the generator stops producing new images and the overall training of the GANs becomes unstable \cite{goodfellow2016nips}.

\subsection{Identification}
\label{section: Instability: Ident}
\begin{figure}[htp!]
    \centering
    \includegraphics[width=1\textwidth]{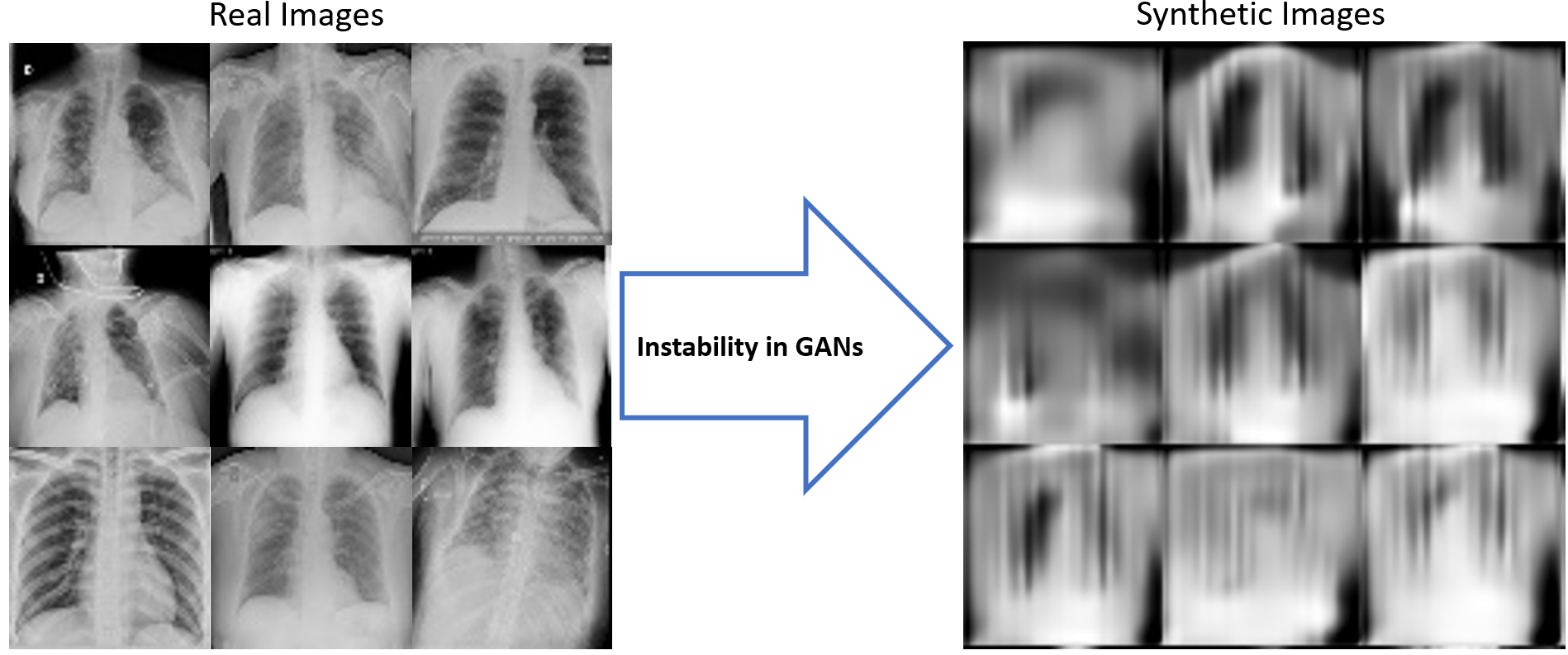}
    \caption{Identification of the instability problem in GANs for X-ray image synthesis. The noisy synthetic X-ray images show unstable training of the GANs for X-ray images. The blurriness in the images is generated due to the vanishing gradient problem which is a basic reason for the unstable training of GANs.}
    \label{Fig.Instability}
\end{figure}

The instability during the training of GANs is identified by the generation of blurry or low-quality synthetic images as indicated in Fig. \ref{Fig.Instability}. Moreover, the underlying problem takes a longer time to train GANs with unstopping behavior which results in generating poor-quality images. Another drawback of the instability problem is that it will lead the generator to produce synthetic images with artifacts. These artifacts include noise or additional objects that are not meant for generated.

\subsection{Quantification}
\label{section: Instability: Quant}
The instability problem of training GANs can be evaluated by the same metrics that are used for mode collapse and non-convergence problems such as MS-SSIM \cite{odena2017conditional}, FID \cite{heusel2017gans}, and PSNR \cite{borji2019pros}. The quality of generated images can be evaluated in terms of similarity measures as discussed in Section \ref{eval_metrics}. Furthermore, classification metrics such as recall and precision are also used to quantify the quality of synthetic images.

\subsection{Solutions to the Problem}
\label{section: Instability: Soltn}
In synthetic image generation using GANs, the stability of GANs is an important aspect to consider. If the training of GANs becomes unstable, the network could not generate high-resolution realistic images. To alleviate this problem, the following possible solutions are proposed for the domain of biomedical imagery.

\subsubsection{Modified Architecture}

\begin{figure}[htp!]
    \centering
    \includegraphics[width=0.7\textwidth]{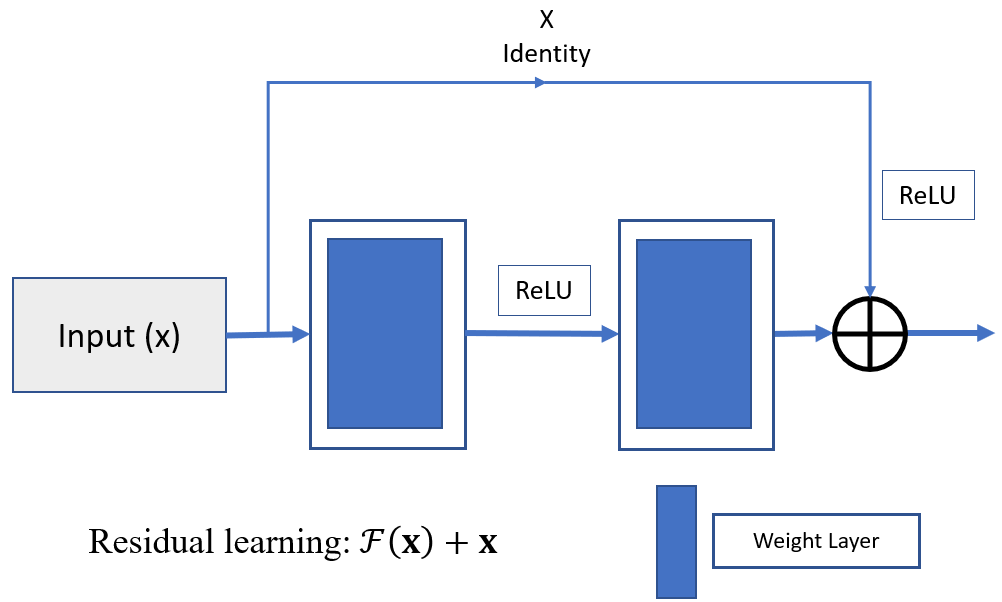}
    \caption{A flow of Residual learning in CF-SAGAN. The figure is redesigned from \cite{he2016deep}.}
    \label{Fig.Residual}
\end{figure}
The architecture of GANs plays a key role to avoid the vanishing gradient problem. The selection of the generator and the discriminator have a great impact on the training performance of GANs. To synthesize PET images from multi-sequence MR images, a Refined CF-SAGAN is proposed \cite{wei2020predicting}. In the proposed architecture \cite{wei2020predicting}, the problem of vanishing gradient occurs when the long-skip connections are used in the generator to recover the lost spatial information during the down-sampling operations. Then, short skip connections are used to handle this problem. This process is known as the residual connections \cite{he2016deep}. The residual connection helps to mitigate the problem of vanishing gradient by allowing an alternative shortcut track for the gradient to flow through as shown in Fig. \ref{Fig.Residual} the training of GANs. It also enhanced the feature exchanges across layers. The generated synthetic PET images are evaluated with the PSNR for image quality. The proposed Refined CF-SAGAN outperformed by $9.07\%$ in PSNR (p $<$ 0.05).

\begin{figure}[htp!]
    \centering
    \includegraphics[width=1\textwidth]{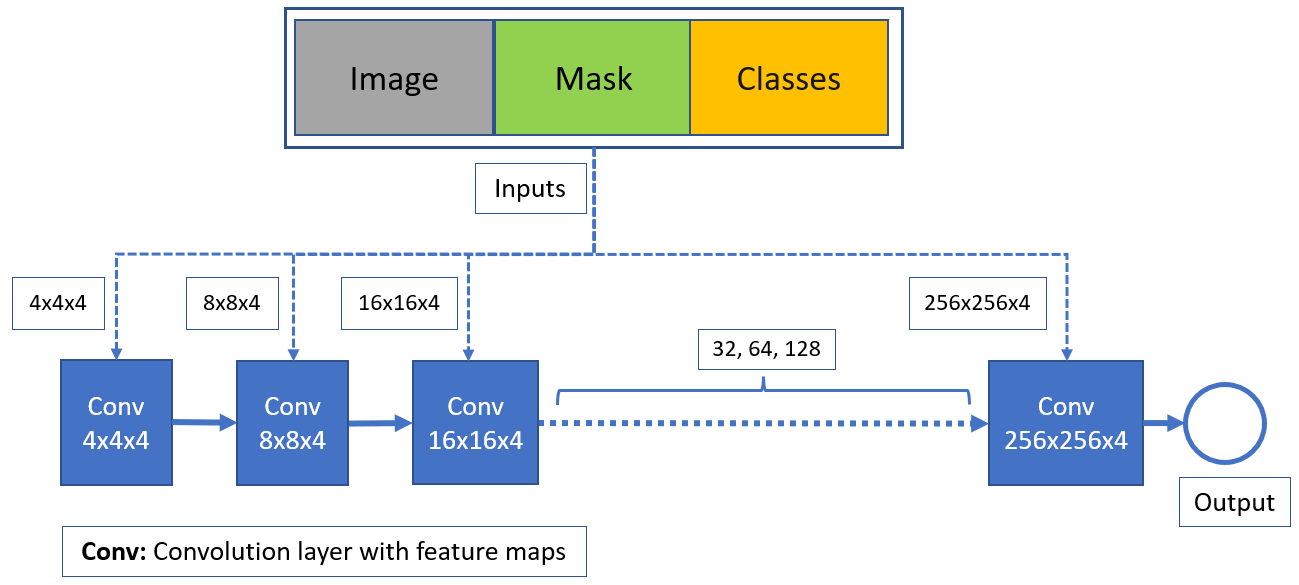}
    \caption{The architecture of multi-scale Generator of ciGAN. The generator takes input at multi-level layers with multiple image resolutions. This figure is redesigned from \cite{ciGAN2018}.}
    \label{Fig.GenciGAN}
\end{figure}
The generation of high-dimensional synthetic images is a challenging task using GANs. To address this problem in biomedical imaging, a modified architecture namely ciGAN is proposed \cite{ciGAN2018}. The ciGAN \cite{ciGAN2018} utilizes a multi-scale generator architecture as depicted in Fig. \ref{Fig.GenciGAN} to infill a segmented area in a target image of breast Mammography. The proposed generator uses a cascaded refinement network that helps to generate features at multiple scales before being concatenated. This process improves the training stability at high resolutions. The generated synthetic images are used for data augmentation in the cancer detection task using ResNet-50. Traditional augmentation techniques like rotation, flip, and rescaling are also used. The proposed ciGAN with traditional augmentation achieved an area under the curve (AUC) score of 0.896 while the real dataset with no augmentation achieves a 0.882 AUC score.

\subsubsection{Loss Function}

\paragraph{Adversarial}
In vanilla GANs, a cross-entropy loss is introduced that is usually described as an adversarial loss. This loss can cause a vanishing gradient problem. To address this problem, WGAN loss is introduced to utilize as an adversarial loss. (Please refer to section 2.4.3 (WGAN) for more detail). A similar study was found in the task of reconstructing low-dose PET images from full-dose PET images \cite{zhao2020study}. Authors \cite{zhao2020study} use a 1-Wasserstein distance instead of cross-entropy in supervised CycleGAN namely S-CycleGAN to improve the training stability of the proposed network. To evaluate the quality of generated low-dose images, authors \cite{zhao2020study} utilized a learned perceptual image patch similarity (LPIPS) score. The lower value of the score shows better image quality regarding the actual image patches. The S-CycleGAN achieved a 0.026 LPIPS score which is small compared to the actual low-dose PET images of 0.035. The results show better performance of S-CycleGAN regarding training stability.

Saad et al. \cite{saad2022self} proposed a novel MSG-SAGAN with a relativistic hinge loss function. Relativism in the hinge loss helps the discriminator to improve its learning using approximate predictions of the real images as half of the images are fake on average instead of taking them all as real. This prior training information helps the discriminator to classify and predict the real and fake images more accurately and stabilizes the training of MSG-SAGAN. An improved FID score of 139.6 is achieved for X-ray image synthesis using the proposed MSG-SAGAN.

\paragraph{Regularization}
This section elaborates on the use of regularization terms with additional loss functions in GANs to stabilize the training of GANs.

Gradient penalization (GP) is used to force the discriminator for producing meaningful gradients. For this, the discriminator D is enforced to be Lipschitz continuous \cite{gulrajani2017improved}. GP enables D to target the $\|D\|_{L i p}$ to 1. The $\|D\|_{L i p}$ is defined as Lipschitz continuity as shown in Eq. \eqref{Lip} reproduced from \cite{lee2020regularization}.

\begin{equation}
\frac{\left|D\left(x_{1}\right)-D\left(x_{2}\right)\right|}{\left|x_{1}-x_{2}\right|} \leq K\label{Lip}
\end{equation}

In the Eq. \eqref{Lip}, $\|D\|_{L i p}$ denotes the left side of the equation. K is the real constant, known as Lipschitz constant \cite{lee2020regularization}, and implies within the range $K \geq 0$ where $\forall x_{1}, x_{2} \in \mathbb{R}^{p}$.
To address the training instability problem, GP is applied as a $\mathcal{L}_{G P}$ using $L$2 norm. The $\mathcal{L}_{G P}$ is defined as $\mathbb{E}_{\hat{x}}\left[\left(\left\|\nabla_{\hat{x}} D(\hat{x})\right\|_{2}-1\right)^{2}\right]$. In this way, gradients that vary from one are penalized.

The gradient penalty regularization term is investigated by Gulrajani et.al \cite{gulrajani2017improved} with WGAN loss to improve the training stability of the network.

In the biomedical imaging domain, WGAN-GP loss is used as an additional loss in many biomedical image analysis tasks such as synthesis of cervical histopathology images \cite{xue2019synthetic} and MR images \cite{segato2020data} \cite{kwon2019generation} to improve training stability. In multi-scale Gradient GAN (MSG-GAN) \cite{deepak2020msg}, a WGAN-GP loss is used to train the MSG-GAN and improved the training stability.

\subsubsection{Summary}

\begin{table}[htp!]
\centering
\caption{\textbf{An overview of existing solutions via preprocessing, modified architectures, adversarial training, and loss functions to address the instability problem in GANs for biomedical image analysis.}}

\begin{tabular}{lp{1cm}p{1.4cm}p{0.3cm}p{0.3cm}p{0.3cm}p{0.3cm}p{7cm}}

\toprule

\textbf{References} & \textbf{GANs Variant} & \textbf{Image Modality} & \textbf{Pre-proc:} & \textbf{Mod. Arch.} & \textbf{Adv. Tr.} & \textbf{Ls. Funs.} & \textbf{Summary of Solutions} \\

\midrule

 Xue et al. \cite{xue2019synthetic} & Modified CGAN & Histopathology Images & x & x & x & \checkmark & WGAN-GP loss is used to address the vanishing gradient problem in CGAN for the generation of high-quality Histopathology images. It uses a Wasserstein distance to train the generator and discriminator models. \\
& \\
 Segato et al. \cite{segato2020data} & DCR-AEGAN & MR Images & x & x & x & \checkmark & WGAN-GP loss is used to stabilize the training of DCR-AEGAN. It helps a DCR-AEGAN to generate 3D MR images using the WGAN-GP distance evaluation during the training of GAN. \\
& \\
 Kwon et al. \cite{kwon2019generation} & AEGAN & MR Images & x & x & x & \checkmark & WGAN-GP loss is used to train the AEGAN for 3D MR image synthesis. WGAN-GP provides stabilized training for generating complex 3D MR images using Wasserstein distance measures. \\
& \\
 Wei et al. \cite{wei2020predicting} & CF-SAGAN & MR Images & x & \checkmark & x & x & The residual connections are used to address the training instability problem for MR-PET image translation task. The residual connections use ResNet-based long and short skip connections in the generator to recover the lost spatial information during the down-sampling operations. It helps to mitigate the problem of vanishing gradient by allowing an alternative track for the gradient to flow during the training of GAN. \\
& \\
 Wu et al. \cite{ciGAN2018} & ciGAN & Mammography Images & x & \checkmark & x & x & A multi-scale generator is proposed to address the instability problem in the ciGAN to generate segmented Mammography images. The proposed generator uses a cascaded refinement network that helps to generate features at multiple scales before being concatenated. This process improves the training stability for the generation of high-resolution synthetic images. \\
& \\
 Zhao et al. \cite{zhao2020study} & S-CycleGAN & PET Images & x & x & x & \checkmark & A loss function of 1-Wasserstein distance is used instead of cross-entropy to stabilize the train of CycleGAN for PET image translation. Wasserstein distance helps to generate high-quality synthetic images using the distance evaluation of real and synthetic images during the training of GANs. \\
& \\       
 Deepak et al. \cite{deepak2020msg} & MSG-GAN & MR Images & x & x & x & \checkmark & WGAN-GP loss is used to stabilize the training of  GAN. WGAN-GP loss uses a Wasserstein distance to evaluate the performance of the generator and discriminator models during the training of GAN for finding the stabilized training to generate high-quality synthetic MR images. \\

\bottomrule
\multicolumn{8}{l}{Pre-proc: Pre-processing; Mod. Arch: Modified Architectures; Adv. Tr: Adversarial Training; Ls. Func: Loss Function}
\end{tabular}
\label{ganoverviewhowinstability}
\end{table}

\begin{figure}
    \centering
\begin{center}
        \resizebox{0.5\textwidth}{!}{%
            \begin{forest}
                for tree={
                    draw,
                    rounded corners,
                    node options={align=center},
                    text width=2.7cm,
                    anchor=center,
                          },
                where level=0{%
      }{%
        folder,
        grow'=0,
        if level=1{%
          before typesetting nodes={child anchor=north},
          edge path'={(!u.parent anchor) -- ++(0,-20pt) -| (.child anchor)},
        }{},
        }
                [Proposed Solutions for Instability Problem \\ in GANs, fill=gray!30, parent
                [Modified \\ Architecture, for tree={fill=red!30, child}
                [Generator: \\ ciGAN \cite{ciGAN2018} (2018) \\ CF-SAGAN \cite{wei2020predicting} (2020)]]
                [Loss Function, for tree={fill=green!50, child}
                [Adversarial: S-CycleGAN \cite{zhao2020study} (2020) \\ MSG-SAGAN \cite{saad2022self} (2022)]
                [Regularization: \\ Modified CGAN \cite{xue2019synthetic} (2019) \\ AEGAN \cite{kwon2019generation} (2019) \\ DCR AE Alpha GAN \cite{segato2020data} (2020) \\ MSG-GAN \cite{deepak2020msg} (2020)]
                ]]
            \end{forest}
        }
    \end{center}
    \caption{Taxonomy of different proposed solutions for addressing the instability problem of GANs in biomedical imagery analysis.}
    \label{fig: ins}
    \end{figure}
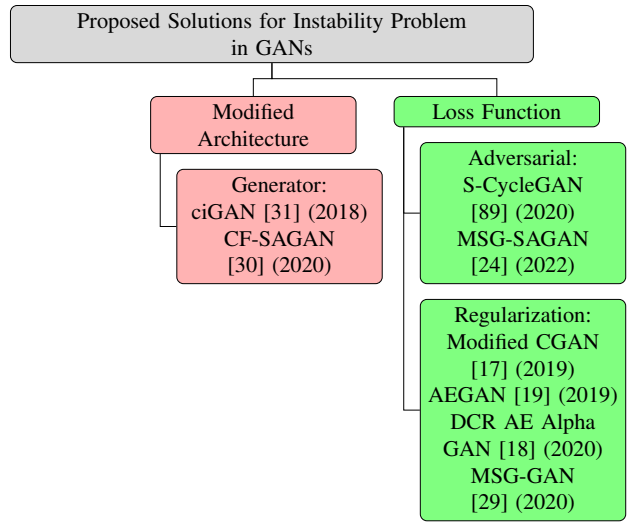
    
\begin{figure}
    \centering
\begin{center}
        \resizebox{0.5\textwidth}{!}{%
            \begin{forest}
                for tree={
                    draw,
                    rounded corners,
                    node options={align=center},
                    text width=2.7cm,
                    anchor=center,
                          },
                where level=0{%
      }{%
        folder,
        grow'=0,
        if level=1{%
          before typesetting nodes={child anchor=north},
          edge path'={(!u.parent anchor) -- ++(0,-20pt) -| (.child anchor)},
        }{},
        }
                [Instability Problem in Different Applications \\ of GANs, rotate = 0, fill=gray!30, parent
                [Image \\ Segmentation, rotate = 0, for tree={fill=red!30, child}
                [Mammography Images: \\ ciGAN \cite{ciGAN2018} (2018), rotate = 0]
                ]
                [Image Synthesis, rotate = 0, for tree={fill=green!50, child},  calign with current edge
                [Conditional \\ Image Synthesis: \\ CF-SAGAN \cite{wei2020predicting} (2020), rotate = 0]
                [Unconditional \\ Image Synthesis, rotate = 0,
                [MR Images: \\ MSG-GAN \cite{deepak2020msg} (2020), rotate = 0]
                [X-ray Images: \\ MSG-SAGAN \cite{saad2022self} (2022), rotate = 0]
                ]]
                [Image \\ Reconstruction, rotate = 0, for tree={fill=cyan!40, child}
                [PET Images \\ S-CycleGAN \cite{zhao2020study} (2020), rotate = 0]
                ]]
            \end{forest}
        }
    \end{center}
    \caption{An application-based taxonomy of different approaches for addressing the instability problem of GANs in biomedical imagery analysis.}
    \label{fig: instab appl}
    \end{figure}
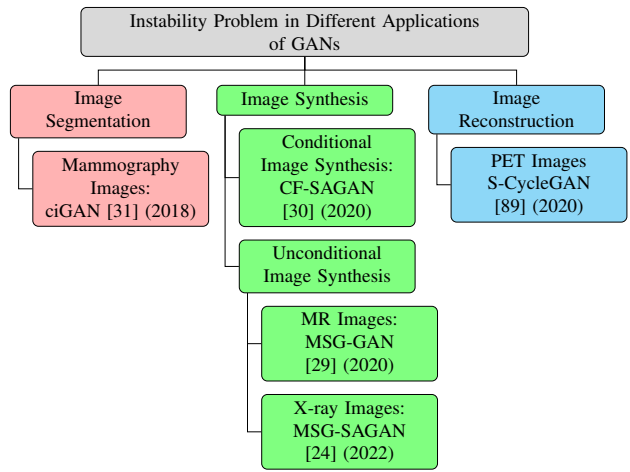

In this section, technical papers of GANs are reviewed to address the instability problem in the domain of biomedical imagery. The problem of unstable training triggers due to the vanishing gradient problem when the discriminator becomes optimal and sends no feedback to update the generator's weights as shown in Fig. \ref{Fig.1}. So, to stabilize the training of GANs, the generator should receive significant feedback in the form of gradients from the discriminator to produce high-quality realistic images. Considering this aim, many work solutions have been proposed in the domain of biomedical imaging. With this aim, the technical papers are classified into two taxonomies. The first is based on solutions in terms of modified architectures and loss functions as shown in Fig. \ref{fig: ins}. The second is based on the applications with different image modalities as shown in Fig. \ref{fig: instab appl}.

With modified architecture, technical papers provide their solutions by changing the generator either its layers such as \cite{wei2020predicting} or complete generator like \cite{ciGAN2018}. Both of the solutions provide a stable conditioned training of proposed GANs but found some artifacts generated in the output images. The loss function plays a key role to address the vanishing gradient problem. The reason behind this phenomenon is that the loss function backpropagated feedback in the form of gradients to update the generator weights. When the discriminator becomes optimal then its loss approaches zero which can't provide feedback to the generator. Technical papers are reviewed that provide solutions in biomedical imagery. In the loss function section, technical papers are further classified into adversarial loss \cite{zhao2020study} and regularization loss \cite{xue2019synthetic} \cite{segato2020data} \cite{kwon2019generation}. The WGAN loss is used as an adversarial loss in \cite{zhao2020study}. The WGAN-GP loss is used as regularization loss in \cite{xue2019synthetic} \cite{segato2020data} \cite{kwon2019generation} \cite{deepak2020msg} to address the instability problem in different application-based solutions.

To address the instability problem in the biomedical imaging domain, Table \ref{tabchalleng} shows a comparative analysis of different approaches provided in the literature. It is analyzed that WGAN-GP loss \cite{gulrajani2017improved} can be a suitable candidate to address the training instability problem in biomedical imagery as it works with various GANs architectures to alleviate the problem. The generated images can be obtained from GANs with high-quality and realistic nature.

Moreover, a detailed overview of existing solutions to address the instability problem in GANs is reported in Table \ref{ganoverviewhownonconverge} where the methodology of each solution to the instability problem in GANs is summarized for the domain of biomedical imagery.

\section{A Comparative Analysis of State-of-the-art GANs on COVID-19 Chest X-ray Image Dataset}
\label{section:comparisongans for covid}

\begin{table}[hbt!]
\centering
\ra{1.1}
\caption{\textbf{A comparison of GAN variants for generating diversified and high-quality synthetic biomedical images. Lower values of MS-SSIM and FID indicate a more diversified generation of synthetic images.}}
\begin{footnotesize}
      \begin{tabular}{llllll}
      \toprule
      \textbf{GAN Variant} & \textbf{Image Type} & \textbf{FID$\Downarrow$} & \textbf{MS-SSIM\_Real} & \textbf{MS-SSIM\_Synthetic$\Downarrow$} \\ \midrule
      DCGAN & X-ray Images & 223 & 0.50 & 0.48 \\
      AIIN-DGAN & X-ray Images & 200 & 0.50 & 0.37 \\
      MSG-GAN & X-ray Images & 160 & 0.50 & 0.46 \\
      MSG-SAGAN & X-ray Images & 135 & 0.50 & 0.45 \\ \bottomrule
      \end{tabular}
      \end{footnotesize}
      \label{gancomparison}
\end{table}

\begin{figure}[htp!]
    \centering
    \includegraphics[width=1\textwidth]{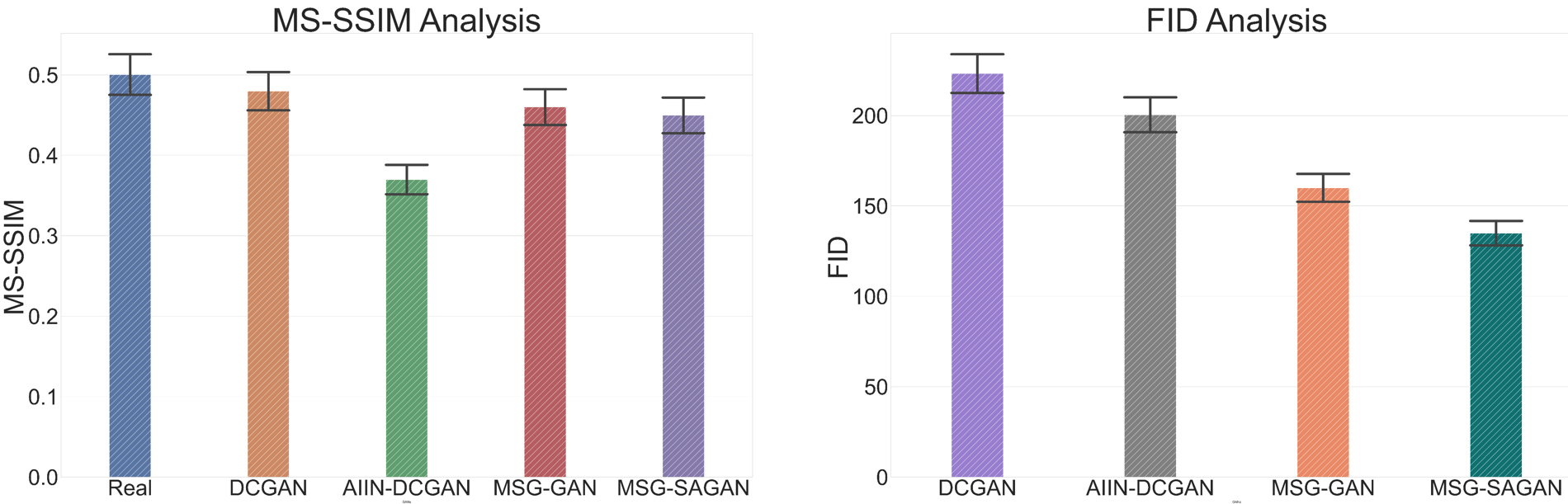}
    \caption{MS-SSIM and FID scores analysis of DCGAN, AIIN-DCGAN, MSG-GAN, and MSG-SAGAN models for COVID-19 chest X-ray images. A lower MS-SSIM score of synthetic images than real images indicates a better diversity and quality of synthetic images as compared to real images. A lower FID score indicates a better diversity and quality of synthetic images as compared to real images.}
    \label{Fig.MS_FID}
\end{figure}

In this section, state-of-the-art GAN architectures such as DCGAN \cite{Neffseg2017} \cite{kora2021evaluation}, AIIN-DCGAN \cite{saad2022addressing}, MSG-GAN \cite{deepak2020msg}, and MSG-SAGAN \cite{saad2022self} are re-implemented for the COVID-19 chest X-ray image dataset \cite{rahman2021exploring}. The same dataset of X-ray images is used to perform all these experiments. The X-ray dataset is selected because it is a widely used image modality to analyze the disease in human beings. X-ray provides a wide spectrum of affected parts of the body. These images are widely used by radiologists and clinicians to inspect the targeted segments of the disease. GANs have been utilized to train on these X-ray image datasets to generate synthetic images \cite{aggarwal2021diagnostic}.

Results are evaluated using two benchmark unified metrics such as MS-SSIM and FID. The MS-SSIM and FID metrics are selected because these metrics provide a compact evaluation of GAN's synthetic images as compared to real images. The MS-SSIM computes salient features of images such as structure, brightness, and contrast to measure the diversity of synthetic images. FID computes the distance based on Inception Version-3 between real and synthetic images. Therefore, a combination of these two metrics provides a fair and significant analysis of a GAN's performance regarding the generation of desirable synthetic images. Literature \cite{odena2017conditional} \cite{miyato2018spectraliclr} \cite{karnewar2020msg} \cite{han2020breaking} demonstrates that these two metrics have been widely adopted for the evaluation of GANs training challenges in the research community and are significant as compared to the alternate evaluation metrics in the domain of natural and biomedical imaging.

A Table \ref{gancomparison} comparing the experimental results of DCGAN, AIIN-DCGAN, MSG-GAN, and MSG-SAGAN for the X-ray image dataset using MS-SSIM and FID evaluation metrics is added. Moreover, a comparative analysis of these GAN architectures via bar graphs of MS-SSIM and FID scores is also shown in Fig. \ref{Fig.MS_FID}. The bar graphs show the impact of different solutions to the training challenges of GANs on the generation of synthetic X-ray images using MS-SSIM and FID scores. The adaptive input-image normalization technique has a significant impact on the MS-SSIM score because MS-SSIM considers the perpetual features of images to measure the similarity score. Therefore, synthetic X-ray images generated by AIIN-DCGAN indicate a lower score of MS-SSIM than the alternate solutions. However, a combined score of MS-SSIM and FID shows that MSG-SAGAN is the most performant GAN architecture to generate diversified synthetic X-ray images. The MSG-SAGAN has the advantages of self-attention and a multi-scale gradient learning scheme that enable the generator and discriminator models to guide each other significantly while focusing on learning the salient features of X-ray images.

\section{Challenges and Future Research Directions}
\label{section:challengesanddirections}

\begin{table}[htp!]
\centering
\caption{\textbf{An overview of experimental effects of benchmark GAN variants for biomedical image generation.}}

\begin{tabular}{p{1cm}p{1.5cm}p{1cm}p{1cm}p{5cm}p{5cm}}
\toprule
\textbf{GAN Variant} & \textbf{Image Type} & \textbf{Comp: Ct.} & \textbf{Mem: Con:} & \textbf{Pros} & \textbf{Cons} \\

\midrule

 DCGAN \cite{Neffseg2017} \cite{saad2022addressing} \cite{wu2018end} & X-rays, CT, Cell & Low & Low & DCGAN has a simple baseline architecture that is easy to understand and implement. It is memory efficient and generalized for several biomedical imaging modalities. & Limited for low-resolution images only. It has unstable training for complex images such as Histopathology or Cell images. It has no option for using conditional information on images. \\
 & \\
 CGAN \cite{xue2019synthetic} \cite{kudo2019virtual} \cite{goel2021automatic} \cite{biswas2019synthetic} & Histopathology, CT, and Retinal & Low & Moderate & CGAN utilizes moderate memory due to the usage of conditional information of input images. It is easy to understand and implement. It is generalized for several biomedical imaging modalities but is dependent on conditional information of imaging modalities. & Less efficient for high-resolution images. It has unstable training for high-resolution images with fine-grained information such as Dermoscopic images. \\
 & \\
 PGGAN \cite{karras2017progressive} \cite{abdelhalim2021data} & Dermoscopic (Skin lesion) & High & High & PGGAN has a complex architecture setting. It is easy to implement and shows efficient performance for all resolutions of images except 1024x1024. It is generalized for several biomedical imaging modalities. & PGGAN uses high memory and computational cost due to its complex architecture compared to DCGAN. \\
& \\
SAGAN \cite{zhang2019self} \cite{abdelhalim2021data} \cite{saad2022self} \cite{wei2020predicting} & X-rays, MR, and Dermoscopic (Skin lesion) & High & High & SAGAN uses a self-attention in the architecture of DCGAN. It is easy to implement and shows efficient performance for all low resolutions of images. It is generalized for several biomedical imaging modalities. & SAGAN uses high memory and computational cost due to its complex architecture compared to DCGAN. It shows unstable training for several high-resolution imaging modalities. \\
& \\
StyleGAN \cite{karras2019style} \cite{qin2020gan} & Dermoscopic (Skin lesion) & High & High & StyleGAN uses a style generator to control the generation of synthetic images with salient features. It shows efficient performance for all resolutions of images except 1024x1024. It is generalized for several biomedical imaging modalities. & StyleGAN uses high memory and computational cost due to its complex architecture compared to DCGAN and PGGAN. \\
& \\
MSG-GAN \cite{karnewar2020msg} \cite{saad2022self} \cite{deepak2020msg} & X-rays and MR & High & High & MSG-GAN shows a stabilized training for all resolutions of images including 1024x1024. & MSG-GAN uses high memory and computational cost due to its complex architecture compared to DCGAN, PGGAN, and StyleGAN. It shows unstable training for a few biomedical imaging modalities. \\
& \\
SRGAN \cite{ledig2017} \cite{xu2020low} & X-rays & High & High & SRGAN shows a stabilized training and efficient performance for all resolutions of images. & SRGAN uses high memory and computational cost due to its complex architecture. \\
& \\
CycleGAN \cite{modanwal2021normalization} \cite{zhao2020study} \cite{lau2018scargan} & MR and PET & High & High & CycleGAN has a stabilized training for image-to-image translation task. & It uses high memory and computational cost due to its complex architecture and shows unstable training for high-resolution and fine-grained biomedical imaging modalities such as PET and histopathology images. \\
& \\
AEGAN \cite{segato2020data} \cite{kwon2019generation} & MR Images & Moderate & Moderate & AEGAN has a stabilized training for several biomedical images. & AEGAN uses high memory and computational cost for 3D MR images and shows unstable training for high-resolution biomedical imaging modalities such as MR images. \\

\bottomrule
\multicolumn{6}{l}{Comp: Ct: Computational Cost; Mem: Con: Memory Consumption}
\end{tabular}
\label{ganoverviewimplementeffect}
\end{table}

The implementation effects such as computational cost, memory consumption, and pros and cons for alternate biomedical imaging modalities of benchmark GANs architectures have been added in Table \ref{ganoverviewimplementeffect}. Table \ref{ganoverviewimplementeffect} provides a comprehensive overview of experimental effects that impact during the re-implementation of these architectures for diverse biomedical imaging modalities. This table will guide a reader to the best appropriate GAN architecture for targeted biomedical images.

\subsection{The Mode Collapse Problem}
In biomedical image analysis, the mode collapse problem is one of the severe problems that occur during the training of GANs. The mode collapse problem has a direct impact on the diversity of synthetic images generated by GANs. Synthetic images lack diversity as compared to real images. Due to this problem, the generator in the GAN misses salient features of the image and repeats the same features in the generation of new synthetic images. It is challenging for researchers to train a GAN completely to avoid the mode collapse problem and its subsequent impact on the synthetic images. The underlying problem behaves differently for a number of GAN-based applications of biomedical image analysis. For example, a mode collapse occurs when a GAN uses a segmented mask with ground truth chest radiographs to generate segmented radiographs. Similarly, significant features of cell images can be affected and missed during the GAN-based generation of synthetic images. The mode collapse problem also occurs due to the complexity of 3-dimensional brain MR images in the process of image synthesis. For instance, modifications in GANs such as perceptual image hashing \cite{Neffseg2017}, the mixture of distributions in the generator \cite{wu2018end}, and VAEGAN-based architectures \cite{segato2020data} have been used to alleviate the mode collapse problem. In biomedical imaging applications, GANs can also cause feature hallucinations when generating new synthetic data \cite{laino2022generative}. Hallucination in GANs refers to the generation of novel, unwanted artificial features or missing significant features in synthetically generated images that can lead to the risk of misdiagnosing diseases \cite{wolterink2021generative}. The hallucinated features are generated due to the problem of mode collapse in GANs. Hallucinated features are usually generated in the synthetic images when performing the image-to-image translation task \cite{cohen2018distribution}. The solutions for alleviating mode collapse can also reduce the effects of hallucination in synthetic biomedical images.

In GANs, several techniques have been used to address the mode collapse problem in biomedical image analysis. It is critical for a GAN to train the generator and the discriminator in such a manner that the generator can learn a complete distribution of features and anatomical structure of biomedical images while the discriminator returns constructive feedback to the generator. The modifications in the generator or discriminator architectures or their loss functions can alleviate the mode collapse but do not solve the problem completely. Thus, there is a research gap to find a significant solution either based on architecture or loss function that should be capable of addressing the mode collapse problem in biomedical image analysis. The proposed solutions may consider the performance of generated images to analyze the effect of mode collapse. The analysis of generated images can better direct researchers to propose an effective solution in this field. However, it is important to address the mode collapse problem during the training of GANs so that the GAN-based applications can be utilized effectively in biomedical image analysis. Future research directions include modified architectures based on state-of-the-art attention networks, novel regularization techniques, capsule networks, and advanced normalization techniques to address the mode collapse problem in biomedical image generation. Autoencoders are also recognized as a significant technique to address the mode collapse problem in GANs, but it generates blurry images. Nevertheless, autoencoders with powerful discriminators can improve the existing solutions in the biomedical imaging domain.

\subsection{The Non-convergence Problem}
In GANs, the non-convergence is a major failure of the generator and the discriminator models to reach an unbalanced state. When the training of GANs becomes unbalanced, there is a direct impact on the performance of synthetic images generated by GANs. Synthetic images can be generated blurry or with artifacts. It is very critical to train a GAN in a way that both models train in a balanced state during the whole training time. One solution is to reach a Nash equilibrium. It is very difficult to reach Nash equilibrium in practice. The issue is that a GAN sticks to the saddle point where the objective function gives minimal weight parameters for one model while the maximal weight parameters are for the other model. However, a minimax game can be used to find a Nash equilibrium. In biomedical image analysis, researchers devise new methodologies to address the non-convergence problem. Like, optimization algorithms such as Whale optimization, improving learning rate, and novel updating algorithms for training the generator and the discriminator have been used.       

The non-convergence problem is a potential challenge faced by GANs during training. Updating algorithms proposed in vanilla GAN are limited to their initial experiments. The updating algorithm of WGAN can work for a few applications to achieve a Nash equilibrium. Similarly, TTUR and hyperparameter optimization techniques can also work for limited architectures and lack generalization ability. So, there is a need for a compact and generalized solution to achieve the Nash equilibrium during the training of GANs. Recently, non-convergence is a generic problem for GANs, and researchers use JS divergence to find a balanced state during the training of GANs that is difficult to achieve in practice. Different techniques have been proposed to cope with this problem, such as f-divergence and improved Wasserstein loss functions that still need improvement. These approaches can be used with different GANs architectures to address the underlying problem in biomedical image analysis. However, future research directions should focus on advancing JS divergence to balance the training of GANs while considering different optimization techniques such as stochastic gradient descent, Pareto-optimality, etc. Novel game theories with divergences can also be explored based on existing schemes that will be helpful for GANs to address the non-convergence problem.

\subsection{Instability Problem}
The training stability of GANs is important to achieve for any GAN-based application of biomedical image analysis. The problem occurs due to the vanishing gradient problem. Thus, there are solutions proposed to address the vanishing gradient problem such as modified architectures and modified loss functions. The loss function has a great impact to stabilize the training of GANs so WGAN-GP loss \cite{gulrajani2017improved} is analyzed in almost all of the reviewed technical papers. The WGAN-GP loss helps in acquiring stable training of GANs in the technical solutions but there is no guarantee or generalization criterion about its suitability and utility for other applications as well as other imaging modalities. It is important to consider that if GANs can handle the training strategy to achieve the Nash equilibrium and try to reach the optimal discriminator then a vanishing gradient problem gets triggered due to the optimality of the discriminator as discussed in section VI-A. It is also suspected that the stability of training depends on the mode collapse and non-convergence problems as well but sometimes, it can be seen that architecture is trained in stable conditions but has been affected by mode collapse. So, this situation could be a question of the performance of GANs. Therefore, all of these technical training challenges must be addressed in biomedical image analysis.

Future research directions should consider the above-mentioned constraints and propose novel techniques to address the instability problem in the biomedical imaging domain. There have been several approaches that are experimented with GANs to stabilize the training while addressing the vanishing gradient problem. There is a need for devising novel regularization, normalization, and game theory techniques to be used in the GANs which are unexplored previously. WGAN-GP is a widely used loss to cope with this problem in the general imaging domain yet requires more work and modifications to reach the stable training of GANs. Hybrid multiple GAN-based architectures based on WGAN-GP loss, attention mechanisms, novel regularization, and optimization techniques can also be explored to address the underlying problem. In recent times, alternate generative models such as diffusion models have also become popular in the domain of biomedical imaging \cite{ali2023spot}. Diffusion models have their pros and cons in terms of training stability, computational cost, diversity of synthetic images, and learning high-dimensional latent spaces as compared to GANs. A detailed review of several applications of diffusion models has been conducted for biomedical images \cite{kazerouni2022diffusion}. It is an important research direction to investigate the solutions to the GANs training challenges for addressing the training challenges of diffusion models in biomedical image analysis.

\subsection{Evaluation Metrics}
In GANs, evaluation metrics play a key role to represent the performance of GANs. These metrics provide a quantification of the problems such as mode collapse, non-convergence, and training instability during the training of GANs. Although, evaluation metrics like IS, FID, MS-SSIM, MMD, and PSNR have been used to evaluate the performance of GANs based on the generated images. Nevertheless, these metrics are application-dependent and lack the capacity in visualizing the occurrence of the challenges during the training of GANs.

In relation to the training challenges of GANs, evaluation metrics are used to capture the diversity and quality of the generated images. Generally, for the mode collapse problem, the diversity of images is quantified by the IS, MS-SSIM, and MMD metrics. While, for the non-convergence and instability problems, PSNR and FID are used. IS and FID metrics are frequently used to evaluate generated images via the quality of images. Both of these metrics are pretrained on ImageNet \cite{deng2009imagenet} dataset. The ImageNet dataset lacks the class of biomedical images thus IS and FID metrics are not recommended to be used in the biomedical imaging domain. Similarly, MS-SSIM is a human perceptual metric that only considers luminance and contrast estimations to measure the similarity of image features between two images. PSNR is a widely used metric to measure the quality of images but is limited to monochrome images. In biomedical image analysis, the performance parameters vary based on the type of imagery domain as every domain-specific images have different image features and properties.

Several unified evaluation metrics such as MS-SSIM, IS, MMD, FID, and PSNR have been utilized to evaluate the training performance of GANs based on the nature of application tasks in different imaging domains. There are two evaluation methods used to measure the performance of GANs. One method includes the direct comparison of the real image dataset to the synthetic image dataset using distance-based evaluation metrics such as IS, MMD, and FID. Another method is to measure the similarity and diversity of synthetic images using some features-based evaluation metrics such as SSIM, MS-SSIM, and PSNR. The score of the evaluation metric of the synthetic image dataset is compared with the score of the similar evaluation metric of the real image dataset. However, the deep learning community mostly relies on using two benchmark metrics such as MS-SSIM for feature-based evaluation and FID for distance-based evaluation to measure the performance of GANs \cite{borji2019pros}.

In biomedical image analysis, researchers utilize traditional pixel-wise evaluation metrics to quantify the performance of GANs. Most traditional metrics are suitable for supervised learning tasks that require reference images. In the biomedical imagery domain, the availability of reference images is limited due to privacy issues and inaccurate manual annotation. This ensures the use of unsupervised learning in the biomedical imagery domain. Furthermore, it is also important to evaluate the training performance of GANs because of the randomization of initialization, optimization, and technical challenges. The evaluation of generated images as compared to real images remains challenging and needs to be explored. There has been a list of metrics reported in \cite{borji2019pros} to evaluate the performance of GANs. In spite of all these proposed metrics, still there is a research gap in finding a metric that can capture the salient features such as the texture and shape of objects in the biomedical images. It is important to analyze the symptoms of each individual training problem of GANs for a number of applications in biomedical image analysis. An evaluation metric that can capture the pre and post-training dynamics of a GAN model is important to investigate. The proposed metric should work with most of the image modalities such as X-rays, MR images, Dermoscopic images, Ultrasound, and PET images to measure the efficacy of GANs in the domain of biomedical imaging.

\section{Conclusion}
\label{section:conclusion}
In this survey, training challenges of GANs such as mode collapse, non-convergence, and instability have been reviewed in detail for the domain of biomedical imagery. The three challenges are discussed via definitions, identifications, quantifications, and possible solutions. To address these training challenges in the biomedical imagery domain, technical literature has been discussed based on applications and solutions taxonomies. Existing literature shows that addressing these challenges entirely is a challenging task, but few techniques have been proposed that can partially alleviate these training challenges. In the architecture of GANs, the mode collapse problem can be addressed by using minibatch discrimination, skip connections, VAEGAN as part of the generator and discriminator, varying layers of generator and discriminator, spectral normalization, perceptual image hashing, Gaussian mixture model as a generator, discriminator with conditional information vector, self-attention mechanism, and adaptive input-image normalization. The non-convergence problem can be addressed by using modified training updates of the generator and discriminator, the Whale optimization algorithm, and two time-scale update rules. The instability problem can be addressed by using the Wasserstein loss, residual connections, multi-scale generator, and Relativistic hinge loss. Each solution contributes to alleviating the mode collapse problem based on the type of GAN architecture. The effectiveness and suitability of the solutions also depend on the types of GAN architecture and biomedical imagery. Moreover, this survey also elaborated that how each training problem can affect the quality of generated biomedical images in terms of realistic nature, diversity, resolution, and artifacts. This survey also highlights possible future research directions to address the underlying training challenges of GANs for biomedical images. In this survey, it is concluded that all three technical challenges faced during the training of GANs need more research work to bridge this gap for biomedical image analysis. This motivates the researchers to propose advanced solutions to address the underlying training challenges of GANs in the domain of biomedical imagery.



\end{document}